\documentclass{ecai} 

\usepackage{latexsym}
\usepackage{amsthm}
\usepackage{booktabs}
\usepackage{enumitem}
\usepackage{graphicx}

\usepackage{amsmath,amssymb,amsfonts}
\usepackage{algorithm}
\usepackage{algpseudocode}
\usepackage{subcaption}
\usepackage{multirow}
\usepackage{colortbl}
\usepackage{setspace}
\usepackage[dvipsnames]{xcolor}

\usepackage[pagebackref,breaklinks,colorlinks,citecolor=cvprblue]{hyperref}
\definecolor{cvprblue}{rgb}{0.21,0.49,0.74}

\newcommand{\BibTeX}{B\kern-.05em{\sc i\kern-.025em b}\kern-.08em\TeX}

\def\eg{\emph{e.g.} } 
\def\ie{\emph{i.e.} }

\begin{document}


\begin{frontmatter}




\title{TSFool: Crafting Highly-Imperceptible Adversarial Time Series through Multi-Objective Attack}


\author[A]{\fnms{Yanyun}~\snm{Wang}}
\author[B,*]{\fnms{Dehui}~\snm{Du}}
\author[A,*]{\fnms{Haibo}~\snm{Hu}}
\author[A]{\fnms{Zi}~\snm{Liang}}
\author[B]{\fnms{Yuanhao}~\snm{Liu}}

\address[A]{Department of Electrical and Electronic Engineering, The Hong Kong Polytechnic University \vspace{0.1cm}}
\address[B]{Software Engineering Institute, East China Normal University \vspace{0.15cm}}
\address[*]{Corresponding Authors. Email: dhdu@sei.ecnu.edu.cn; haibo.hu@polyu.edu.hk}


\begin{abstract}
Recent years have witnessed the success of recurrent neural network (RNN) models in time series classification (TSC). However, neural networks (NNs) are vulnerable to adversarial samples, which cause real-life adversarial attacks that undermine the robustness of AI models. To date, most existing attacks target at feed-forward NNs and image recognition tasks, but they cannot perform well on RNN-based TSC. This is due to the cyclical computation of RNN, which prevents direct model differentiation. In addition, the high visual sensitivity of time series to perturbations also poses challenges to local objective optimization of adversarial samples. In this paper, we propose an efficient method called TSFool to craft highly-imperceptible adversarial time series for RNN-based TSC. The core idea is a new global optimization objective known as ``Camouflage Coefficient" that captures the imperceptibility of adversarial samples from the class distribution. Based on this, we reduce the adversarial attack problem to a multi-objective optimization problem that enhances the perturbation quality. Furthermore, to speed up the optimization process, we propose to use a representation model for RNN to capture deeply embedded vulnerable samples whose features deviate from the latent manifold. Experiments on 11 UCR and UEA datasets showcase that TSFool significantly outperforms six white-box and three black-box benchmark attacks in terms of effectiveness, efficiency and imperceptibility from various perspectives including standard measure, human study and real-world defense.
\end{abstract}

\end{frontmatter}


\section{Introduction}\label{sec:intro}

\begin{figure}[htb]
\centering
\includegraphics[width=8.6cm]{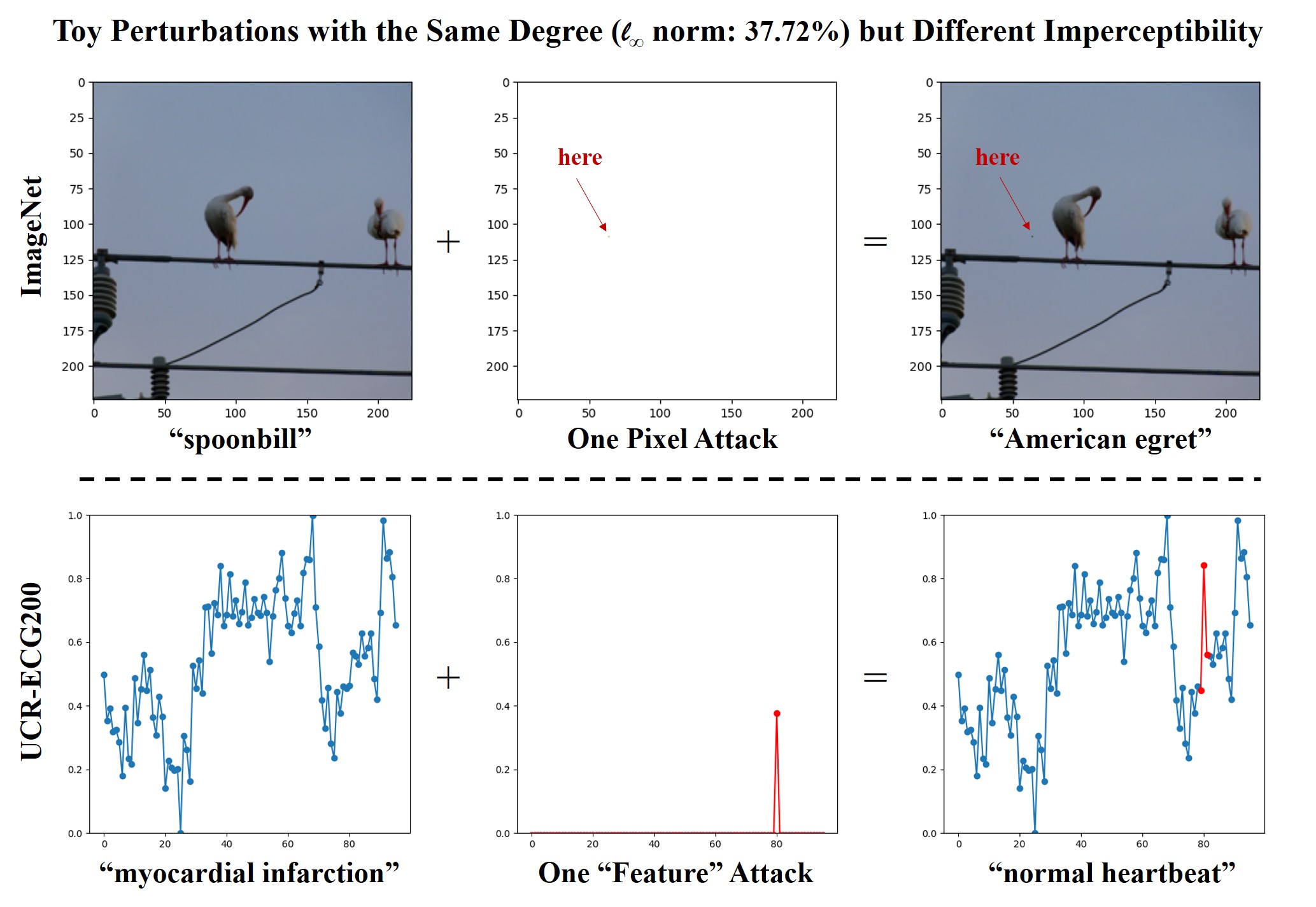}
\vspace{0.0001cm}
\caption{The figure shows that image and time series data have significantly different visual sensitivities to perturbations of the same level. The upper line is an example of One Pixel Attack \citep{su2019one} against a pre-trained ResNet-18 \citep{marcel2010torchvision} on ImageNet \citep{deng2009imagenet}. Although the $\ell_{\infty}$ norm of the generated perturbation is up to 37.72\%, it is still hard to be noticed by human eyes. On the contrary, as the lower line, a similar attack (\ie also merely perturb one feature by 37.72\% under $\ell_{\infty}$ norm) against one of our experimental RNN classifiers on UCR-ECG200 time series dataset \citep{dau2019ucr} is more than visible. This makes it more difficult to craft imperceptible adversarial time series than images. Please notice that this is just a toy case to reveal the current gaps, instead of the illustration of our proposed approach. \vspace{0.8cm}}
\label{fig:new1}
\end{figure}


NNs are vulnerable to adversarial attacks \citep{szegedy2014intriguing}, which means imperceptible perturbations added to the input can cause the output to change significantly \citep{huang2020survey}. A rich body of adversarial attacks has been investigated to generate adversarial samples that can enhance the robustness of the models through adversarial training. For instance, gradient-based attacks \citep{carlini2017towards, goodfellow2015explaining, kurakin2017adversarial, madry2018towards, moosavi2016deepfool, papernot2016limitations} have achieved impressive performance on feed-forward NN classifiers and image recognition tasks. Nevertheless, those gradient-based adversarial attacks cannot perform well in RNN where the unique time recurrent structure of RNN prevents differentiation on it \citep{papernot2016crafting}. Furthermore, existing attacks always adopt a local optimization objective to minimize the perturbation amount for every single sample. However, as shown in Figure~\ref{fig:new1}, time series are far more visually sensitive to perturbation than image data, undermining the effectiveness of these attacks in RNN models for time series. 

Due to these challenges, despite the popularity of TSC and RNN as its solution \citep{ding2022towards, ding2023black, wu2022small}, to date, there is yet not much effective study on crafting qualified adversarial samples for RNN-based TSC \citep{ding2023black, galib2023susceptibility, karim2020adversarial}. A few works focus on making RNN completely differentiable by \textit{cyclical computational graph unfolding} \citep{mozer2013focused, papernot2016crafting}, which turns out to be inefficient and hard to stably scale \citep{campos2018skip, gao2018low, zhang2016architectural}. Black-box attacks by model querying \citep{andriushchenko2020square, brendel2018decision, chen2020hopskipjumpattack, su2019one} or adversarial transferability \citep{papernot2016transferability, papernot2017practical} are either inapplicable to RNN-based TSC or very limited in effectiveness.

In this paper, we propose an efficient method named TSFool to craft highly-imperceptible adversarial samples for RNN-based TSC. With an argument that local optimal perturbation under the conventional objective does not always lead to imperceptible adversarial samples, we propose a novel global optimization objective named Camouflage Coefficient, and add it to reduce the adversarial attack problem to a multi-objective optimization problem. In this way, we can take the relative position between adversarial samples and class clusters into consideration, to measure the imperceptibility of adversarial samples from the perspective of class distribution. Since the full gradient information of an RNN is not directly available, to efficiently approximate the optimization solution, we introduce a representation model built only upon the classifier's outputs. It can fit the manifold hyperplane of a classifier but distinguish samples by their features like humans, and capture deeply embedded vulnerable samples whose features deviate from the latent manifold as guidance. Then we can pick target samples to craft perturbation in the direction of their interpolation, while imperceptibly crossing the classification hyperplane.

With six white-box and three black-box adversarial attacks from basic to state-of-the-art ones as the benchmarks, we evaluate our approach on 11 univariate or multivariate time series datasets respectively from the public UCR \citep{dau2019ucr} and UEA \citep{bagnall2018uea} archives. The results demonstrate that TSFool has significant advantages in effectiveness, efficiency and imperceptibility. In addition, extensive experiments on different hyper-parameters and settings confirm that our results are fair and of general significance. Beyond standard measures, we also conduct two human studies verifying the ability of Camouflage Coefficient to capture the real-world imperceptibility of adversarial samples, as well as implement four anomaly detection methods showing the imperceptibility of TSFool under real-world defense. Our main contributions are summarized as follows:

\vspace{-0.2cm}
\begin{itemize}[leftmargin=0.7cm]
  \item
  By exploring the visual sensitivity of time series data, for the first time, we point out the bias of the conventional local optimization objective of adversarial attack;
  \vspace{0.2cm}
  \item
  We propose a novel optimization objective named ``Camouflage Coefficient" to enhance the global imperceptibility of adversarial samples, with which we reduce the adversarial attack problem to a multi-objective optimization problem; and
  \vspace{0.2cm}
  \item
  We propose a general methodology based on \textit{Manifold Hypothesis} to solve the new optimization problem, and accordingly realize TSFool, the first method to our best knowledge, for RNN-based TSC, to craft real-world imperceptible adversarial time series.
\end{itemize}
\vspace{-0.2cm}

The rest of this paper is organized as follows. Section~\ref{sec:backg} reviews the related studies and explains the current gaps. A general methodology outlining our ideas is proposed in Section~\ref{sec:motiv}, followed by Section~\ref{sec:appro} to specifically realize the proposed TSFool in detail. Section~\ref{sec:evalu} evaluates the performance of TSFool. Section~\ref{sec:discu} provides relevant discussions and Section~\ref{sec:concl} concludes this paper.
Section~\ref{sec:sm_8}, Section~\ref{sec:sm_9} and Section~\ref{sec:sm_discu} in the Appendix respectively supplement more details about the approach, evaluation and discussion.


\vspace{0.05cm}

\section{Background and Related Work}\label{sec:backg}

\vspace{0.1cm}

\subsection{RNN-based Time Series Classification}\label{subsec:rnn_tsc}

TSC is an important and challenging problem in modern data mining \citep{ding2022towards, ismail2019deep}, with a variety of real-world applications including health care \citep{lin2019medical}, stock price prediction \citep{zhan2018stock} and food safety inspection \citep{fawaz2019adversarial}. Time series data consists of sampled data points taken from a continuous process over time \citep{langkvist2014review}. It can be defined as a sequence $X \in \mathbb{R}^{T \times D}$, where $T$ is the sequence length, also known as the number of time steps, and $D$ is the feature dimension of every single data point $x_{t} \in X$ ($t \in [1, T]$), based on which there are two categories of time series namely univariate series ($D = 1$) and multivariate series ($D > 1$) \citep{ding2022towards}. The key difficulty of TSC is to recognize the complex temporal pattern and characterize the temporal semantic information encoded in $X$ \citep{ding2022towards}. With the recurrent structure specifically designed for learning key temporal patterns, RNN has become one of the most state-of-the-art models for TSC \citep{wu2022small}. Given an RNN classifier $\mathcal{N} : \mathcal{X} \rightarrow \mathcal{Y}$ with hidden state $h_t \in \mathbb{R}^H$ and non-linear recurrent function $h_t = \mathcal{N}(h_{t-1}, x_t)$, for any time series data $X$, it takes each of the $x_{t} \in X$ as the input at time step $t$ in order and encodes the key information by update $h_t$ accordingly. Then the final state $h_T$ characterizes the whole series for classification.

\subsection{Adversarial Sample and Adversarial Attack}\label{subsec:basics}

The concept of adversarial attack is introduced by \citet{szegedy2014intriguing}, in which an adversarial sample $\Vec{x}^{*}$ crafted from a legitimate sample $\Vec{x}$ is defined by an optimization problem:
\begin{equation}\label{eq:ori}
    \Vec{x}^{*} = \Vec{x} + \delta_{\Vec{x}} = \Vec{x} + \min \Vert\Vec{z}\Vert \text{ s.t.} \hspace{0.15em} f(\Vec{x} + \Vec{z}) \neq f(\Vec{x}),
\end{equation}
where $f : \mathbb{R}^{n} \rightarrow \Vec{y}$ is the target classifier and $\delta_{\Vec{x}}$ is the smallest perturbation according to a norm appropriate for the input domain. Since exactly solving this problem by an optimization method is time-consuming \citep{moosavi2016deepfool} and not always possible, especially in NNs with complex non-convexity and non-linearity \citep{papernot2016crafting}, the common practice is to find the approximative solutions to estimate adversarial samples. 

The most commonly used adversarial attacks to date are gradient-based methods under white-box setting \citep{li2022review}, which means all the information of the target classifier is available and the perturbation is guided by the differentiating functions defined over model structure and parameters. For instance, the \textit{fast gradient sign method} (FGSM) \citep{goodfellow2015explaining} implements the perturbation according to the gradient of cost function $\mathcal{L}$ with respect to the input $\Vec{x}$:
\begin{equation}
    \delta_{\Vec{x}} = \varepsilon \hspace{0.1em} \text{sign}(\nabla_{\Vec{x}} \hspace{0.15em} \mathcal{L}(f, \Vec{x}, \Vec{y})),
\end{equation}
where $\varepsilon$ denotes the magnitude of the perturbation. The \textit{Jacobian-based saliency map attack} (JSMA) \citep{papernot2016limitations} further introduces forward derivative to quantitatively capture how a specific input component modifies the output. Then DeepFool \citep{moosavi2016deepfool} proposes a greedy idea to determine the perturbation direction of specific samples according to their closest classification hyperplane. As one of the state-of-the-art benchmarks to date \citep{athalye2018obfuscated}, the \textit{projected gradient descent attack} (PGD) \citep{madry2018towards} can be viewed as a variant of FGSM. It crafts samples by getting the perturbation as FGSM and projecting it to the $\epsilon$-ball of input iteratively. To be specific, given $\Vec{x}^{t}$ as the intermediate input at step $t$, PGD updates as:
\begin{equation}
    \Vec{x}^{t+1} = \Pi_{\epsilon} \left\{ \hspace{0.15em} \Vec{x}^{t} + \varepsilon \cdot \text{sign}(\nabla_{\Vec{x}} \hspace{0.15em} \mathcal{L}(f, \Vec{x}^{t}, \Vec{y})), \hspace{0.25em} \Vec{x} \hspace{0.15em} \right\}.
\end{equation}
In deep learning, it is sometimes \citep{carlini2017towards} referred to as \textit{basic iterative method} (BIM) \citep{kurakin2017adversarial}. Another state-of-the-art benchmark, C\&W \citep{carlini2017towards}, no longer uses the constraint to ensure small perturbation but formulates a regularization optimization for it. Given $y_0$ as the true label and $f(\Vec{x}_{i})_{y_k}$ as the prediction score of label $y_k$ for the candidate input $\Vec{x}_{i}$, then:
\begin{equation}
\resizebox{7.9cm}{!}{
$\begin{aligned}
    \Vec{x}^{*} \! = \! \mathop{\text{argmin}}\limits_{\Vec{x}_{i}} \{ \! \Vert \Vec{x}_{i} \! - \! \Vec{x} \Vert_{2}^{2} \! + \! \alpha \! \cdot \! \max \{ \! f( \! \Vec{x}_{i} \! )_{y_0} \! - \! \mathop{\max}\limits_{y_j \neq y_0} \! f( \! \Vec{x}_{i} \! )_{y_j}, \! 0 \! \} \! \},
\end{aligned}$
}\vspace{5cm}
\end{equation}
where $\alpha > 0$ controls the trade-off between small distortion and attack success rate. Finally, Auto-Attack \citep{croce2020reliable} forms a parameter-free and user-independent ensemble of attacks for frequent pitfalls in practice like improper tuning of hyper-parameters and gradient obfuscation or masking.

Nevertheless, model information including the gradient is not always available in practice \citep{ding2023black, li2022review}, so some black-box methods have been proposed, which can be roughly classified into two categories. The first one relies on the predictions returned by querying the target model. For instance, the Boundary Attack \citep{brendel2018decision} and its improvement HopSkipJump \citep{chen2020hopskipjumpattack} search for adversarial samples along the decision boundary. While the second one called Transfer Attack \citep{fawaz2019adversarial, papernot2017practical} trains a substitute model similar to the target model in performance, then attacks it instead to generate adversarial samples, and relies on their adversarial transferability to fool the target model \citep{papernot2016transferability}.

\subsection{Challenges for RNN and Time Series}\label{subsec:dilemma}

While existing adversarial attacks achieve great success on feed-forward NNs and image data, such success has not carried over to RNN-based TSC. There are two main reasons for this dilemma. Firstly, as Figure~\ref{fig:new1}, image data can tolerate respectively large real-world perturbations without being noticed by humans, while time series are so visually sensitive to perturbations that it is more difficult to ensure their imperceptibility \citep{wu2022small}. This exposes the drawback of existing methods in the control of perturbation on different data. It is also surprising that the only specialized measure for imperceptibility of adversarial time series is the number of time steps perturbed \citep{papernot2016crafting}, which is not sufficient.

Secondly, the presence of cyclical computations in RNN architecture prevents direct model differentiation \citep{papernot2016crafting}, which means most of the gradient information is no longer directly available through the \textit{chain rule}. This problem seriously challenges the effectiveness of all the gradient-based methods. An intuitive solution is to make the model differentiable through \textit{cyclical computational graph unfolding} to fit the gradient-based methods \citep{mozer2013focused, papernot2016crafting}. However, as the length of the time series is considerable in most real-world cases, the practical efficiency and scalability of the unfolding computation cannot be guaranteed \citep{campos2018skip, gao2018low, zhang2016architectural}. Another idea is to completely ignore the model knowledge given and view it as a black-box setting, to avoid the difficulty of specialized design for RNN. Nevertheless, a majority of the query-based black-box attacks like One Pixel Attack \citep{su2019one} and Square Attack \citep{andriushchenko2020square} can only work for image input. Although Boundary Attack and HopSkipJump are not subject to this limitation, both of them are extremely time-consuming because of the random-walking. On the other hand, Transfer Attack tends to achieve reasonable time and small perturbations, but unstable effectiveness. All in all, the problems brought by the perturbation sensitivity of time series data and the recurrent structure of RNN model remain to be further explored.

\section{Methodology}\label{sec:motiv}

In this section, we propose some novel ideas to build a general methodology for the crafting of highly-imperceptible adversarial samples. Notice that although these ideas are motivated and inspired by the problems exposed by RNN-based TSC, their underlying principles are not limited to this task. So to facilitate further studies in a larger range, we organize them here independently before specifically realizing them by the proposed approach in Section~\ref{sec:appro}.

\subsection{Rethinking Imperceptible Adversarial Sample}\label{subsec:reconsider}

The sensitivity of time series data prompts us to rethink the imperceptibility of perturbation. Firstly, existing methods always implement perturbations to all of the test data, while according to the specific class distribution, there must be individuals among them that are easier or harder to be perturbed respectively. A previous work has proven that a natural data point closer to (or farther from) the class boundary is less (or more) robust \citep{zhang2020geometry}. So without taking this difference into account and picking target samples appropriately, it is almost impossible for the attack method to stably control the perturbation amount by itself. On the contrary, it has to heavily depend on the specific dataset. 

\begin{figure}[htbp]
    \centering
    \includegraphics[width=4.9cm]{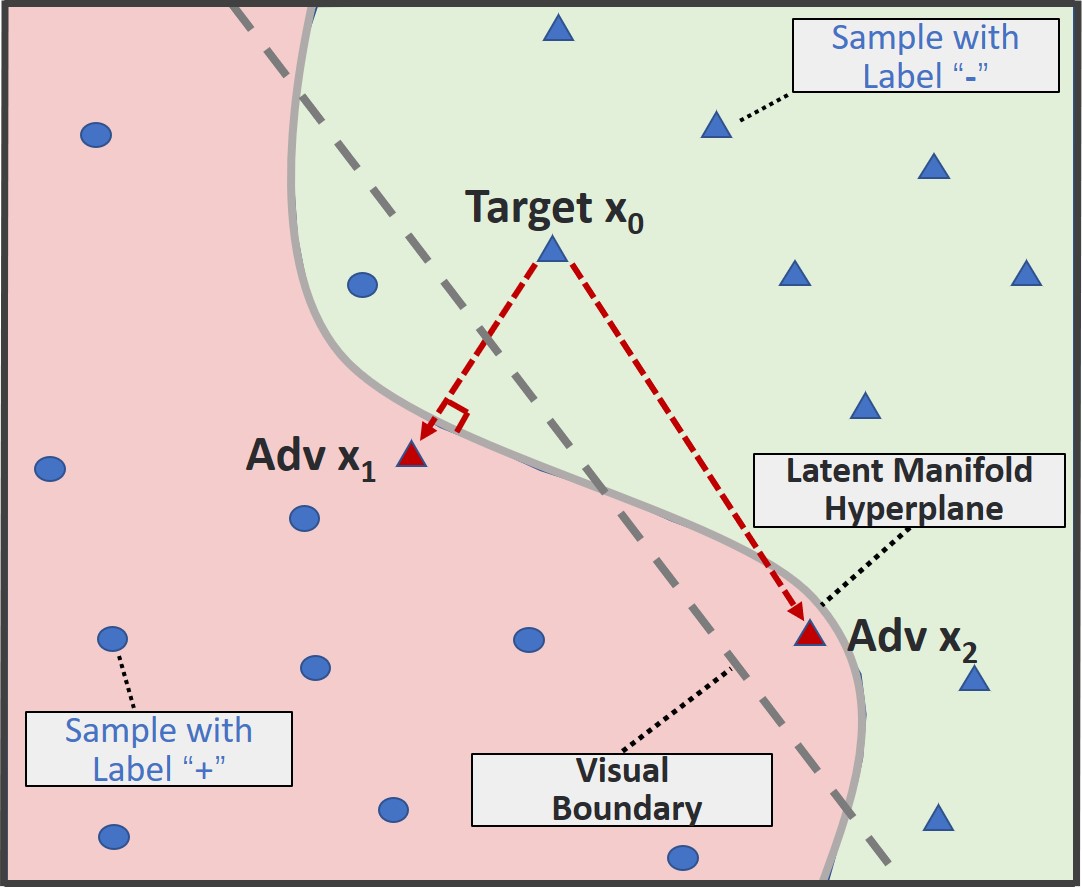}
    \vspace{0.3cm}
    \caption{An intuitive instance that the minimal perturbation (i.e. $\Vert \Vec{x}_1 - \Vec{x}_0 \Vert$) is not necessarily the most imperceptible one from the global perspective. The visual boundary and latent manifold hyperplane are different classification boundaries respectively from human eyes and the learned model. \vspace{0.6cm}}
    \label{fig:exp2}
\end{figure}

Secondly, we argue that the approximation to the optimal perturbation $\delta_{\Vec{x}}$ in Equation~(\ref{eq:ori}) does not always lead to a highly-imperceptible adversarial attack, while most of the existing methods view it as the basic target. This is because when we jump out of a single sample and consider from the perspective of class distribution, it can be found that even the minimal perturbation is not necessarily the most imperceptible one. For instance, in Figure~\ref{fig:exp2}, the adversarial sample $\Vec{x}_1$ is generated from $\Vec{x}_0$ through a minimal perturbation that crosses the closest classification hyperplane, while another adversarial sample $\Vec{x}_2$ is obviously more dangerous as it is semantically closer to and even belong to the original class.

\vspace{0.05cm}

These points inspired us to propose a novel definition of the imperceptibility of adversarial samples from the global perspective. Given a $k$-class classification task and the label $i, j \! \in \! \{0, 1, ..., k-1\}$, $\mathcal{X}_i$ is the set of all the samples belonging to class $i$ in the test set. For adversarial sample $\Vec{x}^{*}$ from $\Vec{x} \in \mathcal{X}_i$, which is wrongly predicted as $f(\Vec{x}^{*}) = j \hspace{0.2em} (j \neq i)$, we define its \textbf{Camouflage Coefficient} (\textbf{CC}) $\mathcal{C}({\Vec{x}^{*}})$ as:
\begin{equation}\label{eq:cc}
    \mathcal{C}({\Vec{x}^{*}}) = \frac{\Vert \Vec{x}^{*} - \Vec{m}_{i} \Vert / d_{i}}{\hspace{0.05em} \Vert \Vec{x}^{*} - \Vec{m}_{j} \Vert / d_{j}},
\end{equation}
where $m_i$ is the center of mass of class $i$ which is also built in the form of a legal sample:
\begin{equation}
    \Vec{m}_{i} = \frac{1}{|\mathcal{X}_i|} \sum_{\Vec{x}' \in \mathcal{X}_i} \Vec{x}',
\end{equation}
and $d_i$ is the average norm distance between $m_i$ and all the samples in $\mathcal{X}_i$, which is used to eliminate the potential bias from the different cluster sizes of the two classes:
\begin{equation}
    d_i = \frac{1}{|\mathcal{X}_i|} \sum_{\Vec{x}' \in \mathcal{X}_i} \Vert \Vec{x}' - \Vec{m}_{i} \Vert.
\end{equation}

As the saying goes, ``\textit{the best place to hide a leaf is the woods}''. The CC represents the proportion of the relative distance between adversarial sample $\Vec{x}^{*}$ and the original class to the relative distance between $\Vec{x}^{*}$ and the class to which it is misclassified. As a result, it can reveal to what extent an adversarial sample can ``hide'' in the original class without being noticed. The smaller the value, the higher the global imperceptibility of the adversarial sample. And if the value exceeds one, the attack somewhat fails. Because in this case, the misclassification of the adversarial sample is no longer surprising as it is already semantically closer to the samples in that wrong class instead of the original class. In our approach, we introduce CC as another optimization objective along with Equation~(\ref{eq:ori}) to craft adversarial samples considering local and global imperceptibility at the same time.

\subsection{Attack through Multi-objective Perturbation}

After adding the Camouflage Coefficient, the adversarial attack becomes a multi-objective optimization problem. Just as the existing methods, considering the efficiency, we do not solve it directly, but find the approximate solution in a logical and practical way. Separately speaking, to optimize Equation~(\ref{eq:cc}), we can just make $\Vec{x}^{*}$ closer to the $\Vec{m}_{i}$, which also means farther away from the $\Vec{m}_{j}$ in general. And to approximate the objective in Equation~(\ref{eq:ori}), perturbing in the direction of the closest classification hyperplane is proven to be a successful approach in DeepFool. So as a compromise to take them into account together, a reasonable idea is to cross a hyperplane that is relatively near the target sample $\Vec{x}$ on an appropriate place that is relatively close to the $\Vec{m}_{i}$.

To realize this idea in practice, we can find a sample $\Vec{x}_v$ from $\mathcal{X}_i$ which is misclassified to class $j$ but deeply embedded in class $i$ as a guide, and accordingly pick a correctly classified sample that is the closest to $\Vec{x}_v$ in class $i$ as the target $\Vec{x}$. Then when we perturb $\Vec{x}$ in the direction of $\Vec{x}_v$ to cross the classification hyperplane between them, we can not only acquire a considerable value of CC as the $\Vec{x}_v$ itself is a sample embedded closer to $\Vec{m}_{i}$ than to $\Vec{m}_{j}$, but also expect the hyperplane is relatively a close one to $\Vec{x}$ as this sample is the closest to $\Vec{x}_v$ among class $i$. In this way, the $\Vec{x}$ is not perturbed to be adversarial along the shortest path, but the more imperceptible one under the multi-objective definition. We define the $\Vec{x}_v$ and $\Vec{x}$ respectively as \textbf{vulnerable negative sample (VNS)} and \textbf{target positive sample (TPS)}, with which the perturbation can be denoted as:
\begin{equation}
    \delta_{\Vec{x}} = \lambda_\varepsilon \Vert \Vec{x}_v - \Vec{x} \Vert + \Vec{x}_\varepsilon,
\end{equation}
where $\lambda_\varepsilon \Vert \Vec{x}_v - \Vec{x} \Vert$ is the maximum interpolation that makes $\Vec{x}$ approach $\Vec{x}_v$ under a given step size $\varepsilon$ without changing its prediction result, and $\Vec{x}_\varepsilon$ is a micro random noise under the same $\varepsilon$ added to cross the classification hyperplane.

\subsection{Capturing Vulnerable Sample by Manifold}\label{subsec:manifold}

Now the only problem left is how to capture the VNS. Our idea to solve it comes from the \textit{manifold hypothesis}. It is one of the influential explanations for the effectiveness of NNs, which holds that many high-dimensional real-world data are actually distributed along low-dimensional manifolds embedded in the high-dimensional space \citep{fefferman2016testing}. This explains why NNs can find latent key features as complex functions of a large number of original features in the data. And it is through learning the manifold of the latent key features of training data that NNs can realize manifold interpolation between input samples and generate accurate predictions of unseen samples \citep{chollet2021deep}. So what matters in NN classification is how to distinguish the latent manifolds instead of the original features of samples with different labels. 

However, due to the limitation of sampling technologies and human cognition, practical label construction and model evaluation have to rely on specific high-dimensional data features. Accordingly, a possible explanation for the existence of adversarial samples is that the features of input data cannot always visually and semantically reflect the latent manifold, which makes it possible for the samples considered to be similar in features to have dramatically different latent manifolds. As a consequence, even a small perturbation in human eyes added to a correctly predicted sample may completely change the perception of NN to its latent manifold, and result in a significantly different prediction.

So if there is a representation model that can imitate the mechanism of a specific NN classifier to predict input data, but distinguish different inputs by their features in the original high-dimensional space just like a human, then it can be introduced to capture deeply embedded vulnerable samples whose features deviate from latent manifold. Specifically, when the prediction of representation model is correct while that of NN classifier is wrong, it means that compared with other samples that are similar in features and belong to the same class in fact, the current sample is perceived by NN at a different manifold cluster, and as a result, wrongly predicted. Such a sample well-meets the definition of VNS.

\section{Proposed Approach}\label{sec:appro}

\begin{figure}[htb]
\centerline{\includegraphics[width=8.5cm]{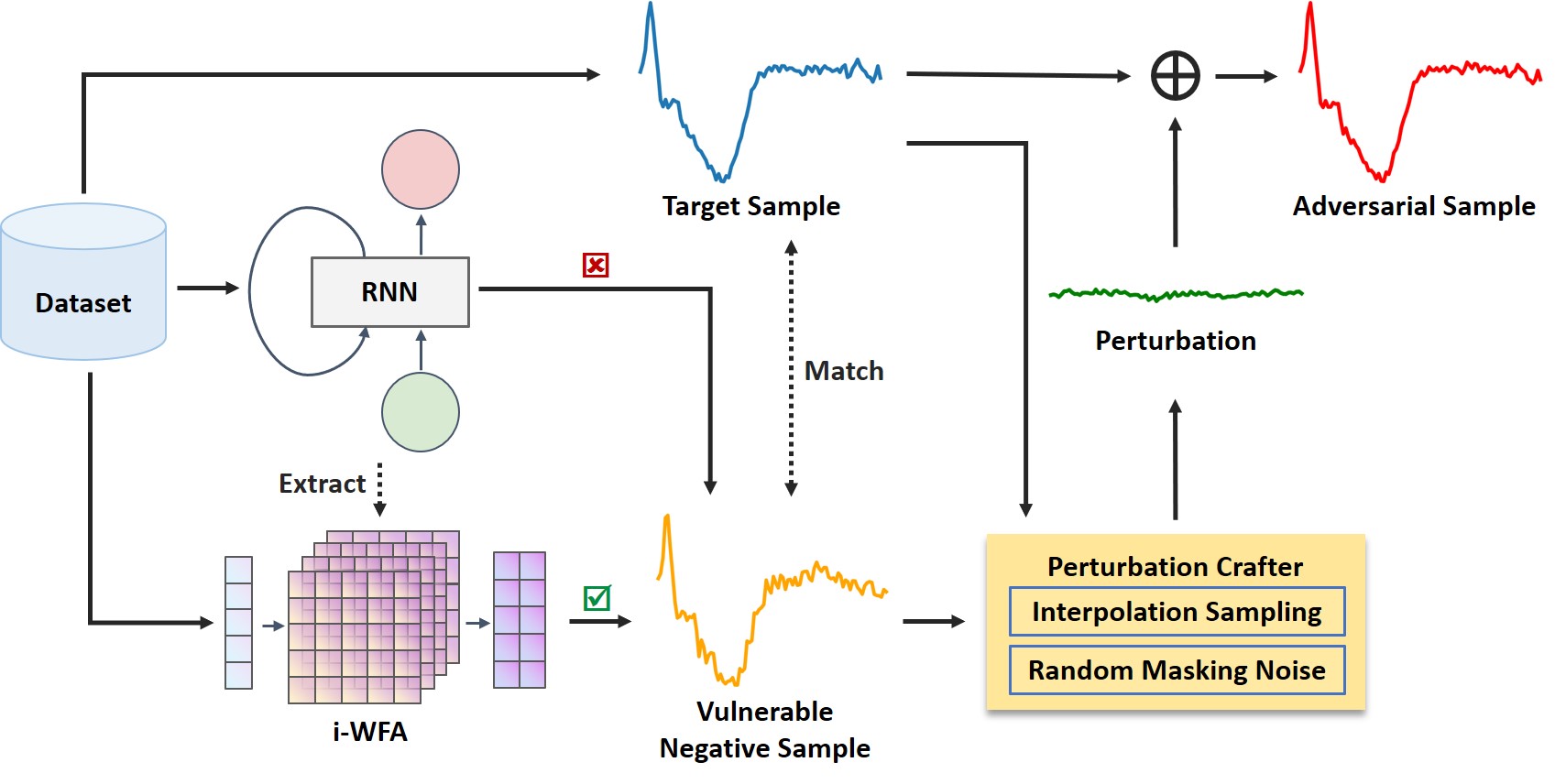}}
\vspace{0.3cm}
\caption{With the representation model i-WFA extracted from the target RNN classifier, the vulnerable samples with features deviating from the latent manifold can be captured according to their predictions. Then the target samples are specifically matched and then perturbed through two steps namely the interpolation sampling and adding random masking noise. \vspace{0.3cm}} \label{fig:map}
\end{figure}

Based on the ideas and methodology above, we propose an efficient approach named TSFool to craft highly-imperceptible adversarial time series for RNN-based TSC. The roadmap of TSFool is shown in Figure~\ref{fig:map}.

\subsection{Extraction of Representation Model}\label{subsec:iwfa}

To model the special recurrent computation of RNN, we introduce \textit{weighted finite automaton} (WFA) \citep{zhang2021decision} that can imitate the execution of RNN based on its hidden state updated at each time step. Nevertheless, WFA also relies on existing clustering methods like $k \text{-} means$ to abstract input data, while an important requirement of the representation model is to distinguish different inputs directly by original features as a human. So we improve WFA by changing its input from data clusters to domain intervals. Specifically, to determine the interval size, for each of the input samples, we calculate the average $\ell_2$ distance between its features in adjacent time steps, and then reduce the result by an order of magnitude as an ``imperceptible distance''. In this way, when the size of the input interval is smaller than this distance, it is almost impossible for features that belong to different input clusters in the original WFA to be assigned to the same interval, so as to ensure that the domain interval is a reasonable substitute of the input cluster. Accordingly, we propose an ideal representation model as below. 

An \textbf{Intervalized Weighted Finite Automaton (i-WFA)} extracted from a $k$-class RNN classifier $\mathcal{N}$ is defined as a tuple $\mathcal{A}=(Z,S,\mathcal{I},(E_{\zeta})_{\zeta \in Z},\mathcal{T})$, where $Z$ is a finite set of intervals that covers the whole input domain, $S$ is a finite set of states abstracted from the hidden states of $\mathcal{N}$, $\mathcal{I}$ is the initial state vector of dimension $|S|$, and $\mathcal{T}$ is the probabilistic output matrix with size $|S| \times k$. For each interval $\zeta \in Z$, the corresponding probabilistic transfer matrix $E_{\zeta}$ with size $|S| \times |S|$ is built according to the hidden transfer of all the samples having features that fall into this interval. In the execution of i-WFA, the $\mathcal{I}$ is iteratively updated to represent the current state with the transfer under different inputs imitated by the corresponding $E_{\zeta}$ at each time step, followed by the $\mathcal{T}$ to imitate the computation of probabilistic predictions finally. The detailed algorithm of i-WFA establishment and an instance of i-WFA are respectively provided in Section~\ref{subsec:sm_iwfa_estab} and Section~\ref{subsec:sm_iwfa_instance} of the \textbf{Appendix} in our Supplementary Material.

\subsection{Craft Imperceptible Adversarial Time Series}

\paragraph{Capturing Vulnerable Negative Sample}
Building an i-WFA as the representation model of the target RNN classifier, we make a comparison between them to capture the VNS. As shown in Algorithm~\ref{alg:TSFool}, we get the prediction results in the test set respectively from the RNN classifier and i-WFA (lines 3-4), followed by the comparison between them in function \textit{CaptureVNS} (line 5).

\paragraph{Interpolation Sampling}\label{subsubsec:perturb1}
As argued in Section~\ref{sec:motiv}, different from the existing methods that attack all the test samples, TSFool picks specific TPS according to the VNS captured. Given that a pair of VNS and TPS are similar in features and with the same label, while they are predicted differently by the classifier, there must be a hyperplane to be discovered between them that divides the latent manifold of the two classes. So we can approximate that hyperplane by the interpolation of their features as the first part of the perturbation (i.e. $\lambda_\varepsilon \Vert \Vec{x}_v - \Vec{x} \Vert$). In Algorithm~\ref{alg:TSFool}, we pick TPS for each of the VNS by the function \textit{PickTPS} (lines 6-7). Then with the function \textit{UpdateSamplingRange} which updates the sampling range each turn by the two currently sampled adjacent examples with different prediction results respectively the same as VNS and TPS (lines 10-11), we do average sampling (line 9) to approximate the maximum feature interpolation iteratively until the step size of sampling is smaller than a given limitation of noise (lines 12-13).

\begin{algorithm}[htb]
\caption{Craft Imperceptible Adversarial Time Series}
\label{alg:TSFool}
\textbf{Input}: RNN $\mathcal{N}=(X,Y,H,f,g)$, i-WFA $\mathcal{A}=(Z,S,\mathcal{I},$ $(E_{\zeta})_{\zeta \in Z},\mathcal{T})$, Test Set $\mathcal{X}$, Test Labels $\mathcal{Y}$, Hyper-parameters $\varepsilon$, $n$ and $p$\\
\textbf{Output}: Adversarial Time Series Set $\mathcal{X}_{\textit{adv}}$

\begin{algorithmic}[1] 
\State Initialize $\mathcal{X}_{adv} \leftarrow []$
\State $\epsilon \leftarrow \textit{ImperceptibleNoise}(\varepsilon)$
\State $Y_{\mathcal{N}} \leftarrow \mathcal{N}(\mathcal{X})$ 
\State $Y_{\mathcal{A}} \leftarrow \mathcal{A}(\mathcal{X})$
\State $X_{v} \leftarrow \textit{CaptureVNS}(\mathcal{X}, \mathcal{Y},Y_{\mathcal{N}},Y_{\mathcal{A}})$
\While{$x_{\textit{neg}} \in X_{v}$}
\State $x_{\textit{pos}} \leftarrow \textit{PickTPS}(\mathcal{X},x_{\textit{neg}})$
\While{$x_{m} \textbf{ not exist}$}
\State $X_{s} \leftarrow \textit{InterpolationSampling}(x_{\textit{neg}},x_{\textit{pos}})$
\State $Y_{s} \leftarrow \mathcal{N}(X_{s})$
\State $x_{\textit{neg}},x_{\textit{pos}} \leftarrow \textit{UpdateSamplingRange}(X_{s},Y_{s})$
\If{$\Vert x_{\textit{pos}} - x_{\textit{neg}} \Vert < \epsilon$}
\State $x_{m} \leftarrow x_{\textit{pos}}$
\EndIf
\EndWhile
\State $X_{\textit{adv}} \leftarrow  \textit{AddRandomMaskingNoise}(x_{m},\epsilon,n,p)$
\State $\mathcal{X}_{\textit{adv}}.\textit{append}(X_{adv})$
\EndWhile
\State \Return $\mathcal{X}_{\textit{adv}}$
\end{algorithmic}
\end{algorithm}

\begin{figure}[htb]
    \centering
    \includegraphics[width=4.9cm]{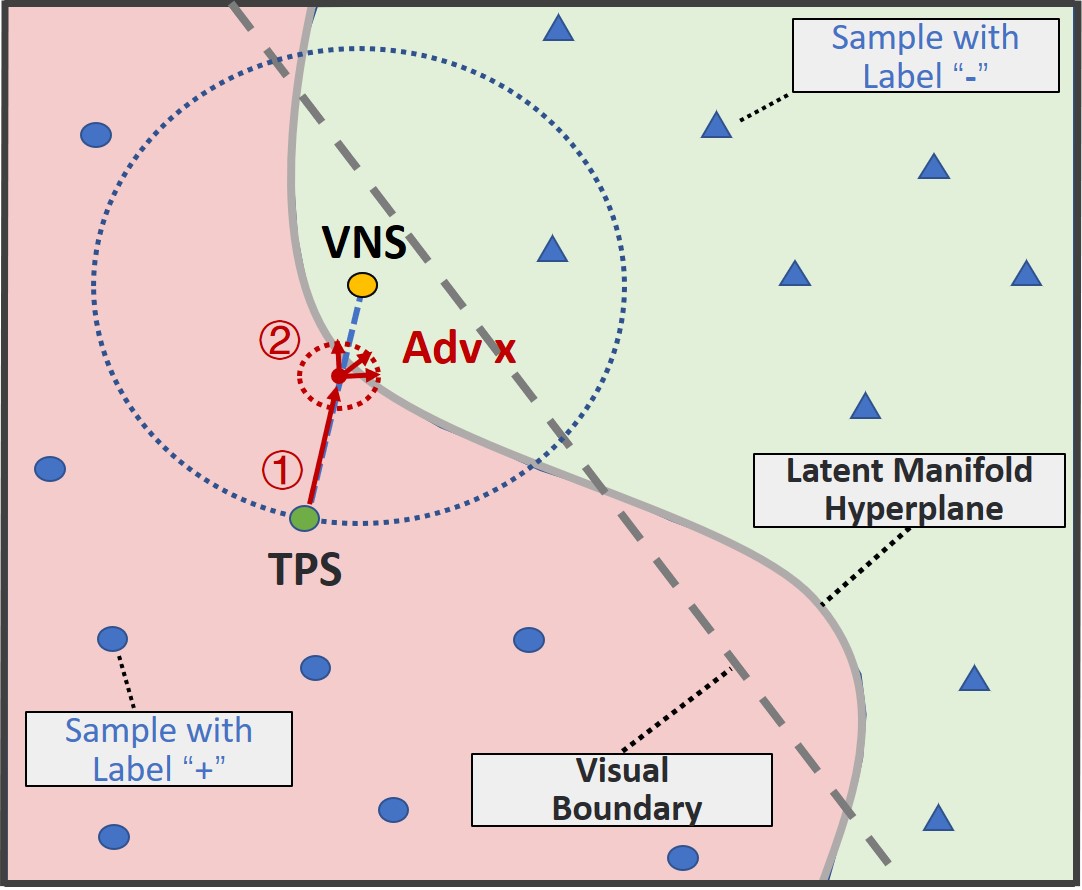}
    \vspace{0.3cm}
    \caption{An intuitive illustration for the crafting of highly-imperceptible adversarial time series. The \textcircled{1} and \textcircled{2} correspond to the two parts of perturbation, respectively from the interpolation sampling and random masking noise. \vspace{0.8cm}}
    \label{fig:explaination3}
\end{figure}

\paragraph{Adding Random Masking Noise}\label{subsubsce:perturb2}
Since the first part of perturbation already approaches the hyperplane of the latent manifold at a given step size, a micro noise $\Vec{x}_\varepsilon$ of the same size can be added to easily cross the hyperplane. In other words, the adversarial property of the perturbed samples is guaranteed by the first part of perturbation, and the only requirement for $\Vec{x}_\varepsilon$ is that it must be micro enough that the semantical consistency between VNS and TPS can also be inherited by the perturbed sample. We implement random masking to generate $\Vec{x}_\varepsilon$ in batches, and add it to finish the attack (line 16). Specifically, we first build a complete noise vector in the direction of interpolation, with the specific noise amount $\epsilon$ calculated by \textit{ImperceptibleNoise} under the constraint mentioned before. Then we randomly mask elements at some time steps of the noise vector (i.e., set value 0) according to a given probability $p$. For each pair of VNS and TPS, this process is run $n$ times to craft a batch of adversarial samples, also with size $n$.

\section{Evaluation}\label{sec:evalu}

\begin{table*}[htbp]
\begin{center}
\renewcommand{\arraystretch}{0.95}
\resizebox{16.8cm}{!}{
\begin{tabular}{c|c|c|c|c|c}
\toprule
\multirow{2}{*}{\textbf{Method}} & \multirow{1.2}{*}{\textbf{Attack}} & \multirow{1.2}{*}{\textbf{Generation}} & \multirow{1.2}{*}{\textbf{Average}} & \multirow{1.2}{*}{\textbf{Perturbation}} & \multirow{1.2}{*}{\textbf{Camouflage}} \\
& \multirow{0.8}{*}{\textbf{Success Rate}} & \multirow{0.8}{*}{\textbf{Number}} & \multirow{0.8}{*}{\textbf{Time Cost (s)}} & \multirow{0.8}{*}{\textbf{Ratio ($\rho^*$)}} & \multirow{0.8}{*}{\textbf{Coefficient}} \\
\midrule
\midrule
FGSM & \hspace{1.8em} 72.12\% \hspace{1.8em} & \hspace{1.8em} \multirow{8}{*}{300.75} \hspace{1.8em} & \hspace{1.8em} \cellcolor{Melon}\underline{0.0018} \hspace{1.8em} & \hspace{1.8em} 37.13\% \hspace{1.8em} & \hspace{1.8em} 1.0804 \hspace{1.8em} \\
JSMA & 83.53\% & & 1.0287 & 15.06\% & 0.9476 \\
DeepFool & 81.58\% & & 0.0276 & 21.45\% & 1.0107 \\
PGD (BIM) & 76.84\% & & 0.1327 & 22.71\% & 0.9938 \\
C\&W & 69.90\% & & 3.2016 & 5.16\% & 0.9372 \\
Auto-Attack & 80.11\% & & 0.1824 & 22.55\% & 0.9745 \\
\hspace{1em} Boundary Attack \hspace{1em} & 79.01\% & & 9.0399 & \cellcolor{Melon}\underline{3.04\%} & 0.8788 \\
HopSkipJump & 83.17\% & & 12.3068 & \cellcolor{Melon}3.86\% & 0.8872 \\
Transfer Attack & 19.54\% & 250 & - & 7.68\% & 1.2010 \\
\textbf{TSFool} & \cellcolor{SpringGreen}\underline{\textbf{87.76\%}} & 305 & \textbf{0.0230} & \textbf{4.63\%} & \cellcolor{SpringGreen}\underline{\textbf{0.8147}} \\
\bottomrule
\end{tabular}
}
\end{center}
\caption{The table shows the average performance of the experimental methods on the 10 UCR univariate time series datasets, including the ASR, the number of samples generated, the time cost and the two measures $\rho^*$ and $\mathcal{C}$ for imperceptibility. TSFool realizes state-of-the-art performance in ASR and the proposed CC. For local perturbation, TSFool is slightly behind Boundary Attack and HopSkipJump, while they are two to three orders slower than TSFool. TSFool also outperforms all the benchmarks in efficiency except FGSM, which is not surprising because FGSM is a simple single-step method without satisfying results under other measures. \vspace{0.6cm}}
\label{tab:average}
\end{table*}

In this section, with 10 univariate and one multivariate time series datasets respectively derived from the public UCR \citep{dau2019ucr} and UEA \citep{bagnall2018uea} time series archives, we evaluate the effectiveness, efficiency and imperceptibility of TSFool. To illustrate its advantages, there are six white-box attacks and three black-box attacks from basic to state-of-the-art that serve as the benchmarks. The detailed experimental setup can be found in Section~\ref{subsec:sm_setup}. Besides, the Python code of TSFool, the pre-trained RNN classifiers, and the raw data of our experiments including the crafted adversarial sets are publicly available in our GitHub repository.\footnote{\vspace{0.1cm}\url{https://github.com/FlaAI/TSFool}\label{fn:1}}

\subsection{Adopted Measures}\label{subsec:measure}

There are mainly four measures adopted for the evaluation. Firstly, we report the original accuracy of the target classifiers and the Attack Success Rate (ASR) to evaluate the effectiveness of attacks. Secondly, we record the average time for crafting a single adversarial sample as the measure of efficiency. Finally, the imperceptibility is considered from two perspectives namely the global Camouflage Coefficient Equation~(\ref{eq:cc}) and the local perturbation. Although the commonly used measure for the latter is the perturbation ratio:
\begin{equation}\label{eq:traditionmetric}
    \rho = \frac{\hspace{0.1em} \Vert \delta_{\Vec{x}} \Vert \hspace{0.1em}}{\Vert \Vec{x} \Vert},
\end{equation}
we propose a variant $\rho^*$ named \textbf{Domain Perturbation Ratio} replacing the original denominator by the specific input domain. This is because typical time series usually have features with large values but narrow distribution ranges, so the input domain can be a better denominator than the absolute value. Given that $\mathcal{X}^{(i)}$ denotes the set of feature values at time step $i$ of all the samples in $\mathcal{X}$, the $\rho^*$ is defined as:
\begin{equation}\label{eq:rhostar}
    \rho^* = \frac{\Vert \delta_{\Vec{x}} \Vert}{ \hspace{0.1em} \sum_{i = 1}^{|\Vec{x}|} \Vert max(\mathcal{X}^{(i)}) - min(\mathcal{X}^{(i)}) \Vert \hspace{0.1em} },
\end{equation}
which always provides similar information as $\rho$ but reflects the relative size of perturbation better for time series data.

\subsection{Overview of Experimental Results}\label{subsec:result1}

\begin{figure*}
  \centering
  \begin{subfigure}{0.196\linewidth}
    \includegraphics[width=1.04\linewidth]{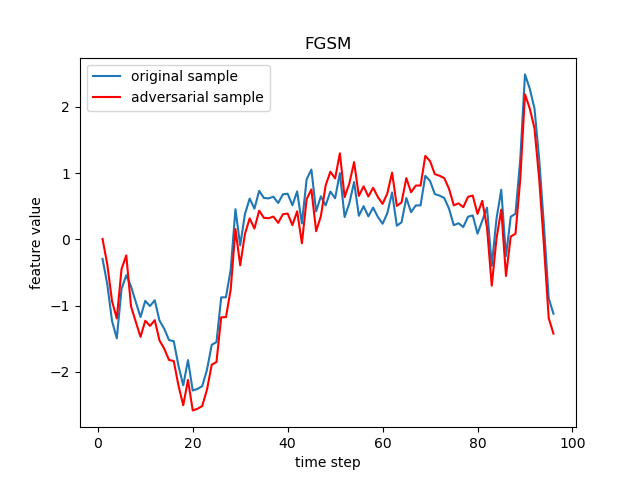}
  \end{subfigure}
  \begin{subfigure}{0.196\linewidth}
    \includegraphics[width=1.04\linewidth]{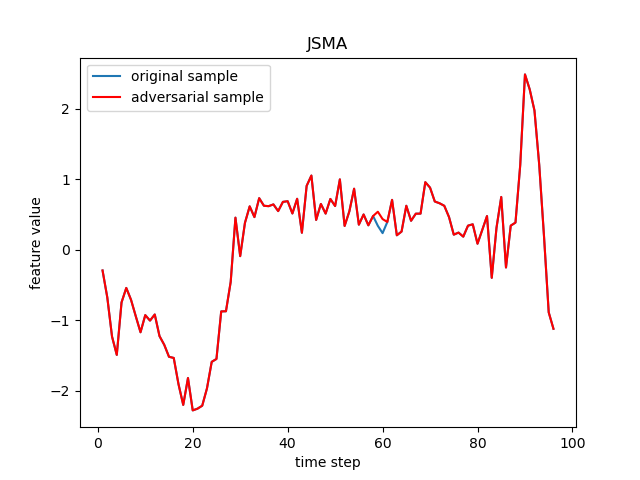}
  \end{subfigure}
  \begin{subfigure}{0.196\linewidth}
    \includegraphics[width=1.04\linewidth]{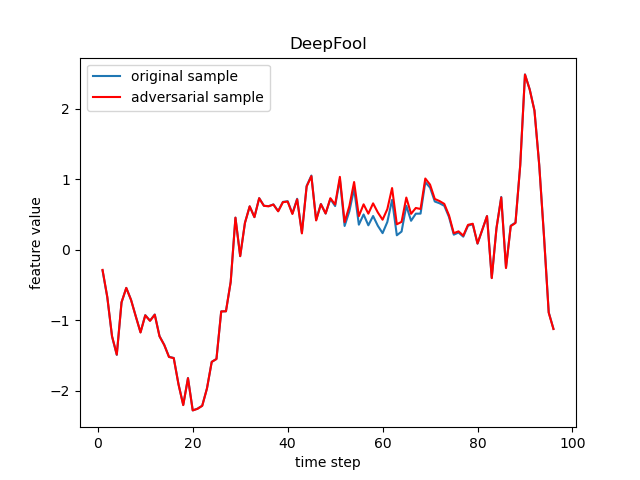}
  \end{subfigure}
  \begin{subfigure}{0.196\linewidth}
    \includegraphics[width=1.04\linewidth]{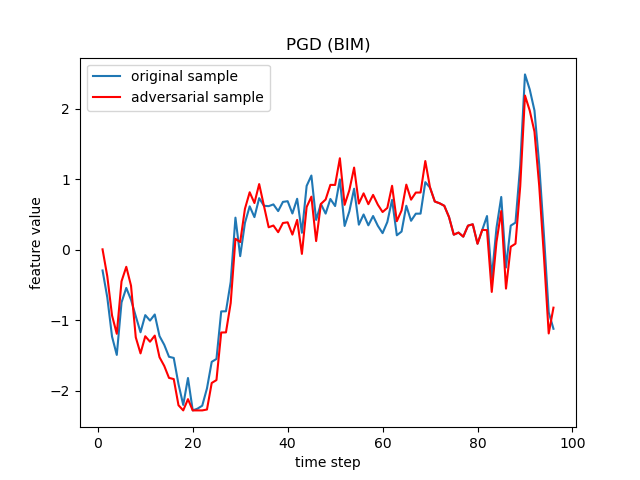}
  \end{subfigure}
  \begin{subfigure}{0.196\linewidth}
    \includegraphics[width=1.04\linewidth]{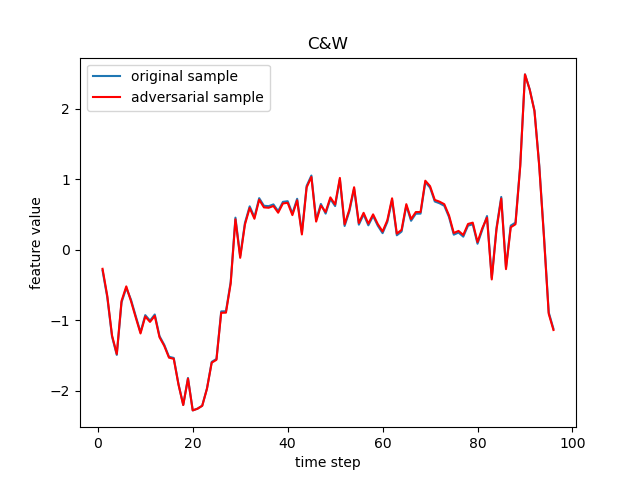}
  \end{subfigure}
  \centering
  \begin{subfigure}{0.196\linewidth}
    \includegraphics[width=1.04\linewidth]{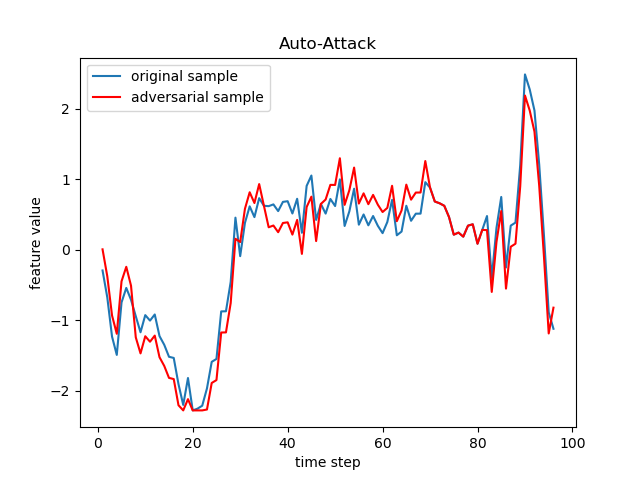}
  \end{subfigure}
  \begin{subfigure}{0.196\linewidth}
    \includegraphics[width=1.04\linewidth]{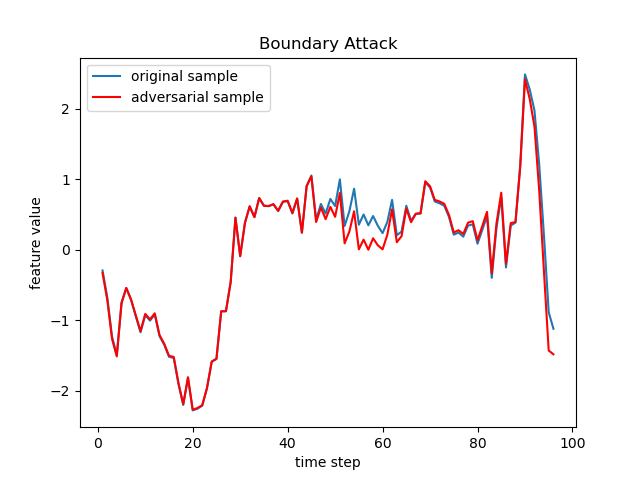}
  \end{subfigure}
  \begin{subfigure}{0.196\linewidth}
    \includegraphics[width=1.04\linewidth]{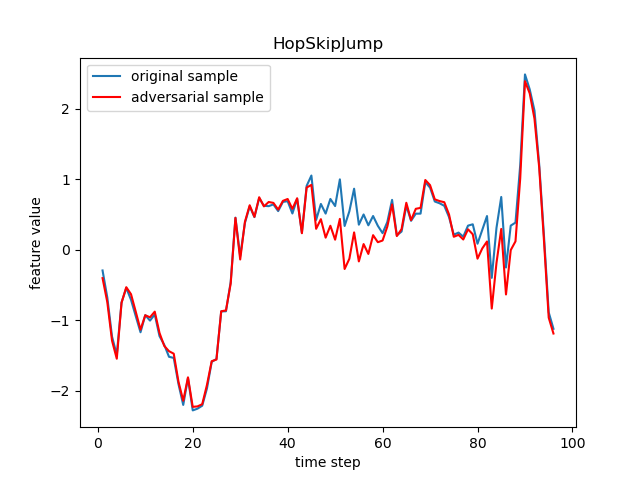}
  \end{subfigure}
  \begin{subfigure}{0.196\linewidth}
    \includegraphics[width=1.04\linewidth]{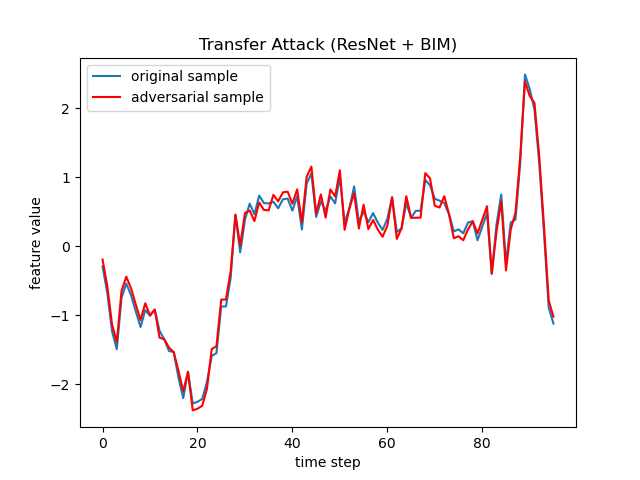}
  \end{subfigure}
  \begin{subfigure}{0.196\linewidth}
    \includegraphics[width=1.04\linewidth]{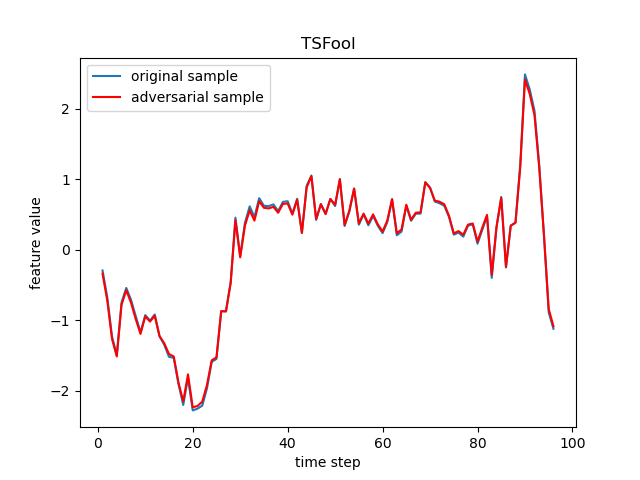}
  \end{subfigure}
  \vspace{0.1cm}
  \caption{The figures show an instance of adversarial attacks through the nine benchmark methods and TSFool on UCR-ECG200 dataset. The results of FGSM, PGD and Auto-Attack are unsatisfying with obvious distortion. While JSMA and DeepFool perform better in this case, their average performances in Table~\ref{tab:average} are just mediocre. On the contrary, Boundary Attack and HopSkipJump are shown more successful in local perturbation control, but this instance reveals that sometimes they can be unstable, not to mention their worst efficiency. Only C\&W, Transfer Attack and TSFool basically realize imperceptible perturbation here, while it should be remembered that on average, C\&W is around two orders slower than TSFool and Transfer Attack achieves the worst ASR among all the experimental methods. \vspace{0.6cm}}
  \label{fig:compare}
\end{figure*}

The average performances (\ie \textit{modified mean} \citep{2009statistics}) of TSFool and the benchmark methods on 10 UCR univariate time series datasets are shown in Table~\ref{tab:average}. For \textbf{\textit{effectiveness}}, the average ASR of TSFool is significantly higher than the benchmarks, with their gaps from 4.23\% to 68.22\%. The \textbf{\textit{efficiency}} of TSFool and DeepFool are at the same level, behind FGSM but better than the other seven benchmarks by one to three orders of magnitude. As FGSM is a basic method not outstanding in other measures, we can state that TSFool is efficient enough. For \textbf{\textit{imperceptibility}}, TSFool not only performs the best under the proposed CC, but also beats seven in nine benchmarks under the conventional local perturbation, with the rest two to three orders slower. This confirms the imperceptibility of TSFool is impressive from the global perspective, and also considerable from local. An intuitive example of attack results by TSFool and the benchmarks is given in Figure~\ref{fig:compare}.

Due to the limited space, this section is just a brief overview of our experimental results, and we leave the detailed analysis in Section~\ref{subsec:sm_result1} of the \textbf{Appendix} in our \textbf{Supplementary Material}. There are also a number of additional experiments that support our points and strengthen the confidence of the above results, including \uppercase\expandafter{\romannumeral1}) implementing TSFool on a multivariate time series dataset from the UEA archive in Section~\ref{subsec:sm_result2}, where it achieves similar performance to the univariate cases; \uppercase\expandafter{\romannumeral2}) exploring the impact of hyper-parameters for all the experimental methods in Section~\ref{subsec:sm_addExp}, to confirm the above results are of general significance; \uppercase\expandafter{\romannumeral3}) additionally evaluating the benchmarks on TPSs in Section~\ref{subsec:sm_fair}, to dispel a possible concern about the consistency of the final compared data and further confirm the fairness of our experiments; \uppercase\expandafter{\romannumeral4}) conducting subjective human studies with 65 volunteers in Section~\ref{subsec:sm_human_study}, to illustrate the benefit of using Camouflage Coefficient in representing real-world imperceptibility of adversarial samples; and \uppercase\expandafter{\romannumeral5}) evaluating TSFool by four common anomaly detection methods in Section~\ref{subsec:sm_anomaly}, to show its advantages beyond existing attacks in imperceptibility under the challenge of real-world defense.

\section{Discussion}\label{sec:discu}

While the surprising results in Section~\ref{sec:evalu} are revealed by RNN-based TSC specifically, they are enough to reflect the fact that general consideration beyond image data and feed-forward models is still lacking in the existing knowledge. As a consequence, we believe the current theory of adversarial attack is incomplete and needs to be further refined. Another problem to be noticed is that imperceptibility should have been one of the most important measures of an adversarial sample as this is part of its definition \citep{szegedy2014intriguing}, without which it should not be believed really ``adversarial''. However, imperceptibility has not received sufficient attention to date, at least not at the same level as ASR. So constructing widely recognized measures for the imperceptibility of adversarial samples is still an important and promising direction. We hope some preliminary ideas regarding these two problems in this paper could raise the community's attention and be instructive for future research.

There are a few related works making preliminary explorations in these directions. For instance, \citet{belkhouja2022dynamic} provide theoretical and empirical evidence to demonstrate the effectiveness of \textit{dynamic time warping} (DTW) over the standard \textit{Euclidean} distance metric regarding the robustness of NNs for time series domain, which also emphasizes the significance of new measures in this field. We provide further discussion in Section~\ref{subsec:sm_dtw} and showcase that, by using DTW as a measure of distortion, TSFool still outperforms PGD. \citet{zhang2020geometry} argue that considering the distance to class boundary, adversarial samples should have unequal importance and should be assigned with different weights, which also lays a solid foundation for our idea to pick TPS in Section~\ref{subsec:reconsider}. But still, we understand someone may view this as a weakness of TSFool as this violates the conventional setting in the evaluation of attack methods to exactly generate an adversarial sample for every single benign one. Fortunately, picking TPS is vital but not indispensable for TSFool. We further discuss the details in Section~\ref{subsec:sm_extend}, and provide an extended version of TSFool without the above weakness. Additional experiments there confirm that the extended TSFool is still the most competitive choice compared with the benchmarks.


Although as aforementioned, our contribution is not limited to proposing the TSFool approach, someone may still doubt whether it is indeed motivated enough to specifically design for RNN-based TSC, as more state-of-the-art solutions for TSC tasks are based on convolutional NNs or transformers \citep{cheng2023classification}. Nevertheless, the fact is that to date RNN-based TSC applications are still popular in real-world practice \citep{ding2022towards, wu2022small, ding2023black}, without an effective approach to measuring their robustness \citep{galib2023susceptibility, ding2023black}, which leaves potential threats to the public. TSFool can be viewed as a ``gray-box'' method. In short, considering the impact of specific modes of RNN running to i-WFA extraction in different applications, TSFool may either be implemented in a black-box way, or rely on a part of white-box information. We leave a more detailed explanation in Section~\ref{subsec:sm_gray}. Please notice that this is just about the property of TSFool, instead of the background setting of this paper. Another point to be noticed is that as TSFool relies on existing vulnerable samples wrongly predicted by the target RNN classifier, a natural requirement is that such samples must exist. Generally, this should not be a serious concern, as they can be any real-world sample at inference time instead of just from the supervised dataset. In practice, TSFool provides several hyper-parameters for fine-tuning as shown in Section~\ref{subsec:sm_addExp} and supports both targeted and untargeted attacks, which makes it widely applicable. We also showcase its potential for adversarial training in Section~\ref{subsec:sm_at}.



\section{Conclusion}\label{sec:concl}

In this paper, given the lack of research and applicable approach in the field of adversarial samples for RNN-based TSC, we propose the TSFool attack, significantly outperforming existing methods in effectiveness, efficiency and imperceptibility. What's important, the novel global optimization objective "Camouflage Coefficient" proposed to refine the adversarial attack as a multi-objective optimization problem may be instructive for the improvement of the current theory, and the methodology proposed based on latent manifold to heuristically approximate the solution of such a new optimization problem can also be easily transferred to other types of models and data, providing a new feasible way to craft imperceptible adversarial samples. For future works, further exploring the newly defined multi-objective optimization problem to find better approximation solutions is an interesting topic, and the attempt to realize our methodology in other kinds of real-world tasks is in progress at present.






\bibliography{mybibfile}

\begin{thebibliography}{59}
\providecommand{\natexlab}[1]{#1}
\providecommand{\url}[1]{\texttt{#1}}
\expandafter\ifx\csname urlstyle\endcsname\relax
  \providecommand{\doi}[1]{doi: #1}\else
  \providecommand{\doi}{doi: \begingroup \urlstyle{rm}\Url}\fi

\bibitem[Andriushchenko et~al.(2020)Andriushchenko, Croce, Flammarion, and Hein]{andriushchenko2020square}
M.~Andriushchenko, F.~Croce, N.~Flammarion, and M.~Hein.
\newblock Square attack: A query-efficient black-box adversarial attack via random search.
\newblock In \emph{European Conference on Computer Vision (ECCV)}, 2020.

\bibitem[Arulmozhi(2009)]{2009statistics}
G.~Arulmozhi.
\newblock \emph{Statistics For Management, 2nd Edition}.
\newblock Tata McGraw-Hill Education, 2009.
\newblock ISBN 9780070153684.

\bibitem[Athalye et~al.(2018)Athalye, Carlini, and Wagner]{athalye2018obfuscated}
A.~Athalye, N.~Carlini, and D.~Wagner.
\newblock Obfuscated gradients give a false sense of security: Circumventing defenses to adversarial examples.
\newblock In \emph{International Conference on Machine Learning (ICML)}, 2018.

\bibitem[Bagnall et~al.(2018)Bagnall, Dau, Lines, Flynn, Large, Bostrom, Southam, and Keogh]{bagnall2018uea}
A.~Bagnall, H.~A. Dau, J.~Lines, M.~Flynn, J.~Large, A.~Bostrom, P.~Southam, and E.~Keogh.
\newblock The uea multivariate time series classification archive.
\newblock \emph{arXiv preprint arXiv:1811.00075}, 2018.

\bibitem[Belkhouja et~al.(2022)Belkhouja, Yan, and Doppa]{belkhouja2022dynamic}
T.~Belkhouja, Y.~Yan, and J.~R. Doppa.
\newblock Dynamic time warping based adversarial framework for time-series domain.
\newblock \emph{IEEE Transactions on Pattern Analysis and Machine Intelligence (TPAMI)}, 2022.

\bibitem[Biggio et~al.(2013)Biggio, Corona, Maiorca, Nelson, {\v{S}}rndi{\'c}, Laskov, Giacinto, and Roli]{biggio2013evasion}
B.~Biggio, I.~Corona, D.~Maiorca, B.~Nelson, N.~{\v{S}}rndi{\'c}, P.~Laskov, G.~Giacinto, and F.~Roli.
\newblock Evasion attacks against machine learning at test time.
\newblock In \emph{European Conference on Machine Learning and Principles and Practice of Knowledge Discovery in Databases (ECML-PKDD)}. Springer, 2013.

\bibitem[Bl{\'a}zquez-Garc{\'\i}a et~al.(2021)Bl{\'a}zquez-Garc{\'\i}a, Conde, Mori, and Lozano]{blazquez2021review}
A.~Bl{\'a}zquez-Garc{\'\i}a, A.~Conde, U.~Mori, and J.~A. Lozano.
\newblock A review on outlier/anomaly detection in time series data.
\newblock \emph{ACM Computing Surveys (CSUR)}, 2021.

\bibitem[Brendel et~al.(2018)Brendel, Rauber, and Bethge]{brendel2018decision}
W.~Brendel, J.~Rauber, and M.~Bethge.
\newblock Decision-based adversarial attacks: Reliable attacks against black-box machine learning models.
\newblock In \emph{International Conference on Learning Representations (ICLR)}, 2018.

\bibitem[Breunig et~al.(2000)Breunig, Kriegel, Ng, and Sander]{breunig2000lof}
M.~M. Breunig, H.-P. Kriegel, R.~T. Ng, and J.~Sander.
\newblock Lof: Identifying density-based local outliers.
\newblock In \emph{International Conference on Management of Data (SIGMOD)}, 2000.

\bibitem[Campos et~al.(2018)Campos, Jou, Gir{\'o}-i Nieto, Torres, and Chang]{campos2018skip}
V.~Campos, B.~Jou, X.~Gir{\'o}-i Nieto, J.~Torres, and S.-F. Chang.
\newblock Skip rnn: Learning to skip state updates in recurrent neural networks.
\newblock In \emph{International Conference on Learning Representations (ICLR)}, 2018.

\bibitem[Carlini and Wagner(2017)]{carlini2017towards}
N.~Carlini and D.~Wagner.
\newblock Towards evaluating the robustness of neural networks.
\newblock In \emph{IEEE Symposium on Security and Privacy (S\&P)}, 2017.

\bibitem[Chambers and Yoder(2020)]{chambers2020filternet}
R.~D. Chambers and N.~C. Yoder.
\newblock Filternet: A many-to-many deep learning architecture for time series classification.
\newblock \emph{Sensors}, 2020.

\bibitem[Chen et~al.(2020)Chen, Jordan, and Wainwright]{chen2020hopskipjumpattack}
J.~Chen, M.~I. Jordan, and M.~J. Wainwright.
\newblock Hopskipjumpattack: A query-efficient decision-based attack.
\newblock In \emph{IEEE Symposium on Security and Privacy (S\&P)}, 2020.

\bibitem[Cheng et~al.(2023)Cheng, Khalitov, Yu, Zhang, and Yang]{cheng2023classification}
L.~Cheng, R.~Khalitov, T.~Yu, J.~Zhang, and Z.~Yang.
\newblock Classification of long sequential data using circular dilated convolutional neural networks.
\newblock \emph{Neurocomputing}, 2023.

\bibitem[Chollet(2021)]{chollet2021deep}
F.~Chollet.
\newblock \emph{Deep learning with Python}.
\newblock Simon and Schuster, 2021.

\bibitem[Croce and Hein(2020)]{croce2020reliable}
F.~Croce and M.~Hein.
\newblock Reliable evaluation of adversarial robustness with an ensemble of diverse parameter-free attacks.
\newblock In \emph{International Conference on Machine Learning (ICML)}, 2020.

\bibitem[Danskin(2012)]{danskin2012theory}
J.~M. Danskin.
\newblock \emph{The theory of max-min and its application to weapons allocation problems}.
\newblock Springer Science \& Business Media, 2012.

\bibitem[Dau et~al.(2019)Dau, Bagnall, Kamgar, Yeh, Zhu, Gharghabi, Ratanamahatana, and Keogh]{dau2019ucr}
H.~A. Dau, A.~Bagnall, K.~Kamgar, C.-C.~M. Yeh, Y.~Zhu, S.~Gharghabi, C.~A. Ratanamahatana, and E.~Keogh.
\newblock The ucr time series archive.
\newblock \emph{IEEE/CAA Journal of Automatica Sinica}, 2019.

\bibitem[Deng et~al.(2009)Deng, Dong, Socher, Li, Li, and Fei-Fei]{deng2009imagenet}
J.~Deng, W.~Dong, R.~Socher, L.-J. Li, K.~Li, and L.~Fei-Fei.
\newblock Imagenet: A large-scale hierarchical image database.
\newblock In \emph{IEEE/CVF Conference on Computer Vision and Pattern Recognition (CVPR)}, 2009.

\bibitem[Ding et~al.(2022)Ding, Zhang, Huang, Pan, Feng, Jiang, and Yang]{ding2022towards}
D.~Ding, M.~Zhang, Y.~Huang, X.~Pan, F.~Feng, E.~Jiang, and M.~Yang.
\newblock Towards backdoor attack on deep learning based time series classification.
\newblock In \emph{International Conference on Data Engineering (ICDE)}. IEEE, 2022.

\bibitem[Ding et~al.(2023)Ding, Zhang, Feng, Huang, Jiang, and Yang]{ding2023black}
D.~Ding, M.~Zhang, F.~Feng, Y.~Huang, E.~Jiang, and M.~Yang.
\newblock Black-box adversarial attack on time series classification.
\newblock In \emph{AAAI Conference on Artificial Intelligence (AAAI)}, 2023.

\bibitem[Ergen and Kozat(2019)]{ergen2019unsupervised}
T.~Ergen and S.~S. Kozat.
\newblock Unsupervised anomaly detection with lstm neural networks.
\newblock \emph{IEEE Transactions on Neural Networks and Learning Systems (TNNLS)}, 2019.

\bibitem[Fawaz et~al.(2019)Fawaz, Forestier, Weber, Idoumghar, and Muller]{fawaz2019adversarial}
H.~I. Fawaz, G.~Forestier, J.~Weber, L.~Idoumghar, and P.-A. Muller.
\newblock Adversarial attacks on deep neural networks for time series classification.
\newblock In \emph{International Joint Conference on Neural Networks (IJCNN)}. IEEE, 2019.

\bibitem[Fefferman et~al.(2016)Fefferman, Mitter, and Narayanan]{fefferman2016testing}
C.~Fefferman, S.~Mitter, and H.~Narayanan.
\newblock Testing the manifold hypothesis.
\newblock \emph{Journal of the American Mathematical Society}, 2016.

\bibitem[Galib and Bashyal(2023)]{galib2023susceptibility}
A.~H. Galib and B.~Bashyal.
\newblock On the susceptibility and robustness of time series models through adversarial attack and defense.
\newblock \emph{arXiv preprint arXiv:2301.03703}, 2023.

\bibitem[Gao et~al.(2018)Gao, Yu, Wu, and Li]{gao2018low}
P.~Gao, L.~Yu, Y.~Wu, and J.~Li.
\newblock Low latency rnn inference with cellular batching.
\newblock In \emph{European Conference on Computer Systems (EuroSys)}, 2018.

\bibitem[Goodfellow et~al.(2015)Goodfellow, Shlens, and Szegedy]{goodfellow2015explaining}
I.~J. Goodfellow, J.~Shlens, and C.~Szegedy.
\newblock Explaining and harnessing adversarial examples.
\newblock In \emph{International Conference on Learning Representations (ICLR)}, 2015.

\bibitem[Huang et~al.(2020)Huang, Kroening, Ruan, Sharp, Sun, Thamo, Wu, and Yi]{huang2020survey}
X.~Huang, D.~Kroening, W.~Ruan, J.~Sharp, Y.~Sun, E.~Thamo, M.~Wu, and X.~Yi.
\newblock A survey of safety and trustworthiness of deep neural networks: Verification, testing, adversarial attack and defence, and interpretability.
\newblock \emph{Computer Science Review}, 2020.

\bibitem[Ismail~Fawaz et~al.(2019)Ismail~Fawaz, Forestier, Weber, Idoumghar, and Muller]{ismail2019deep}
H.~Ismail~Fawaz, G.~Forestier, J.~Weber, L.~Idoumghar, and P.-A. Muller.
\newblock Deep learning for time series classification: a review.
\newblock \emph{Data Mining and Knowledge Discovery (DMKD)}, 2019.

\bibitem[Karim et~al.(2020)Karim, Majumdar, and Darabi]{karim2020adversarial}
F.~Karim, S.~Majumdar, and H.~Darabi.
\newblock Adversarial attacks on time series.
\newblock \emph{IEEE Transactions on Pattern Analysis and Machine Intelligence (TPAMI)}, 2020.

\bibitem[Kolter and Madry(2018)]{kolter2018tutorial}
Z.~Kolter and A.~Madry.
\newblock Tutorial adversarial robustness: Theory and practice.
\newblock In \emph{Advances in Neural Information Processing Systems (NeurIPS)}. 2018.

\bibitem[Kotyan and Vargas(2019)]{kotyan2019adversarial}
S.~Kotyan and D.~V. Vargas.
\newblock Adversarial robustness assessment: Why both $l_0$ and $l_{\infty}$ attacks are necessary.
\newblock \emph{arXiv preprint arXiv:1906.06026}, 2019.

\bibitem[Kurakin et~al.(2017)Kurakin, Goodfellow, and Bengio]{kurakin2017adversarial}
A.~Kurakin, I.~Goodfellow, and S.~Bengio.
\newblock Adversarial examples in the physical world.
\newblock In \emph{Workshop Track @ International Conference on Learning Representations (ICLR)}, 2017.

\bibitem[Lai et~al.(2021)Lai, Zha, Wang, Xu, Zhao, Kumar, Chen, Zumkhawaka, Wan, Martinez, et~al.]{lai2021tods}
K.-H. Lai, D.~Zha, G.~Wang, J.~Xu, Y.~Zhao, D.~Kumar, Y.~Chen, P.~Zumkhawaka, M.~Wan, D.~Martinez, et~al.
\newblock Tods: An automated time series outlier detection system.
\newblock In \emph{AAAI Conference on Artificial Intelligence (AAAI)}, 2021.

\bibitem[L{\"a}ngkvist et~al.(2014)L{\"a}ngkvist, Karlsson, and Loutfi]{langkvist2014review}
M.~L{\"a}ngkvist, L.~Karlsson, and A.~Loutfi.
\newblock A review of unsupervised feature learning and deep learning for time-series modeling.
\newblock \emph{Pattern Recognition Letters (PRL)}, 2014.

\bibitem[Li et~al.(2022)Li, Cheng, Hsieh, and Lee]{li2022review}
Y.~Li, M.~Cheng, C.-J. Hsieh, and T.~C. Lee.
\newblock A review of adversarial attack and defense for classification methods.
\newblock \emph{The American Statistician}, 2022.

\bibitem[Lin et~al.(2019)Lin, Xu, Wu, Richardson, and Bernal]{lin2019medical}
L.~Lin, B.~Xu, W.~Wu, T.~W. Richardson, and E.~A. Bernal.
\newblock Medical time series classification with hierarchical attention-based temporal convolutional networks: A case study of myotonic dystrophy diagnosis.
\newblock In \emph{CVPR Workshops}, 2019.

\bibitem[Liu et~al.(2012)Liu, Ting, and Zhou]{liu2012isolation}
F.~T. Liu, K.~M. Ting, and Z.-H. Zhou.
\newblock Isolation-based anomaly detection.
\newblock \emph{ACM Transactions on Knowledge Discovery from Data (TKDD)}, 2012.

\bibitem[Madry et~al.(2018)Madry, Makelov, Schmidt, Tsipras, and Vladu]{madry2018towards}
A.~Madry, A.~Makelov, L.~Schmidt, D.~Tsipras, and A.~Vladu.
\newblock Towards deep learning models resistant to adversarial attacks.
\newblock In \emph{International Conference on Learning Representations (ICLR)}, 2018.

\bibitem[Manaswi(2018)]{manaswi2018rnn}
N.~K. Manaswi.
\newblock Rnn and lstm.
\newblock \emph{Deep Learning with Applications Using Python: Chatbots and Face, Object, and Speech Recognition with TensorFlow and Keras}, 2018.

\bibitem[Marcel and Rodriguez(2010)]{marcel2010torchvision}
S.~Marcel and Y.~Rodriguez.
\newblock Torchvision the machine-vision package of torch.
\newblock In \emph{ACM International Conference on Multimedia (ACM MM)}, 2010.

\bibitem[Moosavi-Dezfooli et~al.(2016)Moosavi-Dezfooli, Fawzi, and Frossard]{moosavi2016deepfool}
S.-M. Moosavi-Dezfooli, A.~Fawzi, and P.~Frossard.
\newblock Deepfool: a simple and accurate method to fool deep neural networks.
\newblock In \emph{IEEE/CVF Conference on Computer Vision and Pattern Recognition (CVPR)}, 2016.

\bibitem[Mozer(2013)]{mozer2013focused}
M.~C. Mozer.
\newblock A focused backpropagation algorithm for temporal pattern recognition.
\newblock In \emph{Backpropagation}. Psychology Press, 2013.

\bibitem[Nicolae et~al.(2018)Nicolae, Sinn, Tran, Buesser, Rawat, Wistuba, Zantedeschi, Baracaldo, Chen, Ludwig, et~al.]{nicolae2018adversarial}
M.-I. Nicolae, M.~Sinn, M.~N. Tran, B.~Buesser, A.~Rawat, M.~Wistuba, V.~Zantedeschi, N.~Baracaldo, B.~Chen, H.~Ludwig, et~al.
\newblock Adversarial robustness toolbox v1.0.0.
\newblock \emph{arXiv preprint arXiv:1807.01069}, 2018.

\bibitem[Papernot et~al.(2016{\natexlab{a}})Papernot, McDaniel, and Goodfellow]{papernot2016transferability}
N.~Papernot, P.~McDaniel, and I.~Goodfellow.
\newblock Transferability in machine learning: From phenomena to black-box attacks using adversarial samples.
\newblock \emph{arXiv preprint arXiv:1605.07277}, 2016{\natexlab{a}}.

\bibitem[Papernot et~al.(2016{\natexlab{b}})Papernot, McDaniel, Jha, Fredrikson, Celik, and Swami]{papernot2016limitations}
N.~Papernot, P.~McDaniel, S.~Jha, M.~Fredrikson, Z.~B. Celik, and A.~Swami.
\newblock The limitations of deep learning in adversarial settings.
\newblock In \emph{IEEE European Symposium on Security and Privacy (EuroS\&P)}, 2016{\natexlab{b}}.

\bibitem[Papernot et~al.(2016{\natexlab{c}})Papernot, McDaniel, Swami, and Harang]{papernot2016crafting}
N.~Papernot, P.~McDaniel, A.~Swami, and R.~Harang.
\newblock Crafting adversarial input sequences for recurrent neural networks.
\newblock In \emph{IEEE Military Communications Conference (MILCOM)}, 2016{\natexlab{c}}.

\bibitem[Papernot et~al.(2017)Papernot, McDaniel, Goodfellow, Jha, Celik, and Swami]{papernot2017practical}
N.~Papernot, P.~McDaniel, I.~Goodfellow, S.~Jha, Z.~B. Celik, and A.~Swami.
\newblock Practical black-box attacks against machine learning.
\newblock In \emph{ACM Asia Conference on Computer and Communications Security (AsiaCCS)}, 2017.

\bibitem[Schmidl et~al.(2022)Schmidl, Wenig, and Papenbrock]{schmidl2022anomaly}
S.~Schmidl, P.~Wenig, and T.~Papenbrock.
\newblock Anomaly detection in time series: a comprehensive evaluation.
\newblock \emph{International Conference on Very Large Data Bases (VLDB)}, 2022.

\bibitem[Sch{\"o}lkopf et~al.(2001)Sch{\"o}lkopf, Platt, Shawe-Taylor, Smola, and Williamson]{scholkopf2001estimating}
B.~Sch{\"o}lkopf, J.~C. Platt, J.~Shawe-Taylor, A.~J. Smola, and R.~C. Williamson.
\newblock Estimating the support of a high-dimensional distribution.
\newblock \emph{Neural Computation}, 2001.

\bibitem[Su et~al.(2019)Su, Vargas, and Sakurai]{su2019one}
J.~Su, D.~V. Vargas, and K.~Sakurai.
\newblock One pixel attack for fooling deep neural networks.
\newblock \emph{IEEE Transactions on Evolutionary Computation (TEC)}, 2019.

\bibitem[Szegedy et~al.(2014)Szegedy, Zaremba, Sutskever, Bruna, Erhan, Goodfellow, and Fergus]{szegedy2014intriguing}
C.~Szegedy, W.~Zaremba, I.~Sutskever, J.~Bruna, D.~Erhan, I.~Goodfellow, and R.~Fergus.
\newblock Intriguing properties of neural networks.
\newblock In \emph{International Conference on Learning Representations (ICLR)}, 2014.

\bibitem[Wu et~al.(2022)Wu, Wang, Qiao, Xian, Liu, and Zhang]{wu2022small}
T.~Wu, X.~Wang, S.~Qiao, X.~Xian, Y.~Liu, and L.~Zhang.
\newblock Small perturbations are enough: Adversarial attacks on time series prediction.
\newblock \emph{Information Sciences}, 2022.

\bibitem[Xiang et~al.(2020)Xiang, Xu, Li, Ma, Xuan, and Liu]{xiang2020side}
Y.~Xiang, Y.~Xu, Y.~Li, W.~Ma, Q.~Xuan, and Y.~Liu.
\newblock Side-channel gray-box attack for dnns.
\newblock \emph{IEEE Transactions on Circuits and Systems II: Express Briefs}, 2020.

\bibitem[Zhan et~al.(2018)Zhan, Li, Li, Gu, Habimana, and Wang]{zhan2018stock}
X.~Zhan, Y.~Li, R.~Li, X.~Gu, O.~Habimana, and H.~Wang.
\newblock Stock price prediction using time convolution long short-term memory network.
\newblock In \emph{International Conference on Knowledge Science, Engineering and Management (KSEM)}. Springer, 2018.

\bibitem[Zhang et~al.(2020)Zhang, Zhu, Niu, Han, Sugiyama, and Kankanhalli]{zhang2020geometry}
J.~Zhang, J.~Zhu, G.~Niu, B.~Han, M.~Sugiyama, and M.~Kankanhalli.
\newblock Geometry-aware instance-reweighted adversarial training.
\newblock In \emph{International Conference on Learning Representations (ICLR)}, 2020.

\bibitem[Zhang et~al.(2016)Zhang, Wu, Che, Lin, Memisevic, Salakhutdinov, and Bengio]{zhang2016architectural}
S.~Zhang, Y.~Wu, T.~Che, Z.~Lin, R.~Memisevic, R.~R. Salakhutdinov, and Y.~Bengio.
\newblock Architectural complexity measures of recurrent neural networks.
\newblock \emph{Advances in Neural Information Processing Systems (NeurIPS)}, 2016.

\bibitem[Zhang et~al.(2021{\natexlab{a}})Zhang, Du, Xie, Ma, Liu, and Sun]{zhang2021decision}
X.~Zhang, X.~Du, X.~Xie, L.~Ma, Y.~Liu, and M.~Sun.
\newblock Decision-guided weighted automata extraction from recurrent neural networks.
\newblock In \emph{AAAI Conference on Artificial Intelligence (AAAI)}, 2021{\natexlab{a}}.

\bibitem[Zhang et~al.(2021{\natexlab{b}})Zhang, Zheng, and Mao]{zhang2021adversarial}
X.~Zhang, X.~Zheng, and W.~Mao.
\newblock Adversarial perturbation defense on deep neural networks.
\newblock \emph{ACM Computing Surveys (CSUR)}, 2021{\natexlab{b}}.

\end{thebibliography}


\clearpage
\appendix
\section*{Appendix}
\vspace{0.15cm}

\section{The Representation Model i-WFA}\label{sec:sm_8}

In this section, we supplement more details about the representation model, i-WFA, including the algorithm to establish it in Section~\ref{subsec:sm_iwfa_estab}, an intuitive instance with the corresponding visualization of it in Section~\ref{subsec:sm_iwfa_instance}, and an explanation for the gray-box property of TSFool regarding it in Section~\ref{subsec:sm_gray}.

\subsection{Establishment Process of i-WFA}\label{subsec:sm_iwfa_estab}

The detailed algorithm for the establishment of i-WFA is provided in Algorithm~\ref{alg:wfa}. Firstly, we calculate the difference between features for each pair of adjacent time steps of input samples to get the ``imperceptible distance'' as mentioned in Section~\ref{subsec:iwfa}, which can ensure the effectiveness of the input intervals, and accordingly calculate the corresponding input intervals $Z$ for all of the time steps of the input samples (lines 2-9). Secondly, by running the target RNN classifier on the whole test set, the hidden states of RNN can be extracted from each of the time steps through the output layer (lines 10-13), and are then abstracted according to their top-k predictions and confidences to build the abstract state set $S$ of i-WFA (lines 14-19). Next, the state vector $\mathcal{I}$ is initialized as a one-hot format with the first element set to be 1 (lines 20-21), which indicates that the initial status of i-WFA is ``at the initial state with the probability of 100\%''. Then, for every single abstract input interval $\zeta$, we record the state transitions caused by all the input features falling into it, and transfer the statistics to probabilities to build the corresponding probabilistic transfer matrix $E_{\zeta}$ (lines 22-30). Finally, for every single abstract state in $\mathcal{I}$, we record the model outputs corresponding to the instance state falling into it during the whole process of RNN execution, and also transfer the statistics to build the probabilistic output matrix $\mathcal{T}$ (lines 31-35).

\subsection{An Instance of i-WFA}\label{subsec:sm_iwfa_instance}

In a two-class classification task, given an input time series $X=[x_{1},x_{2},x_{3}]_{(x_{i} \in \zeta_{i})}$ with three time steps and the input $x_i$ of each time step falling into the input intervals $\zeta_{i}$. Given the abstract state set $S=[s_{0},s_{1},s_{2}]$, an initial state vector $\mathcal{I}= \left[ \hspace{0.3em} 1 \hspace{0.7em} 0 \hspace{0.7em} 0 \hspace{0.3em} \right]$, as well as the transfer and output matrixes:
\begin{equation*}
\begin{aligned}
    E_{\zeta_{1}} &= 
        \left[
          \begin{array}{ccc}   
            1/4 & 0 & 3/4 \\
            0 & 0 & 0 \\
            0 & 0 & 0
          \end{array}
        \right], \hspace{0.3em}
    E_{\zeta_{2}} = 
        \left[
          \begin{array}{ccc}   
            1/4 & 1/4 & 1/2 \\
            1 & 0 & 0 \\
            1/2 & 1/2 & 0
          \end{array}
        \right],\\
        E_{\zeta_{3}} &= 
        \left[
          \begin{array}{ccc}   
            0 & 1 & 0 \\
            0 & 1/4 & 3/4 \\
            1/4 & 1/2 & 1/4
          \end{array}
        \right], \hspace{0.3em} and \hspace{0.3em}
    \mathcal{T} = 
        \left[
          \begin{array}{cc}   
            1/2 & 1/2 \\
            0 & 1 \\
            3/4 & 1/4
          \end{array}
        \right],
\end{aligned}
\end{equation*}
then the execution process of this instance i-WFA can be illustrated as follows:
\begin{equation*}
\begin{aligned}
    \mathcal{I} \hspace{0.3em} \cdot & \left( \prod_{i=1}^{|X|} E_{\zeta_{i}} \right) \cdot \mathcal{T} = \mathcal{I} \cdot E_{\zeta_{1}} \cdot E_{\zeta_{2}} \cdot E_{\zeta_{3}} \cdot \mathcal{T}\\
    & \hspace{0.3em} = \left[ \hspace{0.3em} 1 \hspace{0.7em} 0 \hspace{0.7em} 0 \hspace{0.3em} \right] \cdot \left[ \begin{array}{ccc}   
    1/4 & 0 & 3/4 \\
    0 & 0 & 0 \\
    0 & 0 & 0 \end{array} \right] \cdot E_{\zeta_{2}} \cdot E_{\zeta_{3}} \cdot \mathcal{T}\\
    & \hspace{0.3em} = \left[ \hspace{0.3em} 1/4 \hspace{0.7em} 0 \hspace{0.7em} 3/4 \hspace{0.3em} \right] \cdot \left[ \begin{array}{ccc} 
    1/4 & 1/4 & 1/2 \\
    1 & 0 & 0 \\
    1/2 & 1/2 & 0 \end{array} \right] \cdot E_{\zeta_{3}} \cdot \mathcal{T}\\
\end{aligned}
\end{equation*}

\begin{algorithm}[H]
\begin{spacing}{1.1}
\caption{Establishment of i-WFA}
\label{alg:wfa}
\textbf{Input}: RNN $\mathcal{N}=(X,Y,H,f,g)$, Test Set $\mathcal{X}$, Hyper-parameters $K$, $T$ and $F$\\
\textbf{Output}: i-WFA $\mathcal{A}=(Z,S,\mathcal{I},(E_{\zeta})_{\zeta \in Z},\mathcal{T})$ 

\begin{algorithmic}[1] 
\State Initialize $Z \leftarrow []$, $S' \leftarrow []$, $S \leftarrow []$, $E \leftarrow []$, $\mathcal{T} \leftarrow []$
\State $d \leftarrow \textit{ImperceptibleDistance}(\mathcal{X},F)$
\State $\mathcal{X}' \leftarrow \textit{FeaturesNormalization}(\mathcal{X})$
\While{$X' \in \mathcal{X}'$} \Comment{\textit{\footnotesize{for component \uppercase\expandafter{\romannumeral1}: input intervals}}}
\While{$x' \in X'$}
\State $\zeta \leftarrow \textit{floor}(x') \times d$
\State $Z.\textit{add}(\zeta)$
\EndWhile
\EndWhile
\While{$X \in \mathcal{X}$} \Comment{\textit{\footnotesize{for component \uppercase\expandafter{\romannumeral2}: abstract states}}}
\State $S'' \leftarrow [g(f^{(i)}(X))]_{i=0}^{|X|}$
\State $S'.\textit{extend}(S'')$
\EndWhile
\While{$s' \in S'$}
\State $s_{\textit{pred}} \leftarrow \textit{argsort}(-s')[:K]$
\State $s_{\textit{conf}} \leftarrow \textit{floor}((-\textit{sort}(-s'[:K]) \times T)$
\State $s \leftarrow \textit{concat}(s_{\textit{pred}},s_{\textit{conf}})$
\State $S.\textit{add}(s)$
\EndWhile
\State $\mathcal{I} \leftarrow \textit{zeros}(|S|+1)$ \Comment{\textit{\footnotesize{for component \uppercase\expandafter{\romannumeral3}: the state vector}}}
\State $\mathcal{I}[0] \leftarrow 1$
\While{$\zeta \in Z$} \Comment{\textit{\footnotesize{for component \uppercase\expandafter{\romannumeral4}: transfer matrixes}}}
\State $E_{\zeta} \leftarrow \textit{RecordingTransition}^{x' \in \zeta}(\mathcal{X}')$
\While{$E_{\zeta}[i] \in E_{\zeta}$}
\If{$\textit{sum}(E_{\zeta}[i]) \neq 0$}
\State $E_{\zeta}[i] \leftarrow E_{\zeta}[i]/\textit{sum}(E_{\zeta}[i])$
\EndIf
\EndWhile
\State $E.\textit{append}(E_{\zeta})$
\EndWhile
\While{$s \in S$} \Comment{\textit{\footnotesize{for component \uppercase\expandafter{\romannumeral5}: the output matrix}}}
\State $\mathcal{T}_{s} \leftarrow \textit{RecordingPrediction}^{s' \in s}(S')$
\State $\mathcal{T}_{s} \leftarrow \mathcal{T}_{s}/\textit{sum}(\mathcal{T}_{s})$
\State $\mathcal{T}.\textit{append}(\mathcal{T}_{s})$
\EndWhile
\State \Return $\mathcal{A}=(Z,S,\mathcal{I},(E_{\zeta})_{\zeta \in Z},\mathcal{T})$
\end{algorithmic}
\end{spacing}
\end{algorithm}

\begin{figure}[H]
    \centering
    \includegraphics[width=5.5cm]{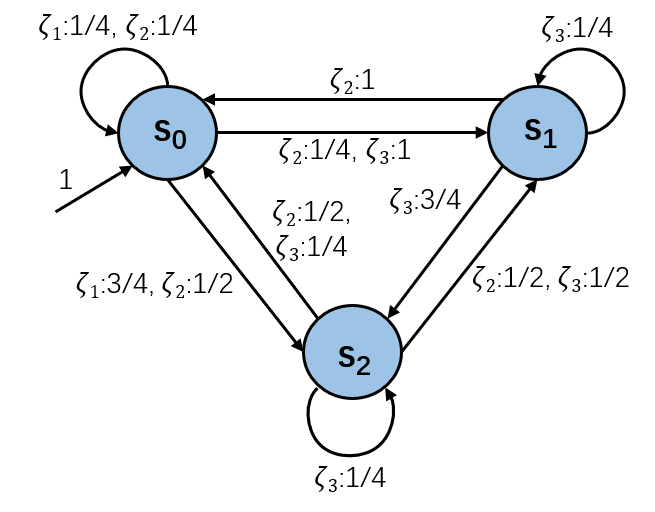}
    \vspace{0.3cm}
    \caption{The visualization of the instance i-WFA.}
    \label{fig:appa}
\end{figure}

\begin{equation*}
\begin{aligned}
    & \hspace{0.3em} = \left[ \hspace{0.3em} 7/16 \hspace{0.7em} 7/16 \hspace{0.7em} 1/8 \hspace{0.3em} \right] \cdot \left[ \begin{array}{ccc}   
    0 & 1 & 0 \\
    0 & 1/4 & 3/4 \\
    1/4 & 1/2 & 1/4 \end{array} \right] \cdot \mathcal{T}\\
    & \hspace{0.3em} = \left[ \hspace{0.3em} 1/32 \hspace{0.7em} 39/64 \hspace{0.7em} 23/64 \hspace{0.3em} \right] \cdot \left[ \begin{array}{cc} 
    1/2 & 1/2 \\
    0 & 1 \\
    3/4 & 1/4 \end{array} \right]\\
    & \hspace{0.3em} = \left[ \hspace{0.3em} 73/256 \hspace{0.7em} 183/256 \hspace{0.3em} \right],
\end{aligned}
\end{equation*}
where the first operator updated recursively represents the ``current state'' of i-WFA with the same size of $\mathcal{I}$ and elements summing to one. It can be found that the probabilistic output is just in the form of the target RNN classifier. This instance i-WFA can be further visualized as Figure~\ref{fig:appa}.

\subsection{Explanation for Gray-Box Property of TSFool}\label{subsec:sm_gray}


Given the white-box setting assumes full model knowledge and the black-box one assumes none, a compromise case between them can be customarily called the gray-box setting \citep{xiang2020side}. When it comes to the usage of model information, the only process to be considered in TSFool is the extraction of the i-WFA from the target RNN classifier. So the reason why we called TSFool a ``gray-box'' method just comes from it. As described in Section~\ref{subsec:sm_iwfa_estab}, the i-WFA extraction only depends on RNN's input, intermediate outputs at each time step, and final output. In this case, given a specific task that can determine their format and value range, the extraction can be done without knowing the specific type, architecture and parameters of the RNN model. 

Then why TSFool is not a black-box method? The problem is, in real-world practice, the situation can be various regarding the intermediate outputs \citep{chambers2020filternet, manaswi2018rnn}. For instance, many-to-many RNNs can provide output at all the time steps, which means the intermediate outputs are directly available. On the other hand, for many-to-one RNNs with the input length variable during inference time, every single intermediate output can be acquired as a final output by stopping inputting the original time series at the corresponding time step, which is still a black-box process. However, for the more basic but commonly used many-to-one RNNs with the input length fixed, we have to artificially construct the intermediate outputs from the RNN hidden state, which is a part of white-box knowledge. Formally, given input length $T$, output function $g_o$, hidden state $h_t$ and output $o_t$ at time step $t$, for $t \in [1, T]$, if $\exists o_t$, s.t. $o_t = g_o(h_t)$, then no matter whether the target model is black- or white-box for attackers, TSFool can be directly used against it. Otherwise, the attackers' capability would be limited to white-box attack as TSFool further requires white-box knowledge (\ie $h_t$ for $t \in [1, T]$ and $g_o$) at this time.

So to ensure the applicability in general, we decided to follow the most strict assumption. That's why we pay more attention to the white-box setting in this paper, and just use the ``gray-box'' to emphasize two properties of TSFool: 1) it can be a black-box method under some specific situations in real-world practice, and 2) it still relies on less model knowledge than gradient-based white-box attacks otherwise. Notice that in our experiments, we never artificially set any gray-box environment to try to limit the performance of the white-box benchmarks and get unfair advantages. We would also like to emphasize again that it is RNN itself that makes it difficult for the benchmark attacks to acquire the full information they need and perform as well as they do upon feed-forward models, even though exactly under the same white-box setting. And this is actually, as mentioned in Section~\ref{subsec:dilemma}, an important motivation for our work.




\section{Detailed Evaluation}\label{sec:sm_9}

In this section, we supplement more details about the evaluation of TSFool, which has been overviewed in Section~\ref{sec:evalu} without sufficient description due to the limited space. Specifically, we firstly provide the experimental setup in Section~\ref{subsec:sm_setup}, followed by the comparisons of TSFool and the benchmarks respectively on univariate and multivariate time series datasets in Section~\ref{subsec:sm_result1} and Section~\ref{subsec:sm_result2}. Next, for fairness, we firstly reveal the impact of a total of 10 hyper-parameters of the experimental methods in Section~\ref{subsec:sm_addExp}, to confirm that the above comparisons under default hyper-parameter settings are representative and of general significance. Also, for a possible concern about the consistency of the final compared data, we provide additional discussion and experimental results in Section~\ref{subsec:sm_fair}, showing whether to evaluate the benchmarks on TPSs or all test samples, our conclusions will remain unchanged. Then in Section~\ref{subsec:sm_human_study}, through two human studies involving 65 volunteers, we verify the advantage of Camouflage Coefficient in representing the real-world imperceptibility of adversarial samples. Finally, in Section~\ref{subsec:sm_anomaly}, given another real-world challenge for the imperceptibility of adversarial time series named anomaly detection, we further confirm that TSFool outperforms existing attacks under four common detection methods.

\begin{table*}[htbp]
\begin{center}
\renewcommand{\arraystretch}{1.1}
\resizebox{17.8cm}{!}{
\begin{tabular}{c|c|c|c|c|c|c|c}
\toprule
\multicolumn{2}{c|}{\textbf{Target Model}} & \hspace{0.7em} \multirow{2}{*}{\textbf{Method}} \hspace{0.7em} & \hspace{0.3em} \multirow{1.2}{*}{\textbf{Attack}} \hspace{0.3em} & \hspace{0.3em} \multirow{1.2}{*}{\textbf{Generation}} \hspace{0.3em} & \hspace{0.3em} \multirow{1.2}{*}{\textbf{Average}} \hspace{0.3em} & \hspace{0.3em} \multirow{1.2}{*}{\textbf{Perturbation}} \hspace{0.3em} & \hspace{0.3em} \multirow{1.2}{*}{\textbf{Camouflage}} \hspace{0.3em} \\
\cline{1-2} 
\multirow{1.2}{*}{\textbf{Dataset}} & \multirow{1.2}{*}{\textbf{Accuracy}} & & \multirow{0.8}{*}{\textbf{Success Rate}} & \multirow{0.8}{*}{\textbf{Number}} & \multirow{0.8}{*}{\textbf{Time Cost (s)}} & \multirow{0.8}{*}{\textbf{Ratio ($\rho^*$)}} & \multirow{0.8}{*}{\textbf{Coefficient}} \\

\midrule
\midrule
\multirow{10}{*}{CBF} & \multirow{10}{*}{0.7511} & FGSM & 77.89\% & \multirow{9}{*}{900} & 0.0024 & 8.24\% & 0.8661 \\
& & JSMA & 87.11\% & & 11.5998 & 53.22\% & 0.8978 \\
& & DeepFool & 83.11\% & & 0.1113 & 13.34\% & 0.9237 \\
& & PGD (BIM) & 77.56\% & & 0.1699 & 7.40\% & 0.8455 \\
& & C\&W & 72.89\% & & 3.9842 & 2.46\% & 0.7937 \\
& & Auto-Attack & 77.67\% & & 0.3968 & 6.33\% & 0.8513 \\
& & Boundary Attack & 81.78\% & & 16.7385 & 1.53\% & 0.7397 \\
& & HopSkipJump & 84.00\% & & 16.4119 & 1.87\% & 0.7394 \\
& & Transfer Attack & 25.78\% & & - & 2.60\% & 1.0105 \\
& & TSFool & 75.69\% & 720 & 0.0297 & 7.48\% & 0.7447 \\

\midrule
\multirow{10}{*}{DPOAG} & \multirow{10}{*}{0.7842} & FGSM & 71.22\% & \multirow{8}{*}{139} & 0.0015 & 36.32\% & 0.9044 \\
& & JSMA & 97.84\% & & 0.0680 & 3.47\% & 0.8981 \\
& & DeepFool & 80.58\% & & 0.0217 & 16.91\% & 0.9356 \\
& & PGD (BIM) & 88.49\% & & 0.1065 & 24.70\% & 1.0163 \\
& & C\&W & 76.26\% & & 2.8604 & 7.89\% & 0.7868 \\
& & Auto-Attack & 88.49\% & & 0.1180 & 24.90\% & 1.0132 \\
& & Boundary Attack & 81.29\% & & 6.9011 & 3.47\% & 0.8737 \\
& & HopSkipJump & 84.89\% & & 10.4413 & 4.02\% & 0.8560 \\
& & Transfer Attack & 14.25\% & 400 & - & 9.62\% & 1.8219 \\
& & TSFool & 87.86\% & 140 & 0.0242 & 4.93\% & 1.6045 \\

\midrule
\multirow{10}{*}{DPOC} & \multirow{10}{*}{0.7319} & FGSM & 73.55\% & \multirow{8}{*}{276} & 0.0014 & 24.00\% & 1.1040 \\
& & JSMA & 73.19\% & & 0.0393 & 1.26\% & 0.9639 \\
& & DeepFool & 73.19\% & & 0.0172 & 9.66\% & 1.0428 \\
& & PGD (BIM) & 72.83\% & & 0.1093 & 17.97\% & 1.0446 \\
& & C\&W & 72.83\% & & 2.5831 & 4.45\% & 0.9820 \\
& & Auto-Attack & 73.19\% & & 0.1130 & 17.98\% & 1.0457 \\
& & Boundary Attack & 69.93\% & & 6.0584 & 1.76\% & 0.9471 \\
& & HopSkipJump & 72.83\% & & 9.9818 & 1.97\% & 0.9531 \\
& & Transfer Attack & 32.33\% & 600 & - & 5.24\% & 1.0938 \\
& & TSFool & \cellcolor{SpringGreen}\underline{93.64\%} & 660 & 0.0055 & 2.93\% & \cellcolor{SpringGreen}\underline{0.8401} \\

\midrule
\multirow{10}{*}{ECG200} & \multirow{10}{*}{0.7400} & FGSM & 64.00\% & \multirow{9}{*}{100} & 0.0018 & 13.91\% & 1.0854 \\
& & JSMA & 70.00\% & & 2.7941 & 41.19\% & 1.1539 \\
& & DeepFool & 74.00\% & & 0.0183 & 21.37\% & 1.1692 \\
& & PGD (BIM) & 67.00\% & & 0.1335 & 9.22\% & 0.9705 \\
& & C\&W & 73.00\% & & 2.8608 & 4.41\% & 0.9526 \\
& & Auto-Attack & 74.00\% & & 0.2824 & 8.57\% & 0.9785 \\
& & Boundary Attack & 68.00\% & & 10.3866 & 1.81\% & 0.8460 \\
& & HopSkipJump & 74.00\% & & 12.4769 & 2.39\% & 0.8607 \\
& & Transfer Attack & 27.00\% & & - & 4.26\% & 1.0422 \\
& & TSFool & \cellcolor{SpringGreen}\underline{94.29\%} & 140 & 0.0086 & 4.41\% & \cellcolor{SpringGreen}\underline{0.6291} \\

\midrule
\multirow{10}{*}{GP} & \multirow{10}{*}{0.9333} & FGSM & 78.67\% & \multirow{9}{*}{150} & 0.0028 & 20.07\% & 1.1798 \\
& & JSMA & 90.67\% & & 0.9790 & 8.11\% & 1.0561 \\
& & DeepFool & 92.67\% & & 0.0449 & 25.98\% & 1.0870 \\
& & PGD (BIM) & 85.33\% & & 0.2079 & 14.07\% & 1.1171 \\
& & C\&W & 76.67\% & & 4.7081 & 3.99\% & 1.0049 \\
& & Auto-Attack & 88.00\% & & 0.3183 & 13.36\% & 1.0983 \\
& & Boundary Attack & 90.67\% & & 22.0909 & 1.83\% & 1.0507 \\
& & HopSkipJump & 93.33\% & & 18.7358 & 2.94\% & 1.0575 \\
& & Transfer Attack & 8.67\% & & - & 5.06\% & 0.7851 \\
& & TSFool & \cellcolor{SpringGreen}\underline{100.00\%} & 80 & 0.1191 & 3.11\% & \cellcolor{SpringGreen}\underline{0.6815} \\

\bottomrule
\end{tabular}
}
\end{center}
\caption{The table shows the performance of the experimental methods in the 10 UCR univariate time series datasets, including the original model accuracy, the ASR, the number of samples generated, the time cost and the two measures $\rho^*$ and $\mathcal{C}$ for imperceptibility. As colored in \textcolor{Green}{green}, TSFool respectively achieves the state-of-the-art ASR and CC on six datasets among the experimental ones.}
\label{tab:average_full}
\end{table*}

\begin{table*}[htbp]
\ContinuedFloat
\begin{center}
\renewcommand{\arraystretch}{1.1}
\resizebox{17.8cm}{!}{
\begin{tabular}{c|c|c|c|c|c|c|c}
\toprule
\multicolumn{2}{c|}{\textbf{Target Model}} & \hspace{0.7em} \multirow{2}{*}{\textbf{Method}} \hspace{0.7em} & \hspace{0.3em} \multirow{1.2}{*}{\textbf{Attack}} \hspace{0.3em} & \hspace{0.3em} \multirow{1.2}{*}{\textbf{Generation}} \hspace{0.3em} & \hspace{0.3em} \multirow{1.2}{*}{\textbf{Average}} \hspace{0.3em} & \hspace{0.3em} \multirow{1.2}{*}{\textbf{Perturbation}} \hspace{0.3em} & \hspace{0.3em} \multirow{1.2}{*}{\textbf{Camouflage}} \hspace{0.3em} \\
\cline{1-2} 
\multirow{1.2}{*}{\textbf{Dataset}} & \multirow{1.2}{*}{\textbf{Accuracy}} & & \multirow{0.8}{*}{\textbf{Success Rate}} & \multirow{0.8}{*}{\textbf{Number}} & \multirow{0.8}{*}{\textbf{Time Cost (s)}} & \multirow{0.8}{*}{\textbf{Ratio ($\rho^*$)}} & \multirow{0.8}{*}{\textbf{Coefficient}} \\

\midrule
\midrule
\multirow{10}{*}{IPD} & \multirow{10}{*}{0.9650} & FGSM & 81.05\% & \multirow{9}{*}{1029} & 0.0003 & 15.32\% & 1.1559 \\
& & JSMA & 96.50\% & & 0.0304 & 6.15\% & 0.8969 \\
& & DeepFool & 96.50\% & & 0.0050 & 67.81\% & 1.2897 \\
& & PGD (BIM) & 83.38\% & & 0.0176 & 15.18\% & 1.1603 \\
& & C\&W & 71.91\% & & 1.0493 & 5.78\% & 0.9439 \\
& & Auto-Attack & 83.58\% & & 0.0369 & 13.21\% & 1.1597 \\
& & Boundary Attack & 93.00\% & & 0.6569 & 7.09\% & 0.8815 \\
& & HopSkipJump & 96.50\% & & 1.9331 & 8.15\% & 0.9172 \\
& & Transfer Attack & 6.80\% & & - & 4.98\% & 1.0780 \\
& & TSFool & 83.53\% & 340 & 0.0056 & \cellcolor{SpringGreen}\underline{2.98\%} & \cellcolor{SpringGreen}\underline{0.8060} \\

\midrule
\multirow{10}{*}{MPOAG} & \multirow{10}{*}{0.6429} & FGSM & 76.62\% & \multirow{8}{*}{154} & 0.0028 & 67.65\% & 1.0604 \\
& & JSMA & 75.97\% & & 4.2364 & 215.51\% & 1.1461 \\
& & DeepFool & 78.57\% & & 0.0441 & 20.08\% & 0.9917 \\
& & PGD (BIM) & 77.92\% & & 0.2168 & 39.03\% & 0.9940 \\
& & C\&W & 67.53\% & & 4.2949 & 5.67\% & 1.0384 \\
& & Auto-Attack & 77.27\% & & 0.2294 & 39.26\% & 0.9902 \\
& & Boundary Attack & 75.32\% & & 12.1604 & 5.65\% & 1.0050 \\
& & HopSkipJump & 82.47\% & & 19.8336 & 7.29\% & 1.0176 \\
& & Transfer Attack & 20.75\% & 400 & - & 9.46\% & 1.4861 \\
& & TSFool & \cellcolor{SpringGreen}\underline{88.33\%} & 60 & 0.0614 & 5.72\% & \cellcolor{SpringGreen}\underline{0.6833} \\

\midrule
\multirow{10}{*}{MPOC} & \multirow{10}{*}{0.6392} & FGSM & 53.95\% & \multirow{8}{*}{291} & 0.0015 & 55.65\% & 0.9892 \\
& & JSMA & 63.92\% & & 0.0364 & 2.61\% & 0.8737 \\
& & DeepFool & 63.92\% & & 0.0225 & 23.09\% & 0.9543 \\
& & PGD (BIM) & 62.20\% & & 0.1103 & 28.79\% & 0.9668 \\
& & C\&W & 50.17\% & & 3.1460 & 8.10\% & 0.8964 \\
& & Auto-Attack & 63.92\% & & 0.1263 & 29.83\% & 0.9659 \\
& & Boundary Attack & 59.79\% & & 6.6208 & 2.61\% & 0.8818 \\
& & HopSkipJump & 62.20\% & & 9.4942 & 3.23\% & 0.8873 \\
& & Transfer Attack & 51.67\% & 600 & - & 9.35\% & 1.0294 \\
& & TSFool & \cellcolor{SpringGreen}\underline{81.75\%} & 400 & 0.0093 & 5.46\% & \cellcolor{SpringGreen}\underline{0.7291} \\

\midrule
\multirow{10}{*}{PPOAG} & \multirow{10}{*}{0.8976} & FGSM & 87.32\% & \multirow{9}{*}{205} & 0.0015 & 71.52\% & 1.1642 \\
& & JSMA & 96.10\% & & 0.0459 & 4.44\% & 0.6142 \\
& & DeepFool & 92.20\% & & 0.0280 & 30.70\% & 0.9812 \\
& & PGD (BIM) & 91.22\% & & 0.1139 & 46.90\% & 0.8188 \\
& & C\&W & 37.56\% & & 2.9188 & 3.90\% & 1.0625 \\
& & Auto-Attack & 95.61\% & & 0.1250 & 47.77\% & 0.8226 \\
& & Boundary Attack & 91.22\% & & 6.9941 & 5.64\% & 0.5684 \\
& & HopSkipJump & 95.12\% & & 10.5637 & 7.14\% & 0.5263 \\
& & Transfer Attack & 11.71\% & & - & 16.46\% & 1.6559 \\
& & TSFool & 87.14\% & 140 & 0.0367 & 15.74\% & 0.9019 \\

\midrule
\multirow{10}{*}{PPOC} & \multirow{10}{*}{0.7869} & FGSM & 42.27\% & \multirow{9}{*}{291} & 0.0016 & 64.13\% & 1.4325 \\
& & JSMA & 78.69\% & & 0.0138 & 1.08\% & 0.8480 \\
& & DeepFool & 78.35\% & & 0.0241 & 20.10\% & 0.9142 \\
& & PGD (BIM) & 60.48\% & & 0.1101 & 32.71\% & 0.9953 \\
& & C\&W & 74.57\% & & 2.9646 & 5.16\% & 0.8856 \\
& & Auto-Attack & 78.69\% & & 0.1468 & 33.26\% & 0.8527 \\
& & Boundary Attack & 73.88\% & & 6.4591 & 1.30\% & 0.8557 \\
& & HopSkipJump & 78.69\% & & 10.3486 & 1.61\% & 0.8664 \\
& & Transfer Attack & 15.81\% & & - & 13.48\% & 1.2120 \\
& & TSFool & \cellcolor{SpringGreen}\underline{85.56\%} & 540 & 0.0081 & 2.40\% & 1.1306 \\

\bottomrule
\end{tabular}
}
\end{center}
\caption{(Cont.)}
\end{table*}

\begin{figure*}[htb]
    \centering
    \includegraphics[width=17.8cm]{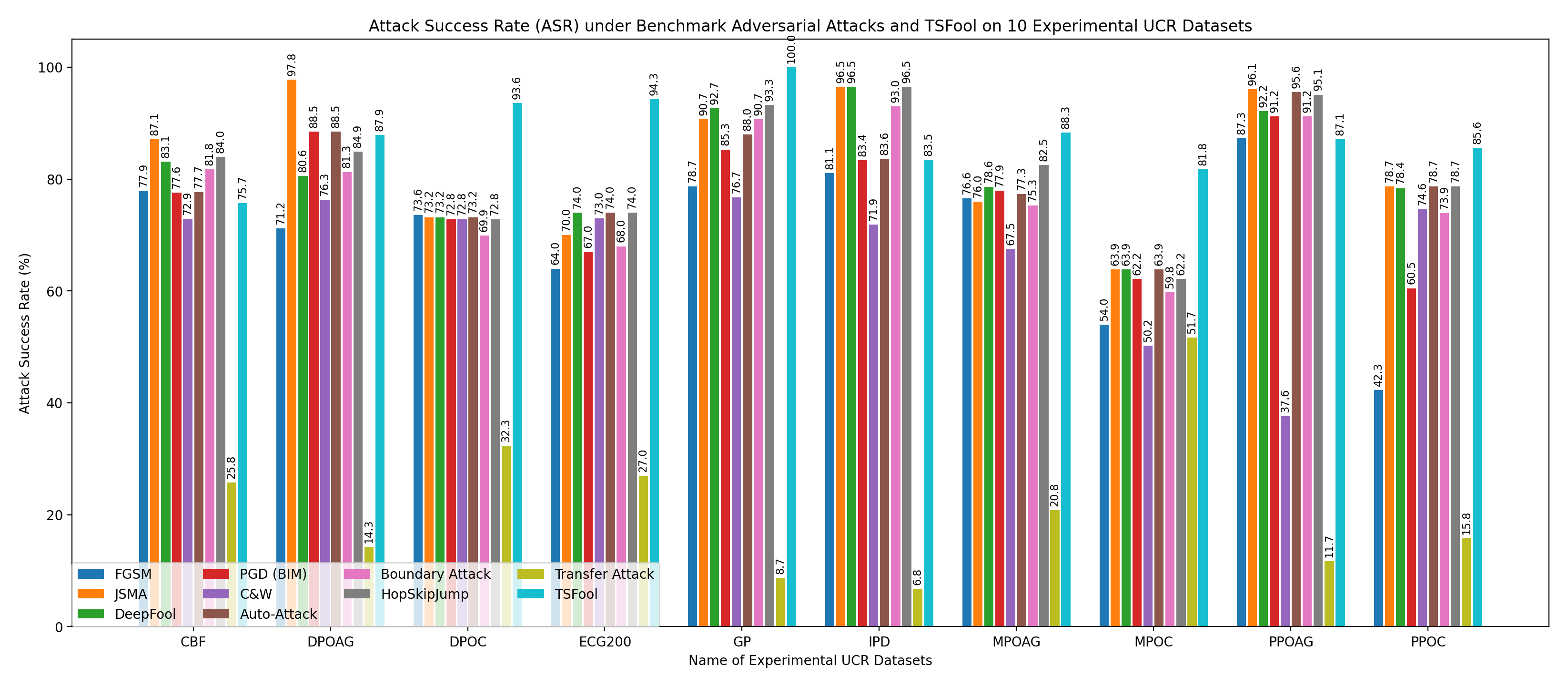}
    \vspace{0.3cm}
    \caption{Corresponding to Table~\ref{tab:average_full}, the figure illustrates the ASR of TSFool and the nine benchmark attacks on the 10 experimental UCR datasets, which empirically confirms that TSFool significantly outperforms the benchmarks in general. \vspace{0.6cm}}
    \label{fig:accuracy}
\end{figure*}

\begin{figure*}[htb]
    \centering
    \includegraphics[width=17.4cm]{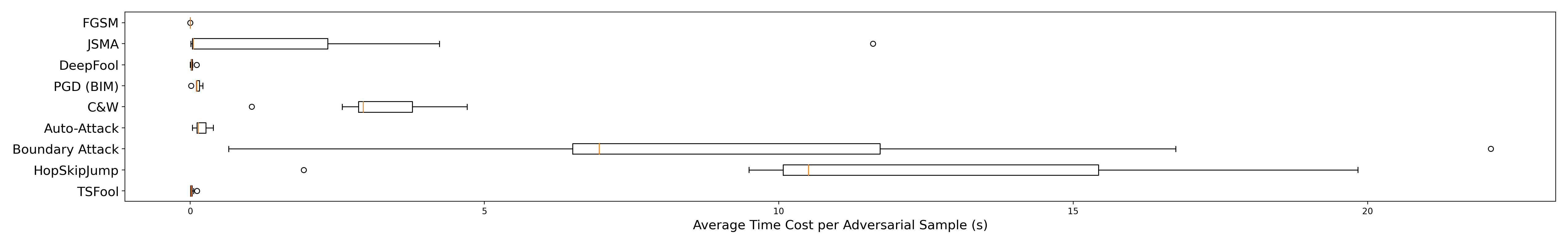}
    \vspace{0.3cm}
    \caption{Corresponding to Table~\ref{tab:average_full}, the figure shows the time cost distributions of the experimental methods, intuitively revealing their gaps in efficiency. Specifically, the time cost of C\&W, Boundary Attack and HopSkipJump are dramatically higher than others, which may even make them intolerable in some real-world practices. JSMA seems fine on average, but is unstable and can be time-consuming on specific cases. Other experimental methods are relatively reasonable in efficiency, with TSFool becoming the suboptimal one only behind FGSM. \vspace{0.6cm}}
    \label{fig:time_new}
\end{figure*}

\subsection{Experimental Setup}\label{subsec:sm_setup}

In our experiments, the \textit{long short-term memory} (LSTM) is adopted for the target RNN classifiers, and 10 univariate plus one multivariate time series datasets (namely CBF, DPOAG, DPOC, ECG200, GP, IPD, MPOAG, MPOC, PPOAG, PPOC and AF) are respectively derived from the public UCR \citep{dau2019ucr} and UEA \citep{bagnall2018uea} time series archives as the target datasets. Our experiments mainly focus on univariate time series since the majority of real-world TSC applications are developed for such data \citep{ding2022towards}. This can also be doubly confirmed by the difference in the size of UCR (containing 128 univariate datasets) and UEA (containing 30 multivariate datasets) archives \citep{bagnall2018uea}. Nevertheless, we still include some simple experiments for multivariate time series to show the wide applicability of TSFool. Notice that we select the target datasets following the UCR briefing document strictly to make sure there is no cherry-picking.

The benchmarks used in our experiments cover the existing adversarial attacks from basic to state-of-the-art ones, including six white-box methods, namely FGSM, JSMA, DeepFool, PGD (BIM), C\&W and Auto-Attack, as well as three black-box methods, namely Boundary Attack, HopSkipJump and Transfer Attack. All of the benchmark methods have already been introduced in Section~\ref{subsec:basics}. The public adversarial set used for Transfer Attack is from \citet{fawaz2019adversarial}, which is generated from a \textit{residual network} (ResNet) by BIM attack. The implementations of other benchmarks come from a public Python library named Adversarial Robustness Toolbox (ART) \citep{nicolae2018adversarial}. Notice that to ensure the fairness of comparison and the general significance of the results, we first report the results with all of the experimental methods including TSFool using default hyper-parameters in Section~\ref{subsec:sm_result1} and Section~\ref{subsec:sm_result2}, and then provide additional experiments for the impact of different hyper-parameter settings in Section~\ref{subsec:sm_addExp}. 
All of our experiments are conducted on Windows 10 Enterprise 20H2 x64 coming with AMD Ryzen 5 3600 6-Core 3.60 GHz CPU and 64GB RAM, and are implemented with Python 3.11.5 and PyTorch 2.0.1.

\begin{figure*}
  \centering
  \begin{subfigure}{0.19\linewidth}
    \includegraphics[width=1\linewidth]{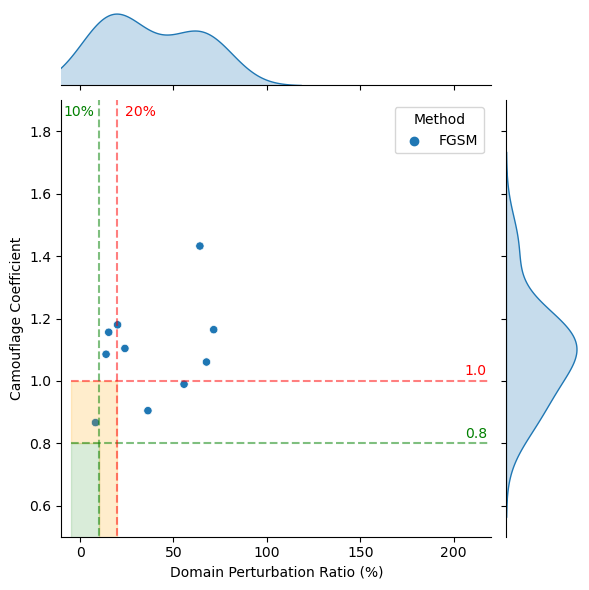}
  \end{subfigure}
  \begin{subfigure}{0.19\linewidth}
    \includegraphics[width=1\linewidth]{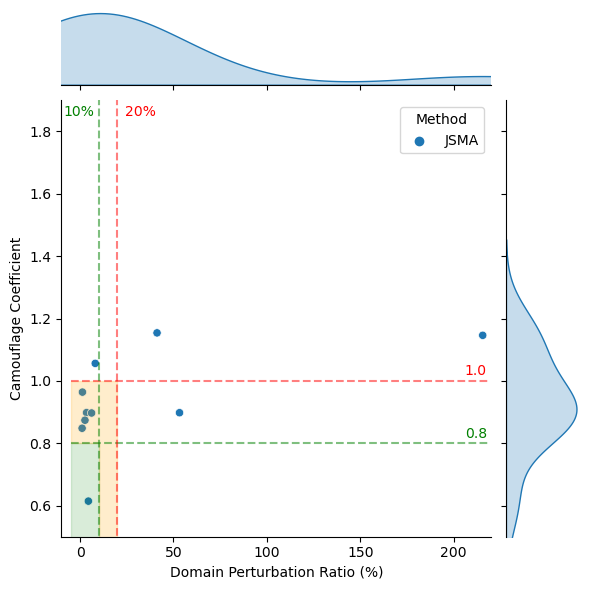}
  \end{subfigure}
  \begin{subfigure}{0.19\linewidth}
    \includegraphics[width=1\linewidth]{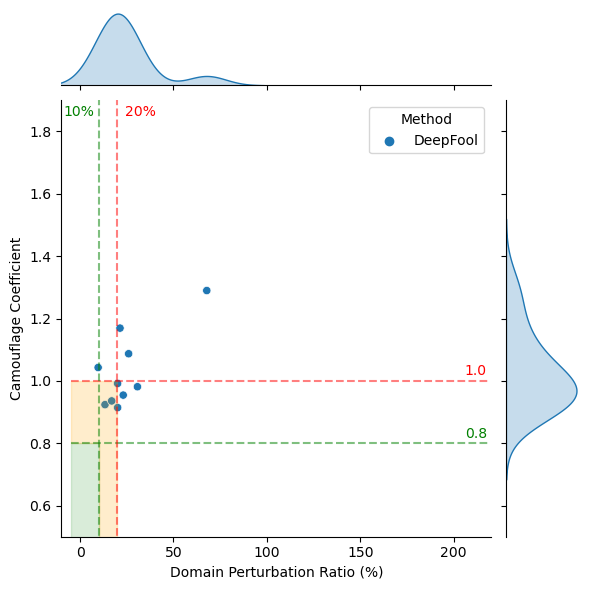}
  \end{subfigure}
  \begin{subfigure}{0.19\linewidth}
    \includegraphics[width=1\linewidth]{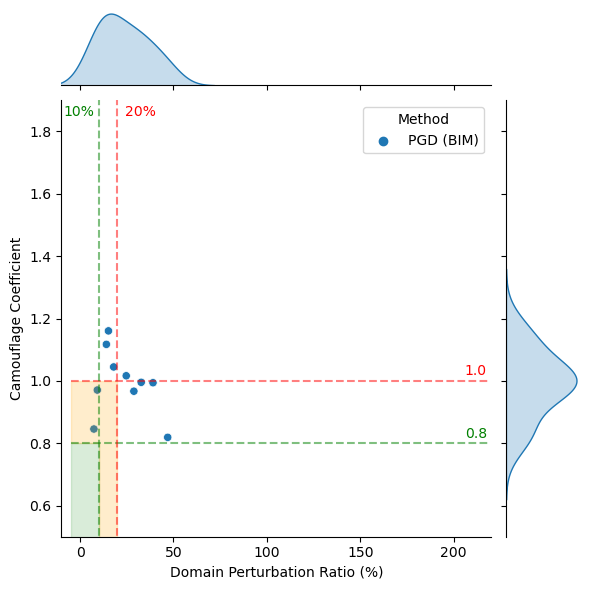}
  \end{subfigure}
  \begin{subfigure}{0.19\linewidth}
    \includegraphics[width=1\linewidth]{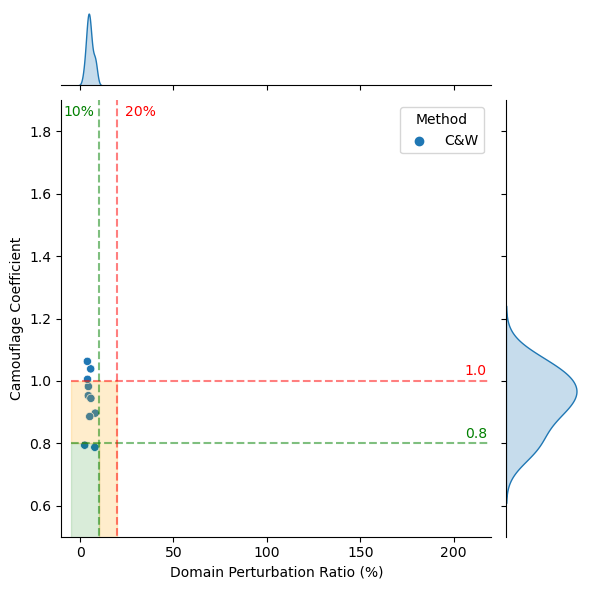}
  \end{subfigure}
  \centering
  \begin{subfigure}{0.19\linewidth}
    \includegraphics[width=1\linewidth]{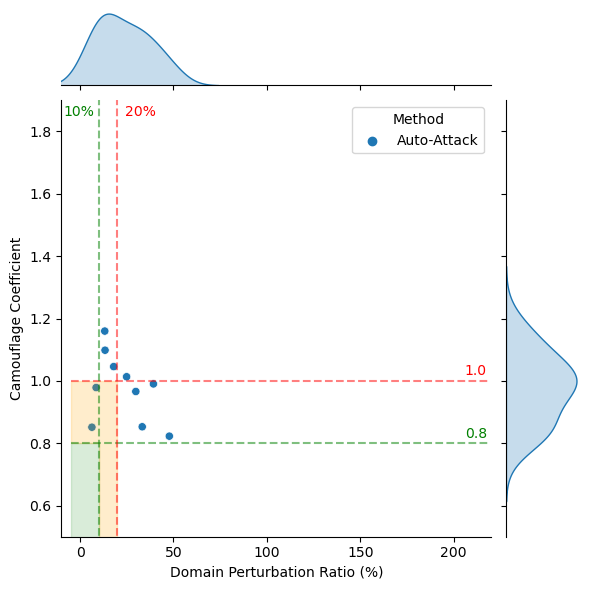}
  \end{subfigure}
  \begin{subfigure}{0.19\linewidth}
    \includegraphics[width=1\linewidth]{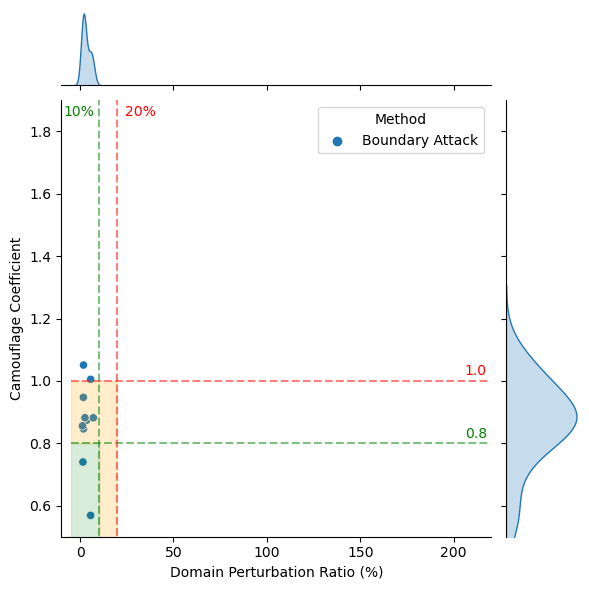}
  \end{subfigure}
  \begin{subfigure}{0.19\linewidth}
    \includegraphics[width=1\linewidth]{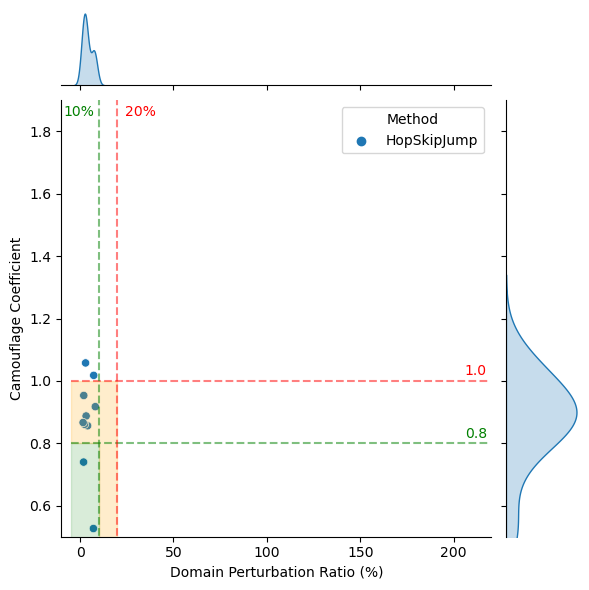}
  \end{subfigure}
  \begin{subfigure}{0.19\linewidth}
    \includegraphics[width=1\linewidth]{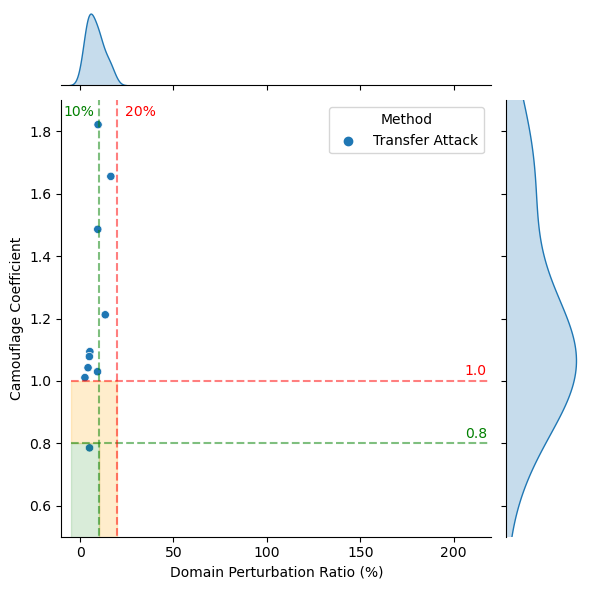}
  \end{subfigure}
  \begin{subfigure}{0.19\linewidth}
    \includegraphics[width=1\linewidth]{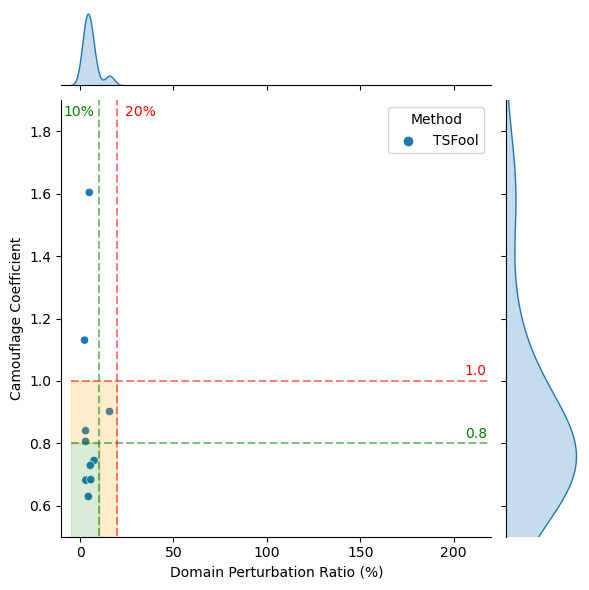}
  \end{subfigure}
  \vspace{0.3cm}
  \caption{Corresponding to Table~\ref{tab:average_full}, the figures illustrate the joint distribution of Domain Perturbation Ratio $\rho^*$ and Camouflage Coefficient $\mathcal{C}$ of the experimental methods on the 10 UCR datasets, with each point representing a specific dataset, to intuitively compare their imperceptibility in general. Here we set $\rho^* < 10\%$ and $\mathcal{C} < 0.8$ as ``Excellent'', then $\rho^* < 20\%$ and $\mathcal{C} < 1$ as ``Reasonable'', just as the \textcolor{Green}{green} and \textcolor{Red}{red} lines, which naturally split the figures into three ranges respectively colored in \textcolor{Green}{green}, \textcolor{Orange}{orange} and none. It can be found that TSFool is the state-of-the-art one whether considering the number of points falling into the ``Excellent'' range (\ie five) or the ``Acceptable'' range (\ie eight). Also, from the density curves on the marginal axes, C\&W, Boundary Attack, HopSkipJump, Transfer Attack and TSFool outperform the other five methods in local perturbation. At the same time, for the global CC, the peak value of TSFool is the smallest among all the experimental methods. These figures empirically enhance our confidence in the imperceptibility of TSFool. \vspace{0.6cm}}
  \label{fig:perturb}
\end{figure*}

\subsection{Results on Univariate Time Series}\label{subsec:sm_result1}

\paragraph{Effectiveness}\label{subsubsec:effective} 
In Figure~\ref{fig:accuracy}, we illustrate the ASR under the experimental methods on the 10 UCR univariate time series datasets. It can be found that all the methods except the Transfer Attack can steadily realize effective attack, while the ASR of TSFool is significantly higher than the benchmark methods in general, respectively beating the benchmarks by 15.64\%, 4.23\%, 6.18\%, 10.92\%, 17.86\%, 7.65\%, 8.75\%, 4.59\%, 68.22\% on average. This empirically illustrates the impressive effectiveness of TSFool in RNN-based TSC. Notice that as shown in Table~\ref{tab:average_full}, the original accuracies of the experimental classifiers are from 63.92\% to 96.50\%, which basically covers most common cases in real-world practice regarding the model quality. As a result, these experiments are of general significance. 
 

\paragraph{Efficiency}\label{subsubsec:efficiency}
To evaluate the efficiency, we record the number of samples generated in the adversarial set and the total execution time of the experimental methods on all the experimental datasets, and accordingly calculate the average time cost for crafting a single sample. As shown in Table~\ref{tab:average} and Table~\ref{tab:average_full}, the efficiency of TSFool and DeepFool are basically at the same level. Compared with them, PGD (BIM) and Auto-Attack are slower by one order of magnitude, then JSMA, C\&W and Boundary Attack by two. HopSkipJump is the most time-consuming method, slower by three orders of magnitude. For the Transfer Attack based on ResNet and BIM, as the time cost is not reported in the original paper \citep{fawaz2019adversarial}, we have to leave the position empty. However, it should not be expected to be faster than directly using PGD (BIM) in our experiment, since typically ResNet is more complex than the LSTM classifiers we used. The only benchmark faster than TSFool is FGSM, which is not surprising since FGSM is a basic single-step method without outstanding performance in the other three measures. As a result, we can state that TSFool is efficient enough. Additionally, we show the distributions of time cost of the experimental methods in Figure~\ref{fig:time_new} to intuitively reveal their dramatic difference in efficiency.

\paragraph{Imperceptibility}\label{subsubsec:impercept}
As the evaluation of imperceptibility, we report both the local Domain Perturbation Ratio $\rho^*$ and the global Camouflage Coefficient $\mathcal{C}$ of the experimental methods in Table~\ref{tab:average} and Table~\ref{tab:average_full}. For local perturbation, it can be found that the $\rho^*$ of FGSM, JSMA, DeepFool, PGD (BIM) and Auto-Attack are relatively large, all of which are beyond 15\%, which means the generated perturbations may be easily detected. On the contrary, the rest five basically control the perturbation in a reasonable range, from 3.04\% to 7.68\%. The only two benchmarks better than TSFool in $\rho^*$ are Boundary Attack and HopSkipJump. Unfortunately, they are also the worst two in efficiency as aforementioned. Then from the global perspective, the $\mathcal{C}$ values of three benchmarks namely FGSM, DeepFool and Transfer Attack exceed one on average. As argued in Section~\ref{subsec:reconsider}, this means the majority of samples are already perturbed to be closer to the adversarial class instead of the original one, in which they should have been hidden. Strictly speaking, these attacks actually fail under our definition. The $\mathcal{C}$ values of the rest six benchmarks are less than one, while TSFool becomes the state-of-the-art method in CC, outperforming them by 0.0641 to 0.1791. In Figure~\ref{fig:perturb}, we further illustrate the joint distribution of $\rho^*$ and $\mathcal{C}$ for TSFool and the benchmarks based on their specific performance on the 10 UCR datasets, to empirically confirm the imperceptibility of TSFool considering the two perspectives together.

\subsection{Results on Multivariate Time Series}\label{subsec:sm_result2}

\begin{figure*}[htb]
  \centering
  \begin{subfigure}{0.19\linewidth}
    \includegraphics[width=1\linewidth]{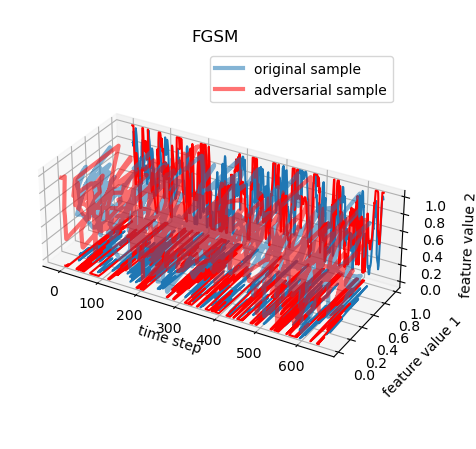}
  \end{subfigure}
  \begin{subfigure}{0.19\linewidth}
    \includegraphics[width=1\linewidth]{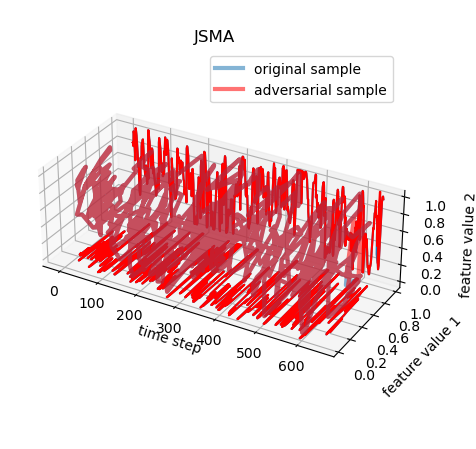}
  \end{subfigure}
  \begin{subfigure}{0.19\linewidth}
    \includegraphics[width=1\linewidth]{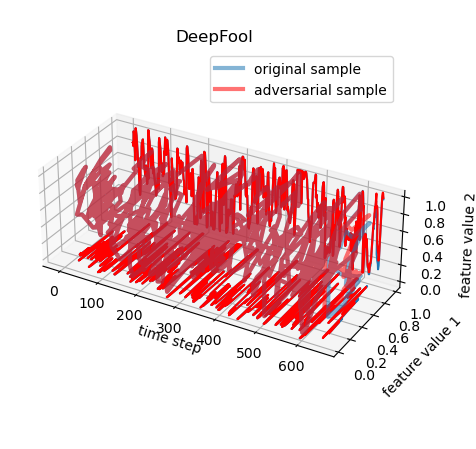}
  \end{subfigure}
  \begin{subfigure}{0.19\linewidth}
    \includegraphics[width=1\linewidth]{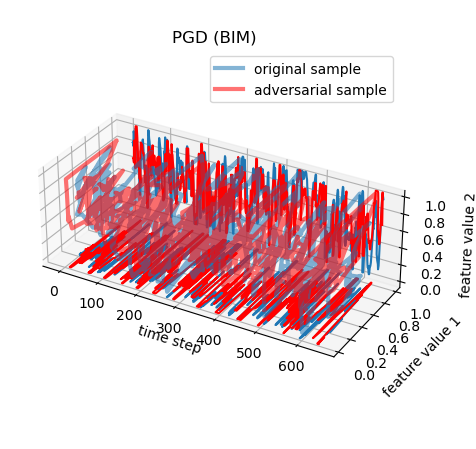}
  \end{subfigure}
  \begin{subfigure}{0.19\linewidth}
    \includegraphics[width=1\linewidth]{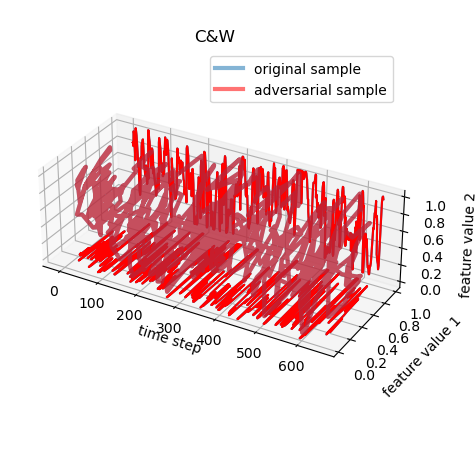}
  \end{subfigure}
  \centering
  \begin{subfigure}{0.19\linewidth}
    \includegraphics[width=1\linewidth]{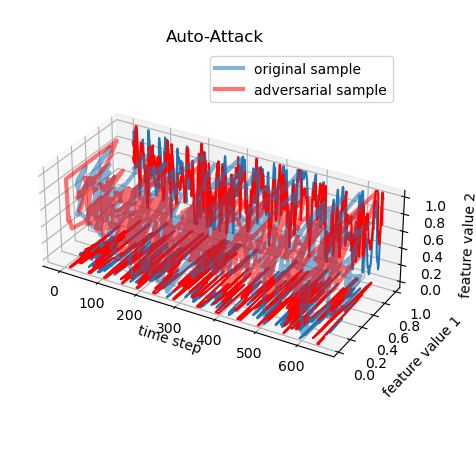}
  \end{subfigure}
  \begin{subfigure}{0.19\linewidth}
    \includegraphics[width=1\linewidth]{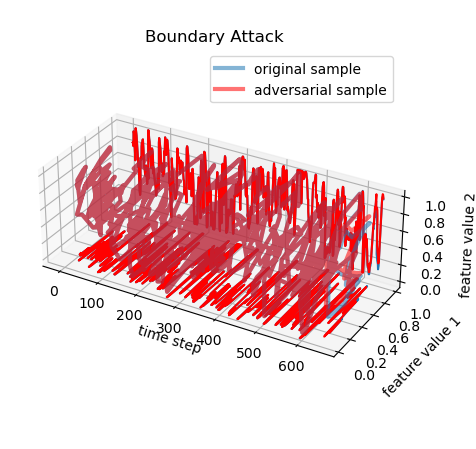}
  \end{subfigure}
  \begin{subfigure}{0.19\linewidth}
    \includegraphics[width=1\linewidth]{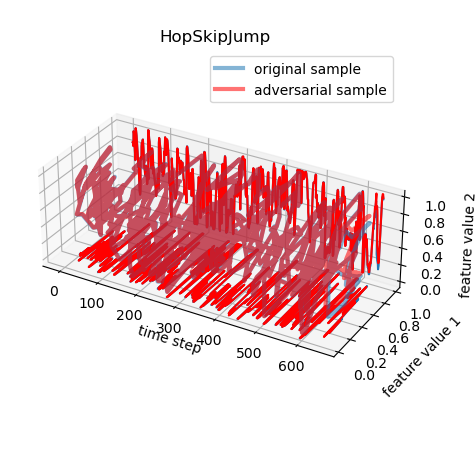}
  \end{subfigure}
  \begin{subfigure}{0.19\linewidth}
    \includegraphics[width=1\linewidth]{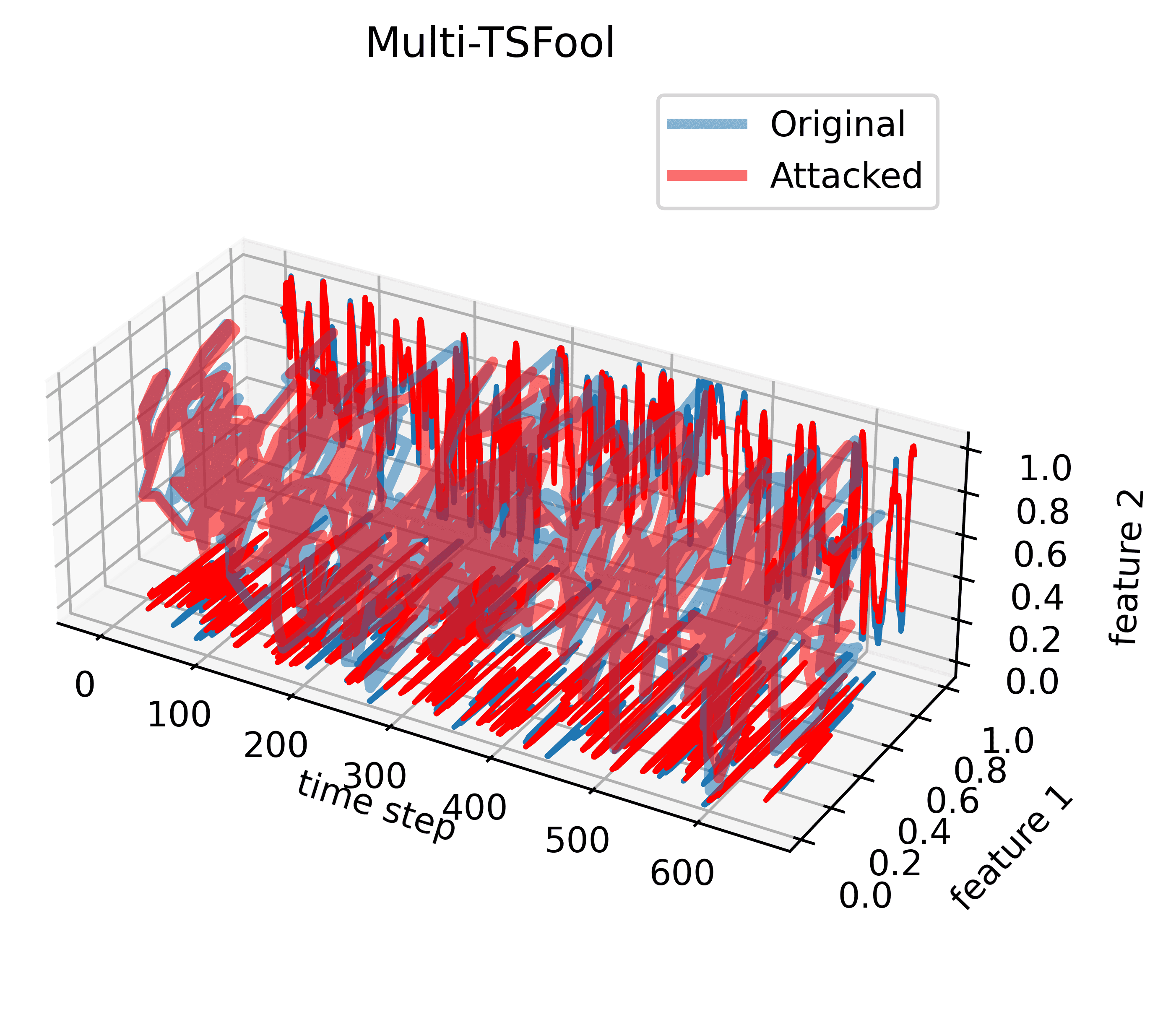}
  \end{subfigure}
  \vspace{0.26cm}
  \caption{Corresponding to Table~\ref{tab:multi}, this is an example of adversarial attacks through the rest eight benchmarks and Multi-TSFool on the AF dataset, in which the samples have two features at each time step. From the projection at each feature dimension, it can be found that the performance of FGSM, PGD (BIM) and Auto-Attack is basically at the same level as in Figure~\ref{fig:compare}, and the problem of HopSkipJump is still the intolerable efficiency. The results of C\&W and Boundary Attack seem impressive, while they are the worst two in ASR, and the instance ``adversarial'' samples from them in this figure actually fail to fool the target classifier. JSMA and DeepFool become competitive in this case, while TSFool still outperforms them in ASR, Time Cost and CC on average. \vspace{0.73cm}}
  \label{fig:compare-multi}
\end{figure*}

For multivariate time series, we evaluate Multi-TSFool on the dataset AF from the UEA archive. Every single sample in AF consists of data points with two features at each of the time steps (\ie $D = 2$). Similarly, we report the ASR, average time cost, perturbation and CC of the experimental methods in Table~\ref{tab:multi}, which also shows the advantages of TSFool on multivariate time series data in effectiveness and imperceptibility. Nevertheless, there are two differences to be noticed: Firstly, there is no existing publicly available adversarial dataset for AF, so considering the fairness of comparison, we have to remove the Transfer Attack here; Secondly, as each data point no longer just has a single feature, there are several options for the distance metric. For the Domain Perturbation Ratio $\rho^*$, we respectively report the results under $\ell_1$, $\ell_2$ and $\ell_{\infty}$, because they can be meaningful for different real-world cases \citep{kotyan2019adversarial}. While for the Camouflage Coefficient $\mathcal{C}$, we simply use the default $\ell_2$ distance to measure the relative position. An intuitive example of the experimental methods on the UEA-AF multivariate dataset is also given in Figure~\ref{fig:compare-multi}.

\begin{table}[H]
\begin{center}
\resizebox{8.6cm}{!}{
\begin{tabular}{c|c|ccccc|c}
\toprule
\multirow{2}{*}{\textbf{Method}} & \textbf{Hyper-} & \multicolumn{5}{c|}{\multirow{0.9}{*}{\textbf{Tried Values}}} & \multirow{2}{*}{\textbf{Default}}\\
\cline{3-7} 
 & \textbf{parameter} & \multirow{1.3}{*}{\textbf{1}} & \multirow{1.3}{*}{\textbf{2}} & \multirow{1.3}{*}{\textbf{3}} & \multirow{1.3}{*}{\textbf{4}} & \multirow{1.3}{*}{\textbf{5}} & \\
\midrule
\midrule
FGSM & $eps$ & 0.01 & 0.05 & 0.1 & 0.5 & 1 & 0.3 \\
\midrule
JSMA & $gamma$ & 0.01 & 0.05 & 0.1 & 0.5 & 1 & 1 \\
\midrule
DeepFool & $max$ $iter$ & 1 & 5 & 10 & 50 & 100 & 100 \\
\midrule
\multirow{3}{*}{PGD (BIM)} & $eps$ & 0.01 & 0.05 & 0.1 & 0.5 & 1 & 0.3 \\
& $eps$ $step$ & 0.01 & 0.05 & 0.1 & 0.5 & 1 & 0.1 \\
& $max$ $iter$ & 1 & 5 & 10 & 50 & 100 & 100 \\
\midrule
\multirow{2}{*}{C\&W} & $learning$ $rate$ & 0.001 & 0.005 & 0.01 & 0.05 & 0.1 & 0.01 \\
& $max$ $iter$ & 1 & 5 & 10 & 50 & 100 & 10 \\
\midrule
Auto-Attack & $eps$ & 0.01 & 0.05 & 0.1 & 0.5 & 1 & 0.3 \\
\midrule
& $delta$ & 0.001 & 0.005 & 0.01 & 0.05 & 0.1 & 0.01 \\
\multirow{2}{*}{Boundary} & $epsilon$ & 0.001 & 0.005 & 0.01 & 0.05 & 0.1 & 0.01 \\
\multirow{2}{*}{Attack} & $min$ $epsilon$ & 0 & 0.005 & 0.01 & 0.05 & 0.1 & 0 \\
& $init$ $size$ & 1 & 5 & 10 & 50 & 100 & 100 \\
& $num$ $trial$ & 1 & 5 & 10 & 25 & 50 & 25 \\
\midrule
\multirow{2}{*}{HopSkipJump} & $max$ $iter$ & 1 & 5 & 10 & 50 & 100 & 50 \\
& $init$ $size$ & 1 & 5 & 10 & 50 & 100 & 100 \\
\bottomrule
\end{tabular}
}
\end{center}
\vspace{0.1cm}
\caption{The table illustrates the common hyper-parameters of the benchmarks involved in our experiments in Section~\ref{subsec:sm_addExp}, with the specific values tried for them and the default values of them. Combining this table with Figure~\ref{fig:addExp2_1} and Figure~\ref{fig:addExp2_2}, it can be found that the default settings are actually the suboptimal or even optimal ones for 13 in the 16 experiments. And for the rest three (\ie $eps$ for FGSM, PGD and Auto-Attack), the default settings are still reasonable instead of the worst. This is to confirm that using the default settings is not the reason why the benchmarks fail to outperform TSFool in Section~\ref{subsec:sm_result1} and Section~\ref{subsec:sm_result2}, and as a result, our comparison experiments are fair enough, sufficiently representative and of general significance.}
\label{tab:hyper_specific}
\end{table}

\begin{table*}[htbp]
\begin{center}
\renewcommand{\arraystretch}{1.1}
\resizebox{17.8cm}{!}{
\begin{tabular}{c|c|c|c|c|c|c|c|c|c}
\toprule
\multicolumn{2}{c|}{\textbf{Target Model}} & \multirow{2}{*}{\textbf{Method}} & \multirow{1.2}{*}{\textbf{Attack}} & \multirow{1.2}{*}{\textbf{Generation}} & \multirow{1.2}{*}{\textbf{Average}} & \multicolumn{3}{c|}{\textbf{Perturbation Ratio ($\rho^*$)}} & \multirow{1.2}{*}{\textbf{Camouflage}}\\
\cline{1-2} \cline{7-9} 
\multirow{1.2}{*}{\textbf{Dataset}} & \multirow{1.2}{*}{\textbf{Accuracy}} & & \multirow{0.8}{*}{\textbf{Success Rate}} & \multirow{0.8}{*}{\textbf{Number}} & \multirow{0.8}{*}{\textbf{Time Cost (s)}} & \multirow{1.2}{*}{\textbf{$\ell_1$}} & \multirow{1.2}{*}{\textbf{$\ell_2$}} & \multirow{1.2}{*}{\textbf{$\ell_{\infty}$}} & \multirow{0.8}{*}{\textbf{Coefficient}} \\
\midrule
\midrule
\multirow{9}{*}{AF} & \multirow{9}{*}{0.8000} & FGSM & 80.00\% & \multirow{8}{*}{15} & 0.0127 & 24.52\% & 25.87\% & 29.04\% & 1.0211 \\
& & JSMA & 86.67\% & & 7.4301 & 3.04\% & 3.74\% & 4.99\% & 0.8126 \\
& & DeepFool & 80.00\% & & 0.9900 & 0.54\% & 0.64\% & 0.85\% & 0.7551 \\
& & PGD (BIM) & 80.00\% & & 0.9406 & 16.79\% & 18.22\% & 21.72\% & 0.9918 \\
& & C\&W & 40.00\% & & 13.2858 & 0.48\% & 0.52\% & 0.63\% & 0.7958 \\
& & Auto-Attack & 80.00\% & & 2.1487 & 15.80\% & 17.11\% & 20.35\% & 0.9918 \\
& & Boundary Attack & 66.67\% & & 418.2511 & 0.43\% & 0.51\% & 0.67\% & 0.8045 \\
& & HopSkipJump & 86.67\% & & 78.3258 & 0.84\% & 0.97\% & 1.27\% & 0.8066 \\
& & TSFool & \cellcolor{SpringGreen}\underline{100.00\%} & 20 & 0.0960 & 5.89\% & 6.69\% & 8.65\% & \cellcolor{SpringGreen}\underline{0.6047} \\
\bottomrule
\end{tabular}
}
\end{center}
\vspace{0.2cm}
\caption{The table illustrates the performance of the experimental methods in the AF multivariate time series dataset from the UEA archive. Notice that Transfer Attack is not available here due to the lack of a relevant public adversarial set. It can be found that the performance of TSFool on multivariate time series is still consistent with our conclusion on univariate ones in Section~\ref{subsec:result1} and Section~\ref{subsec:sm_result1} (\ie optimal in ASR and CC, sub-optimal in Time Cost and still reasonable in Perturbation Ratio). \vspace{0.53cm}}
\label{tab:multi}
\end{table*}

\begin{figure*}[htb]
    \centering
    \begin{subfigure}{0.21\linewidth}
    \includegraphics[width=1.02\linewidth]{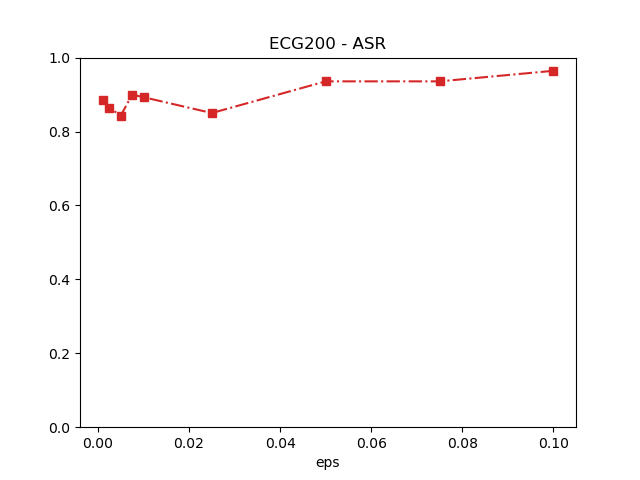}
    \end{subfigure}
    \begin{subfigure}{0.21\linewidth}
    \includegraphics[width=1.02\linewidth]{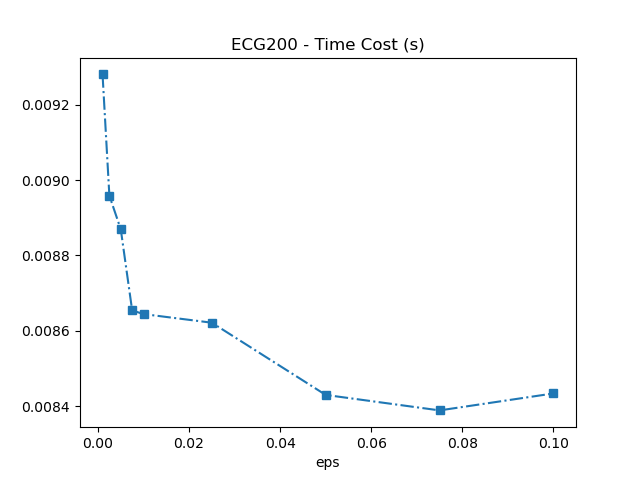}
    \end{subfigure}
    \begin{subfigure}{0.21\linewidth}
    \includegraphics[width=1.02\linewidth]{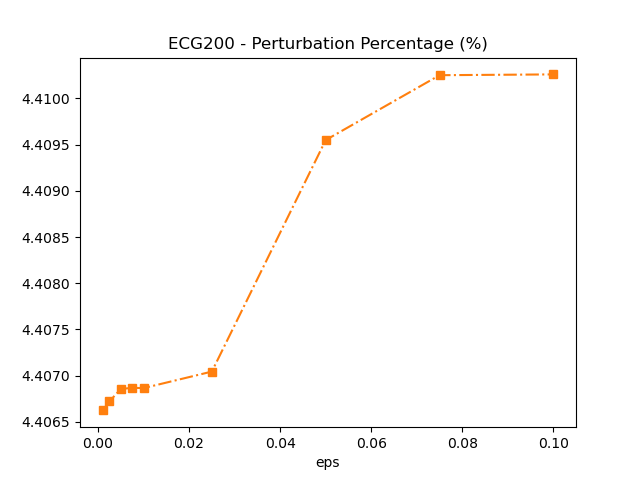}
    \end{subfigure}
    \begin{subfigure}{0.21\linewidth}
    \includegraphics[width=1.02\linewidth]{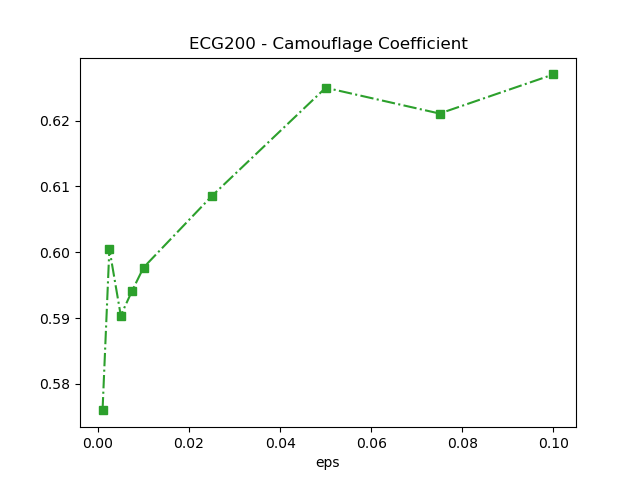}
    \end{subfigure}
    \begin{subfigure}{0.21\linewidth}
    \includegraphics[width=1.02\linewidth]{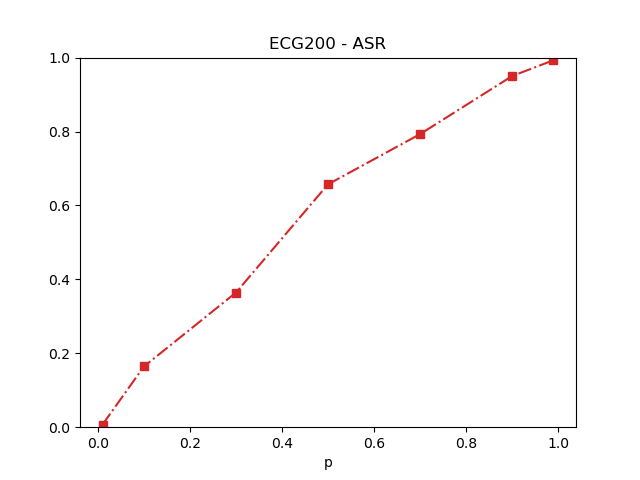}
    \end{subfigure}
    \centering
    \begin{subfigure}{0.21\linewidth}
    \includegraphics[width=1.02\linewidth]{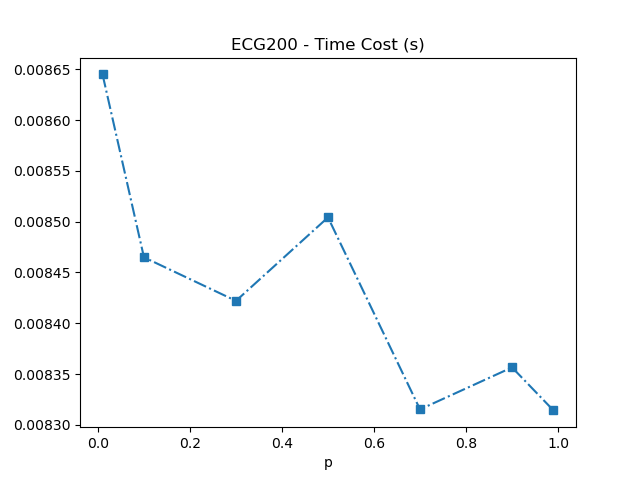}
    \end{subfigure}
    \begin{subfigure}{0.21\linewidth}
    \includegraphics[width=1.02\linewidth]{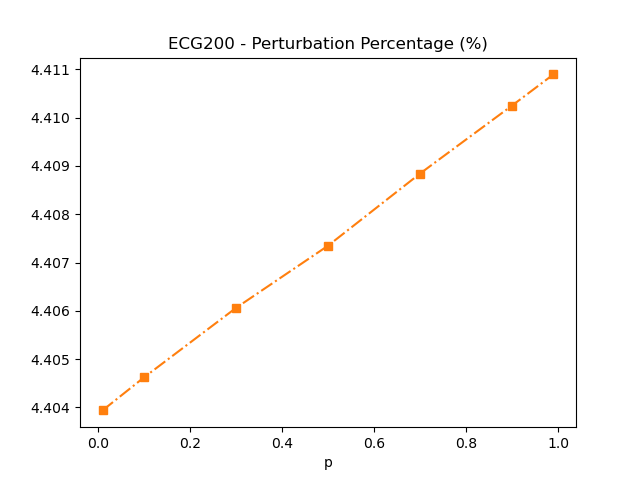}
    \end{subfigure}
    \begin{subfigure}{0.21\linewidth}
    \includegraphics[width=1.02\linewidth]{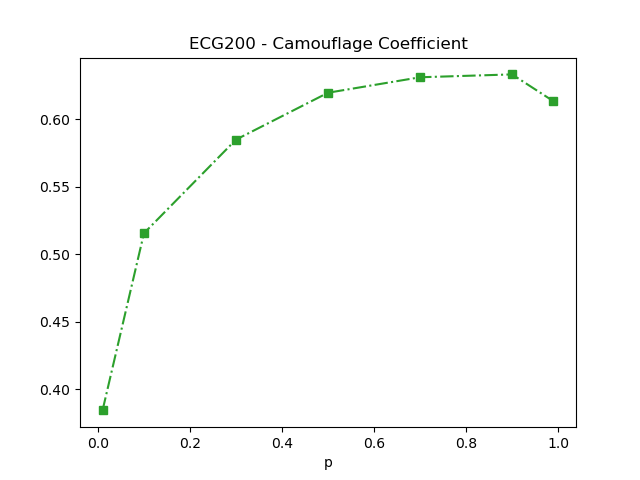}
    \end{subfigure}
    \vspace{0.45cm}
    \caption{The figures show the impact of the two hyper-parameters $eps$ and $P$ to the performance of TSFool. There are nine sample points for $eps$ (\ie 0, 0.0025, 0.005, 0.0075, 0.01, 0.025, 0.05, 0.075, 0.1) and seven for $P$ (\ie 0.01, 0.1, 0.3, 0.5, 0.7, 0.9, 0.99), both of which basically cover their whole value ranges. Notice that in our default setting used for the above experiments in Section~\ref{subsec:sm_result1} and Section~\ref{subsec:sm_result2}, the $eps$ is assigned to be 0.01 and the $P$ to be 0.9, both of which are reasonable but not the optimal. This is to verify that we do not artificially adopt the optimal hyper-parameters for TSFool to get any unfair advantages in the comparisons. Please also notice that to better reveal the trend of the measures, the y-axises of them except ASR are not normalized here. \vspace{0.6cm}}
    \label{fig:addExp1}
\end{figure*}

\begin{figure*}[htb]
    \begin{subfigure}{0.121\linewidth}
    \includegraphics[width=1.05\linewidth]{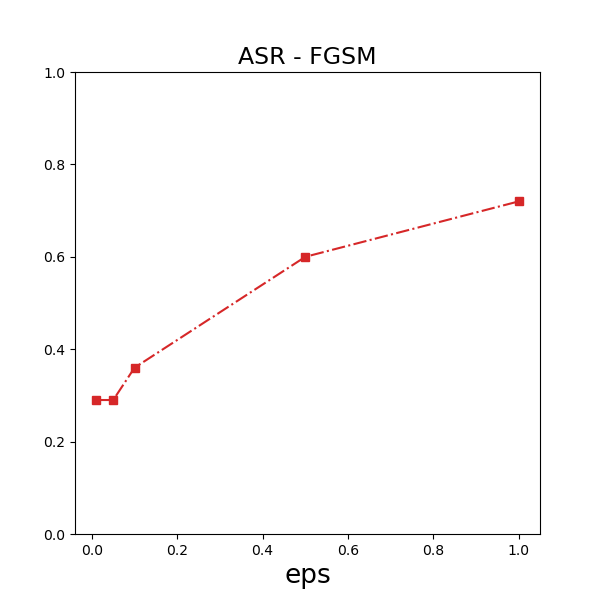}
    \end{subfigure}
    \begin{subfigure}{0.121\linewidth}
    \includegraphics[width=1.05\linewidth]{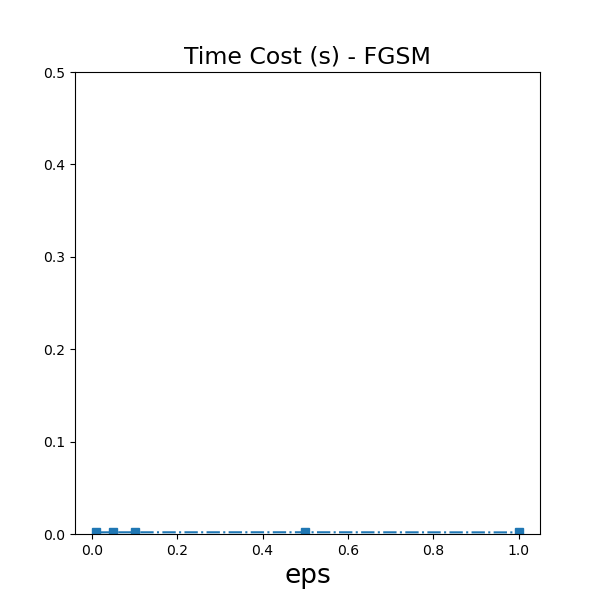}
    \end{subfigure}
    \begin{subfigure}{0.121\linewidth}
    \includegraphics[width=1.05\linewidth]{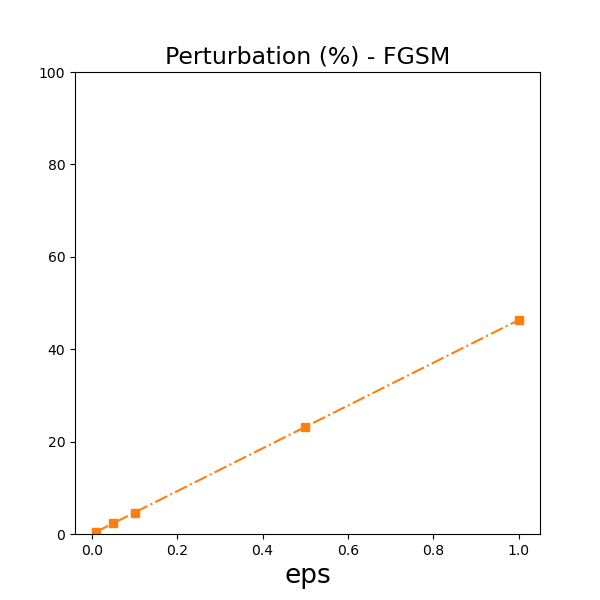}
    \end{subfigure}
    \begin{subfigure}{0.121\linewidth}
    \includegraphics[width=1.05\linewidth]{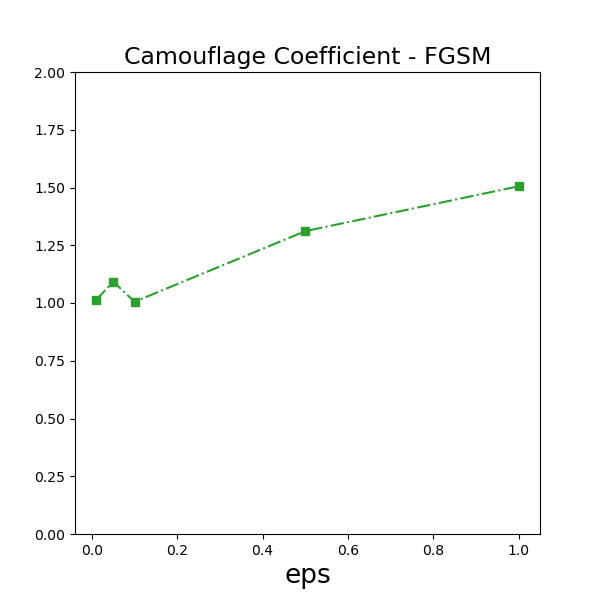}
    \end{subfigure}
    \begin{subfigure}{0.121\linewidth}
    \includegraphics[width=1.05\linewidth]{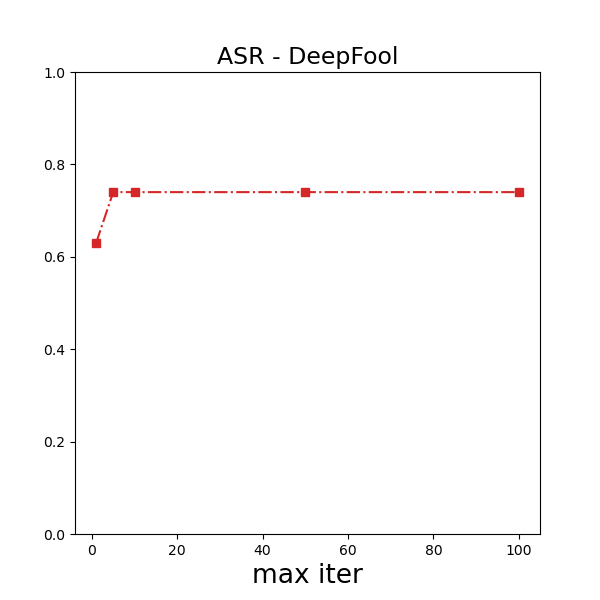}
    \end{subfigure}
    \begin{subfigure}{0.121\linewidth}
    \includegraphics[width=1.05\linewidth]{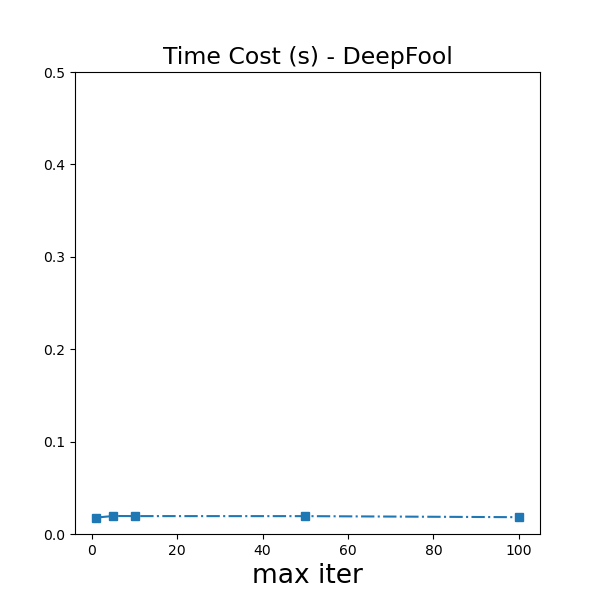}
    \end{subfigure}
    \begin{subfigure}{0.121\linewidth}
    \includegraphics[width=1.05\linewidth]{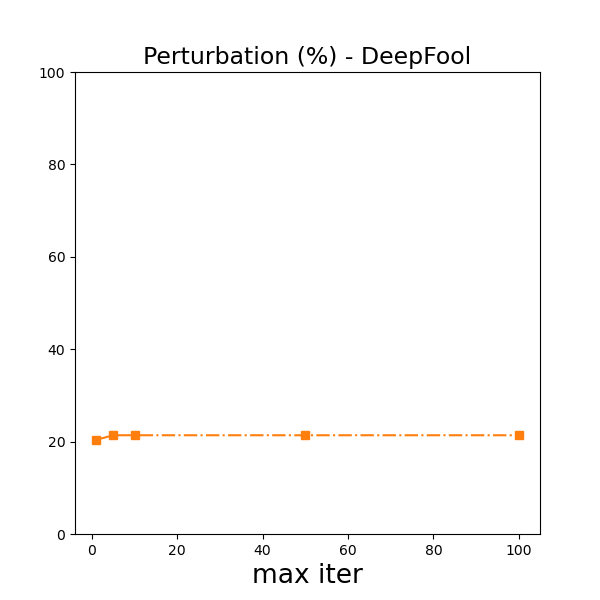}
    \end{subfigure}
    \begin{subfigure}{0.121\linewidth}
    \includegraphics[width=1.05\linewidth]{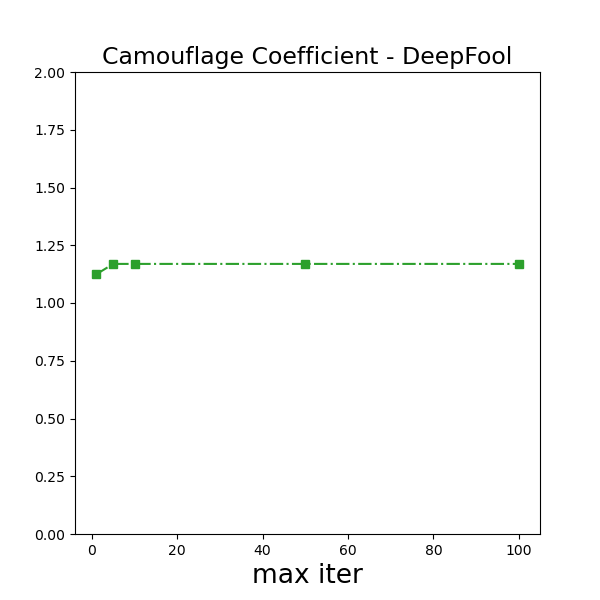}
    \end{subfigure}
    \begin{subfigure}{0.121\linewidth}
    \includegraphics[width=1.05\linewidth]{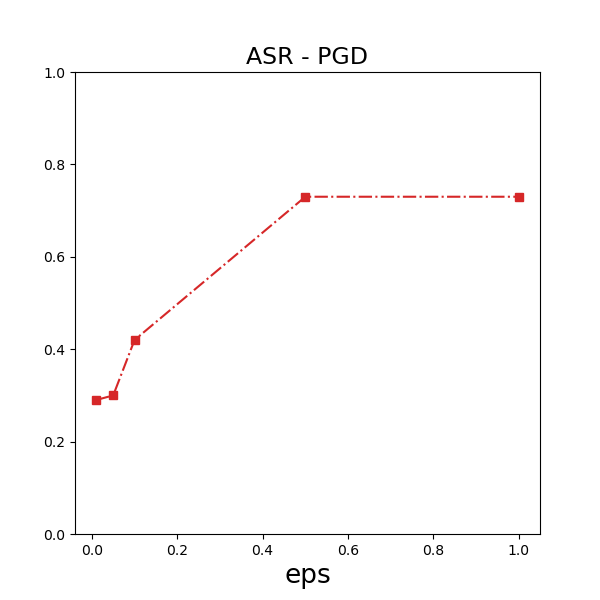}
    \end{subfigure}
    \begin{subfigure}{0.121\linewidth}
    \includegraphics[width=1.05\linewidth]{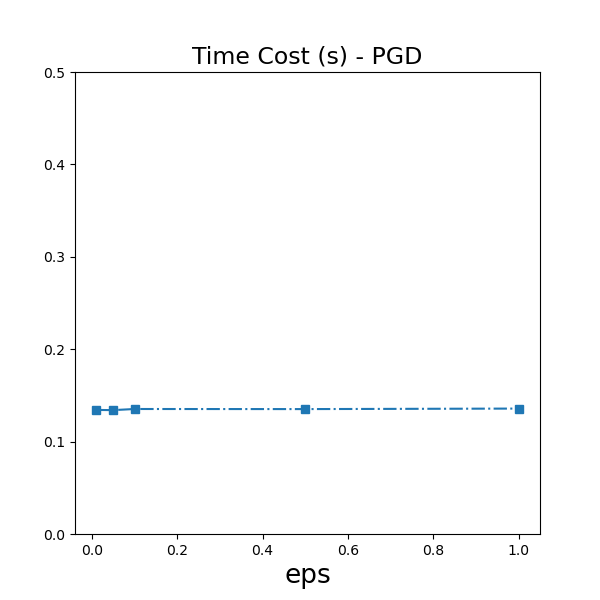}
    \end{subfigure}
    \begin{subfigure}{0.121\linewidth}
    \includegraphics[width=1.05\linewidth]{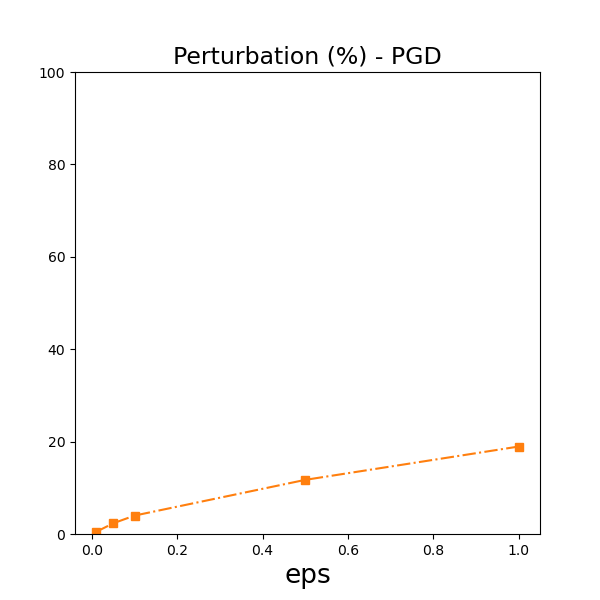}
    \end{subfigure}
    \begin{subfigure}{0.121\linewidth}
    \includegraphics[width=1.05\linewidth]{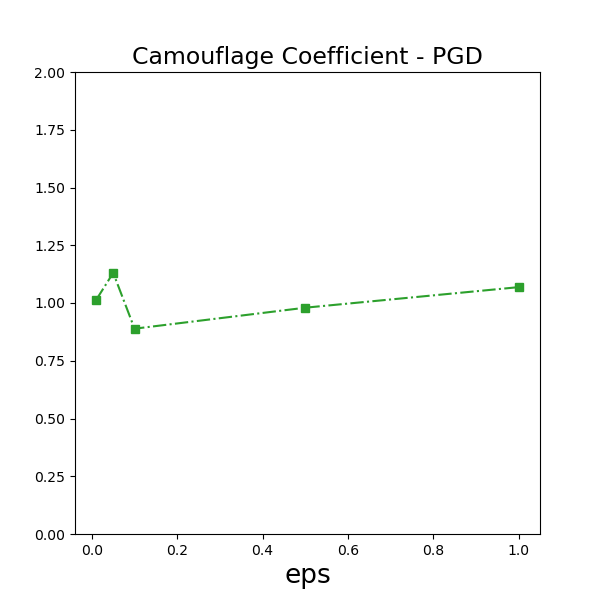}
    \end{subfigure}
    \begin{subfigure}{0.121\linewidth}
    \includegraphics[width=1.05\linewidth]{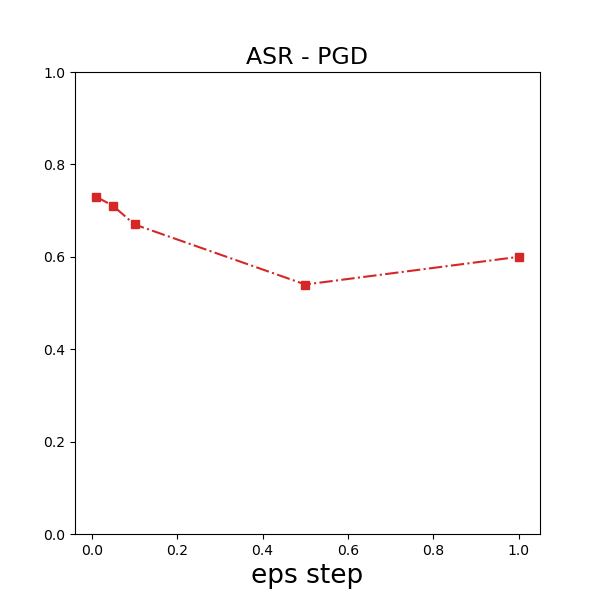}
    \end{subfigure}
    \begin{subfigure}{0.121\linewidth}
    \includegraphics[width=1.05\linewidth]{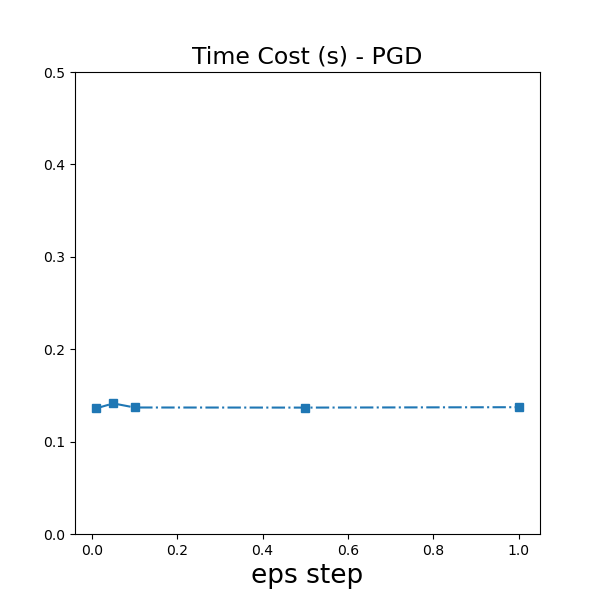}
    \end{subfigure}
    \begin{subfigure}{0.121\linewidth}
    \includegraphics[width=1.05\linewidth]{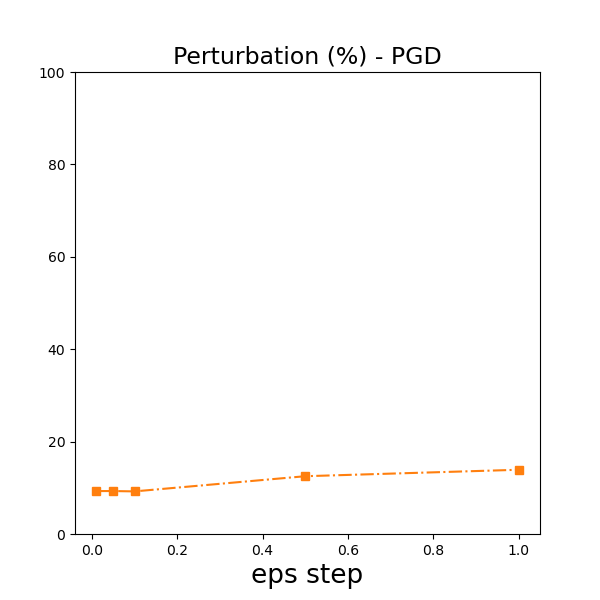}
    \end{subfigure}
    \begin{subfigure}{0.121\linewidth}
    \includegraphics[width=1.05\linewidth]{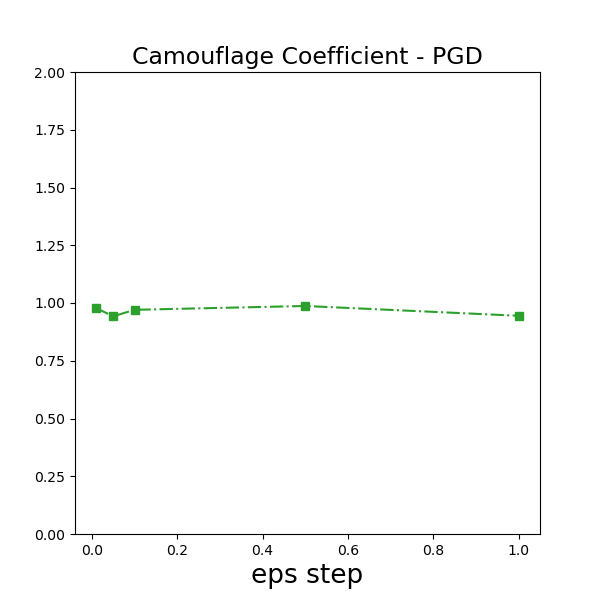}
    \end{subfigure}
    \begin{subfigure}{0.121\linewidth}
    \includegraphics[width=1.05\linewidth]{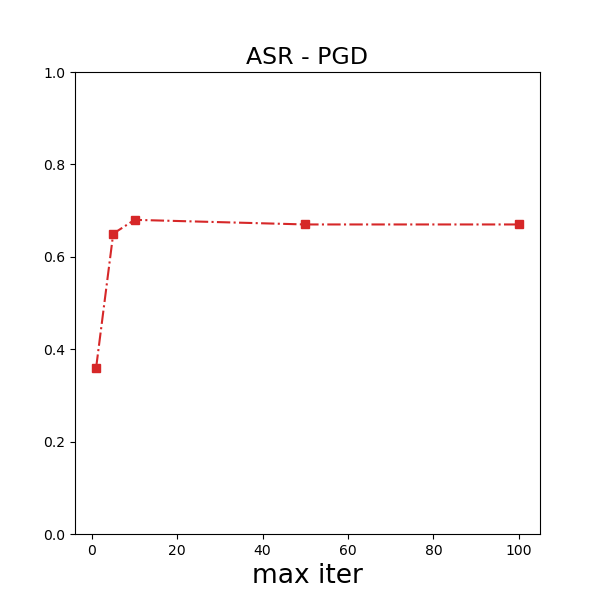}
    \end{subfigure}
    \begin{subfigure}{0.121\linewidth}
    \includegraphics[width=1.05\linewidth]{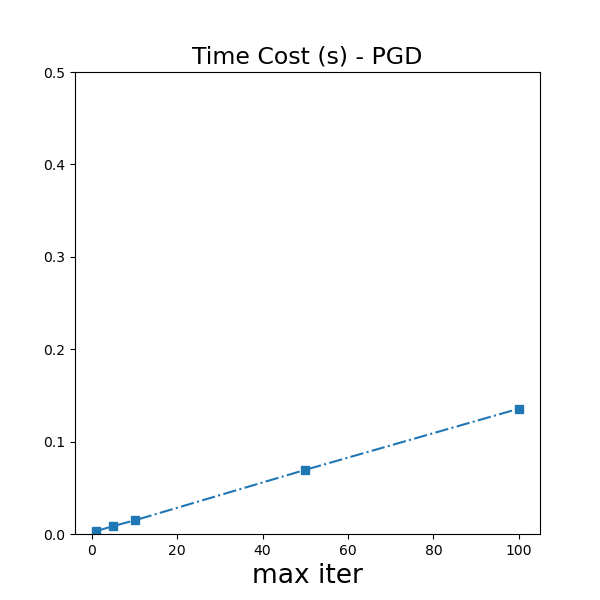}
    \end{subfigure}
    \begin{subfigure}{0.121\linewidth}
    \includegraphics[width=1.05\linewidth]{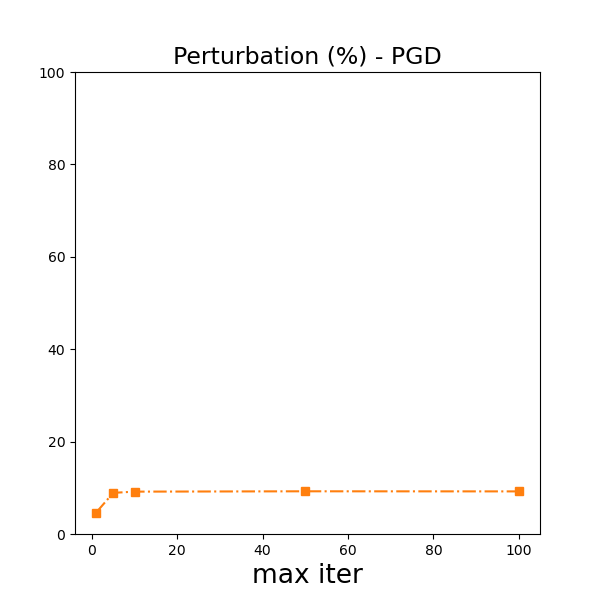}
    \end{subfigure}
    \begin{subfigure}{0.121\linewidth}
    \includegraphics[width=1.05\linewidth]{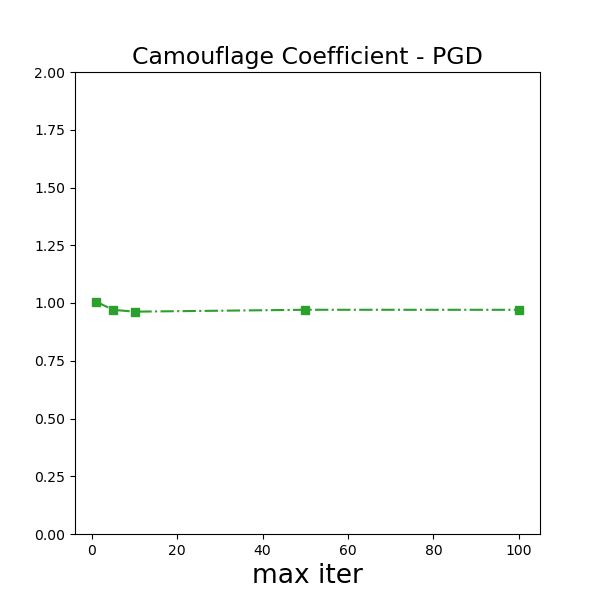}
    \end{subfigure}
    \begin{subfigure}{0.121\linewidth}
    \includegraphics[width=1.05\linewidth]{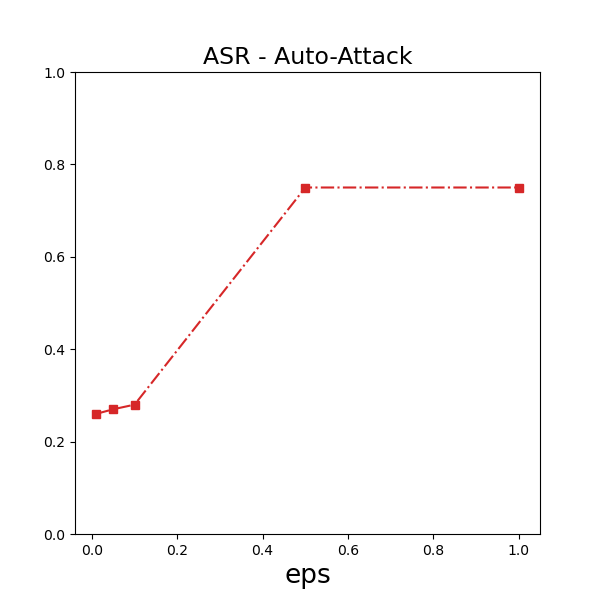}
    \end{subfigure}
    \begin{subfigure}{0.121\linewidth}
    \includegraphics[width=1.05\linewidth]{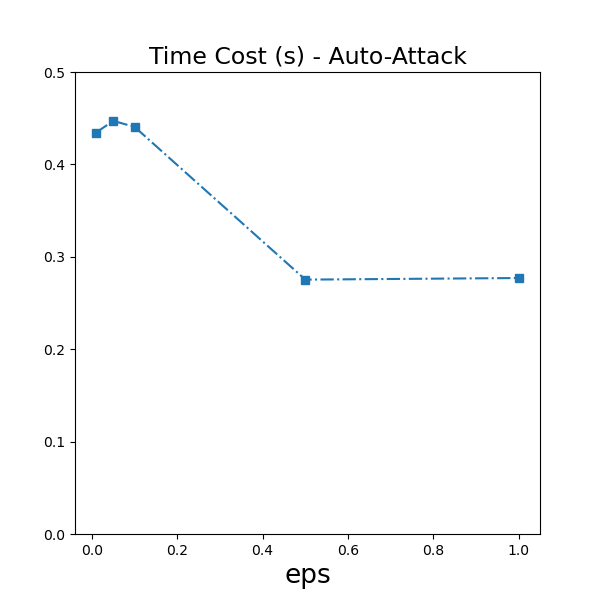}
    \end{subfigure}
    \begin{subfigure}{0.121\linewidth}
    \includegraphics[width=1.05\linewidth]{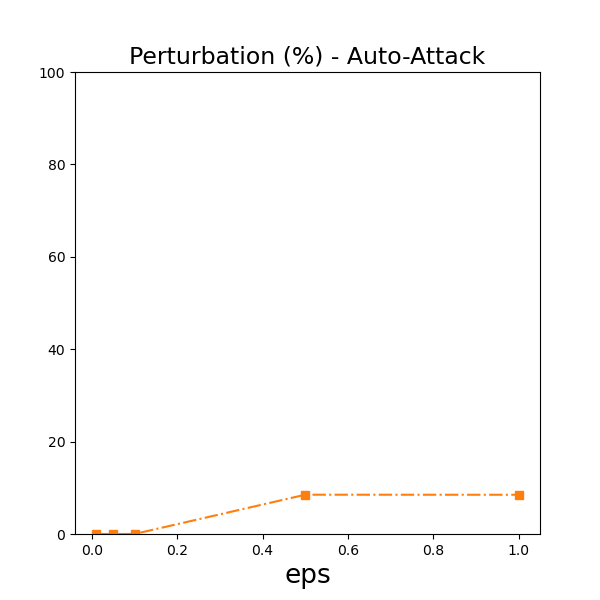}
    \end{subfigure}
    \begin{subfigure}{0.121\linewidth}
    \includegraphics[width=1.05\linewidth]{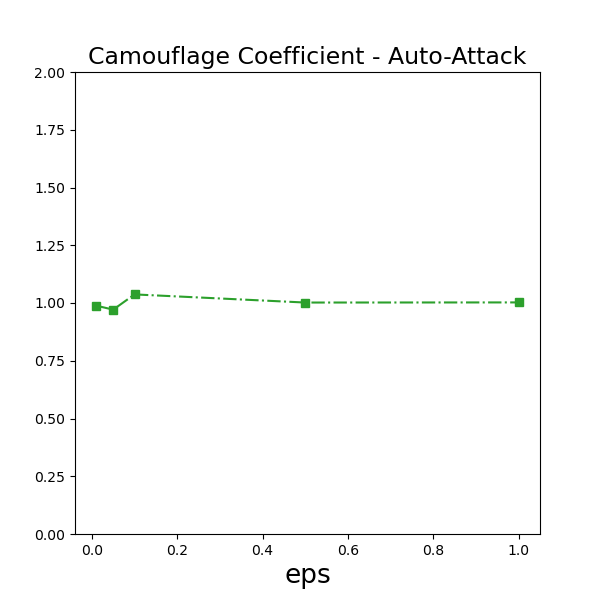}
    \end{subfigure}
\vspace{0.15cm}
\caption{With every four figures corresponding to one hyper-parameter for a specific method, the figures show the impact of some important hyper-parameters on the performance of some benchmarks, including the $eps$ for FGSM, PGD (BIM) and Auto-Attack, the $max$ $iter$ for DeepFool and PGD (BIM), as well as the $eps$ $step$ for PGD (BIM). There are five sample points for all of these hyper-parameters and their specific values tried are listed in Table~\ref{tab:hyper_specific}. Notice that for convenience of comparison, the y-axises of ASR, time cost, $\rho^*$ and $\mathcal{C}$ are respectively normalized to [0, 1], [0, 0.5](s), [0, 100](\%) and [0, 2] here. \vspace{0.6cm}}
\label{fig:addExp2_1}
\end{figure*}

\begin{figure*}[htb]
    \begin{subfigure}{0.121\linewidth}
    \includegraphics[width=1.05\linewidth]{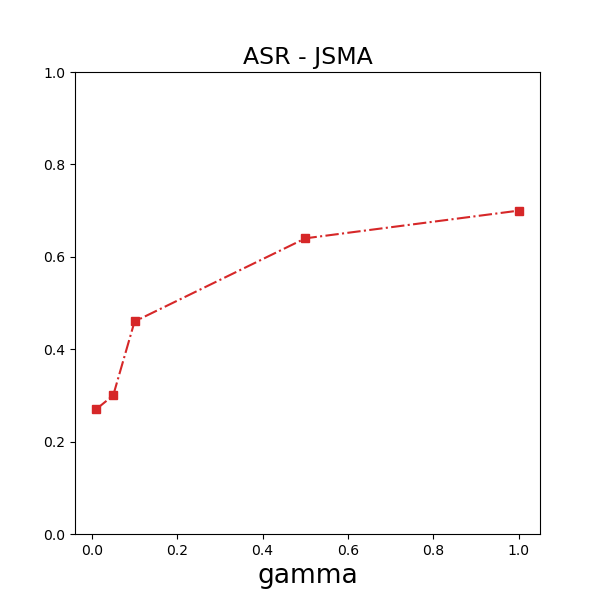}
    \end{subfigure}
    \begin{subfigure}{0.121\linewidth}
    \includegraphics[width=1.05\linewidth]{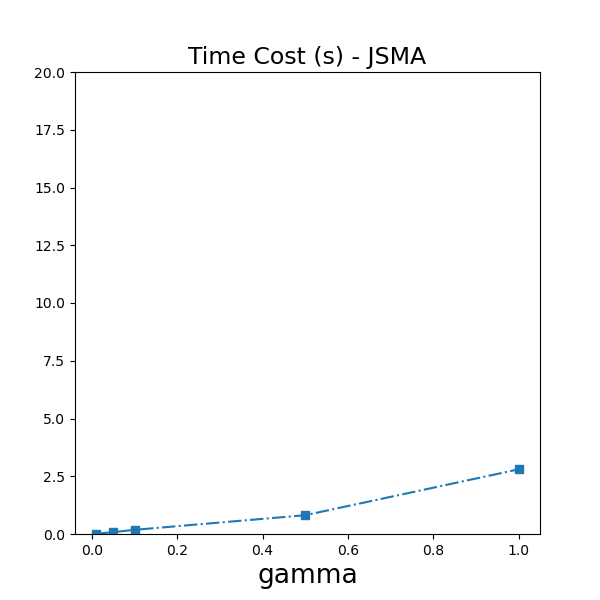}
    \end{subfigure}
    \begin{subfigure}{0.121\linewidth}
    \includegraphics[width=1.05\linewidth]{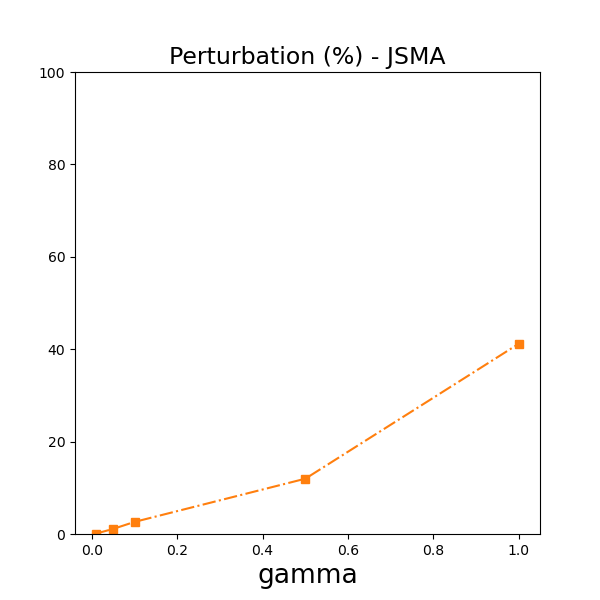}
    \end{subfigure}
    \begin{subfigure}{0.121\linewidth}
    \includegraphics[width=1.05\linewidth]{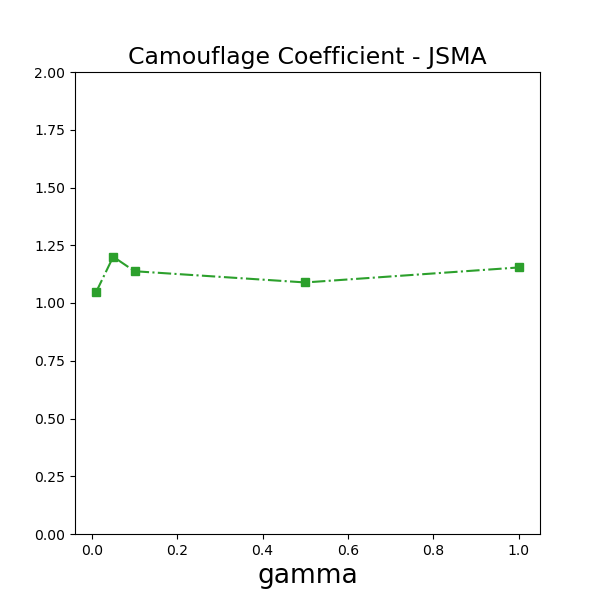}
    \end{subfigure}
    \begin{subfigure}{0.121\linewidth}
    \includegraphics[width=1.05\linewidth]{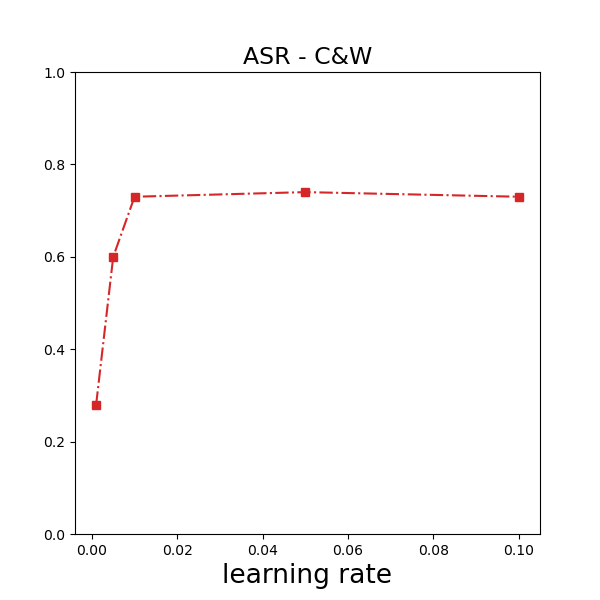}
    \end{subfigure}
    \begin{subfigure}{0.121\linewidth}
    \includegraphics[width=1.05\linewidth]{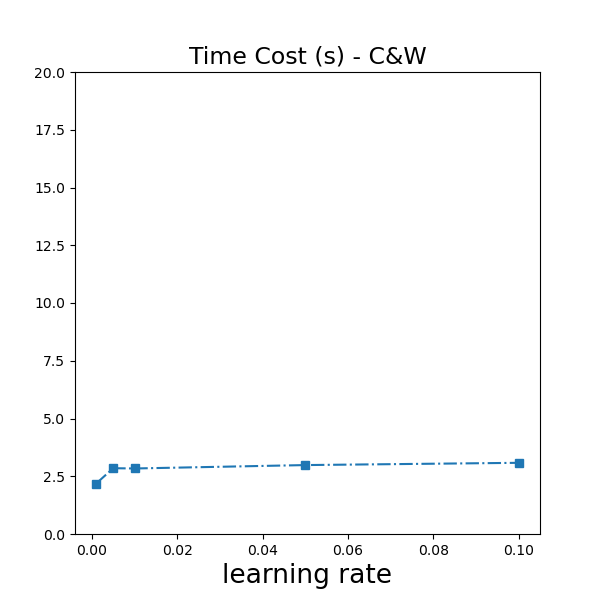}
    \end{subfigure}
    \begin{subfigure}{0.121\linewidth}
    \includegraphics[width=1.05\linewidth]{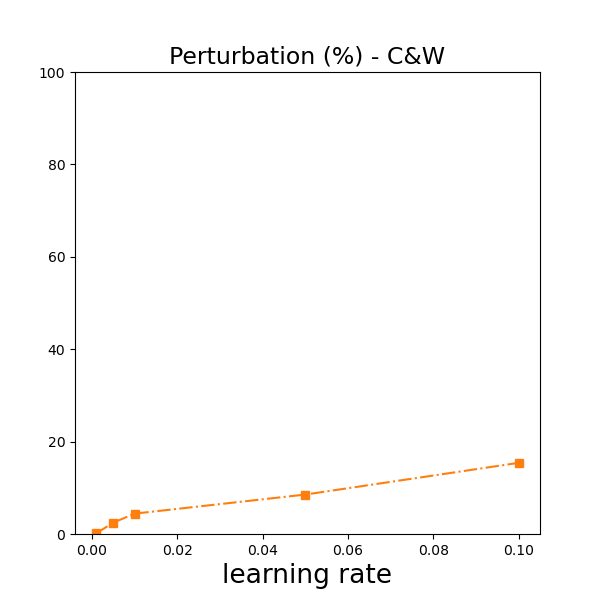}
    \end{subfigure}
    \begin{subfigure}{0.121\linewidth}
    \includegraphics[width=1.05\linewidth]{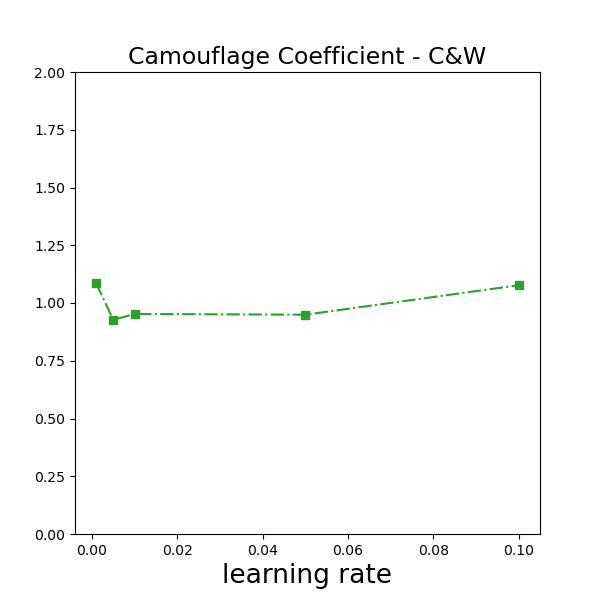}
    \end{subfigure}
    \begin{subfigure}{0.121\linewidth}
    \includegraphics[width=1.05\linewidth]{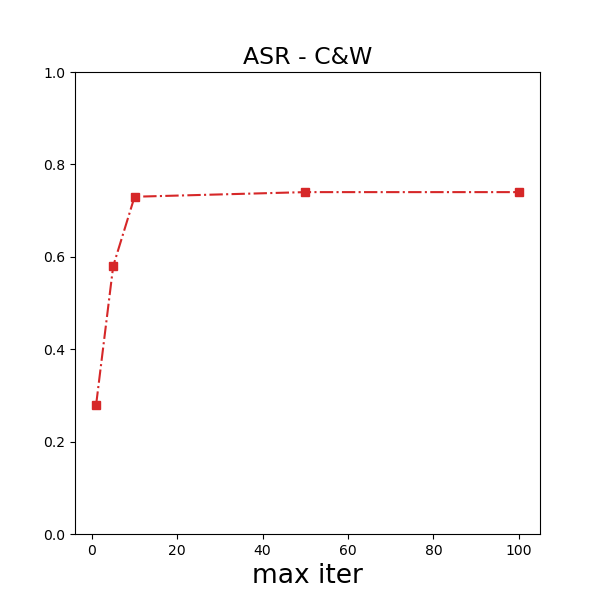}
    \end{subfigure}
    \begin{subfigure}{0.121\linewidth}
    \includegraphics[width=1.05\linewidth]{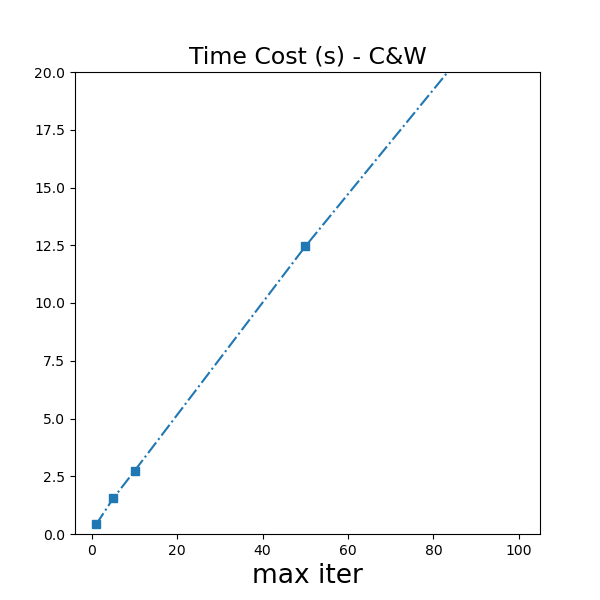}
    \end{subfigure}
    \begin{subfigure}{0.121\linewidth}
    \includegraphics[width=1.05\linewidth]{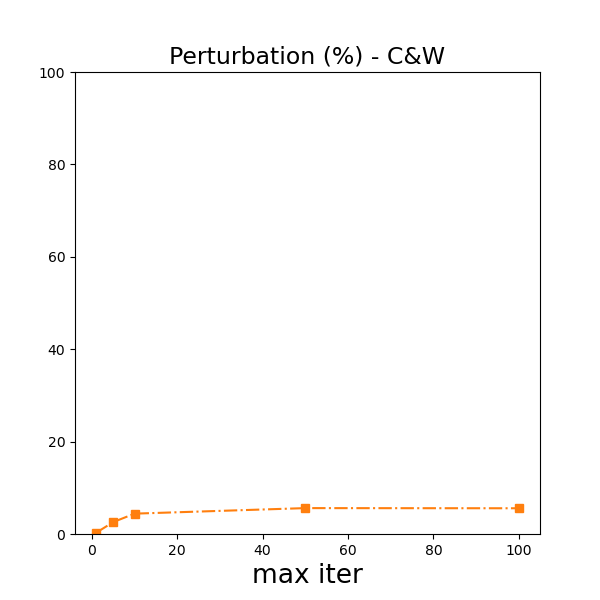}
    \end{subfigure}
    \begin{subfigure}{0.121\linewidth}
    \includegraphics[width=1.05\linewidth]{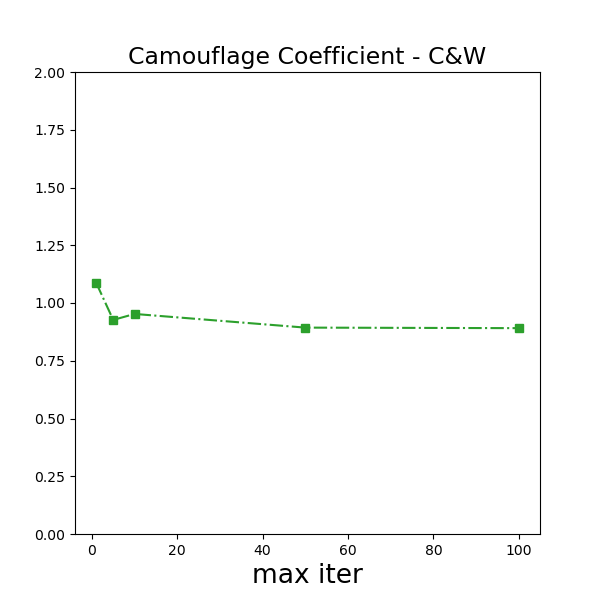}
    \end{subfigure}
    \begin{subfigure}{0.121\linewidth}
    \includegraphics[width=1.05\linewidth]{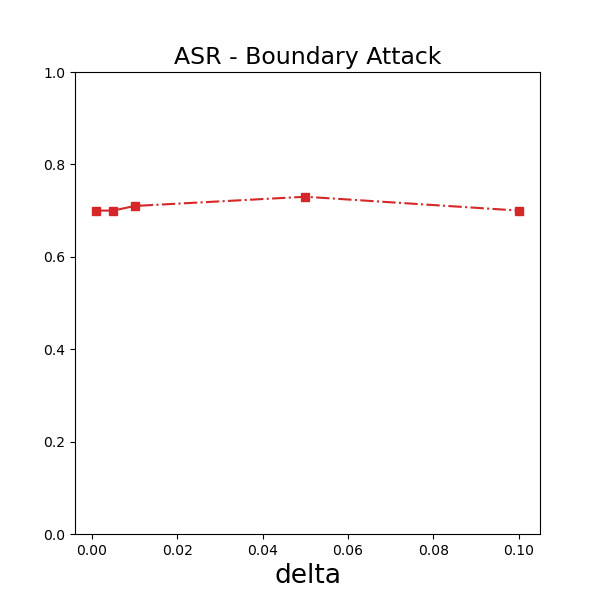}
    \end{subfigure}
    \begin{subfigure}{0.121\linewidth}
    \includegraphics[width=1.05\linewidth]{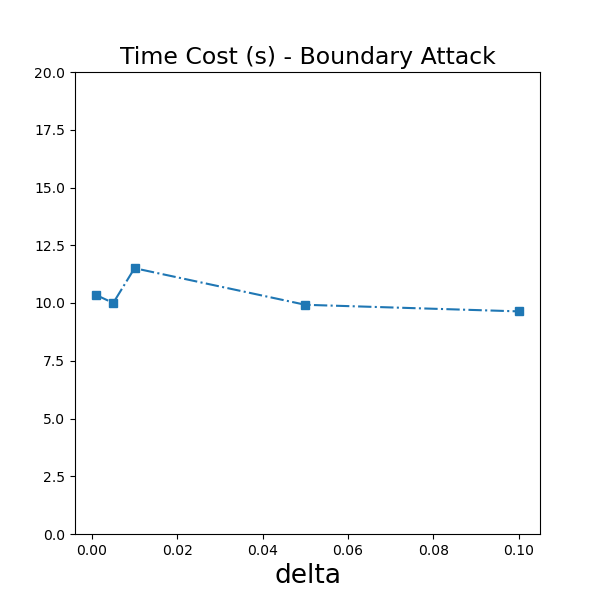}
    \end{subfigure}
    \begin{subfigure}{0.121\linewidth}
    \includegraphics[width=1.05\linewidth]{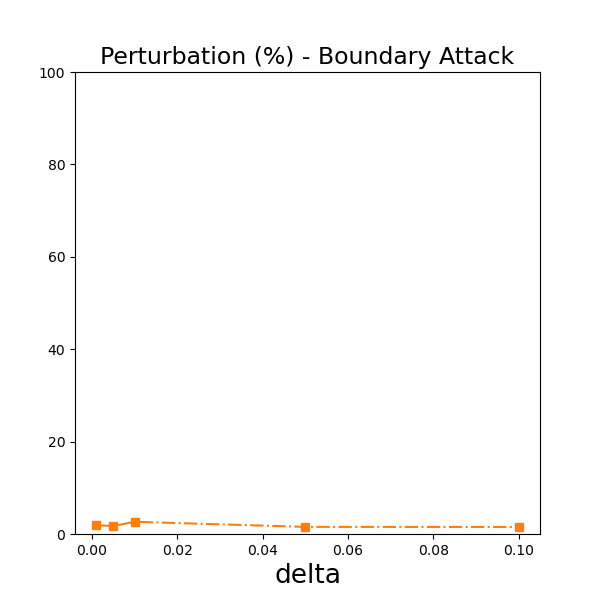}
    \end{subfigure}
    \begin{subfigure}{0.121\linewidth}
    \includegraphics[width=1.05\linewidth]{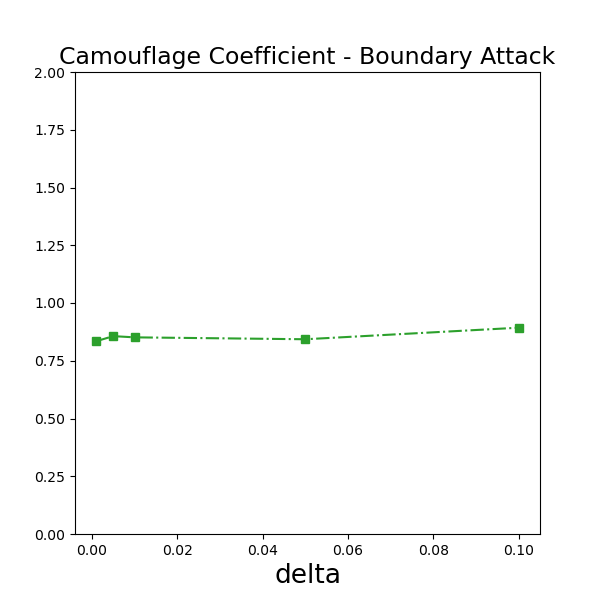}
    \end{subfigure}
    \begin{subfigure}{0.121\linewidth}
    \includegraphics[width=1.05\linewidth]{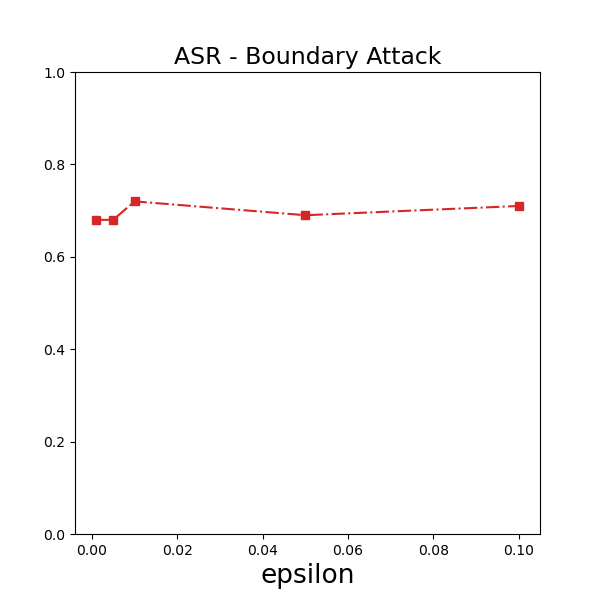}
    \end{subfigure}
    \begin{subfigure}{0.121\linewidth}
    \includegraphics[width=1.05\linewidth]{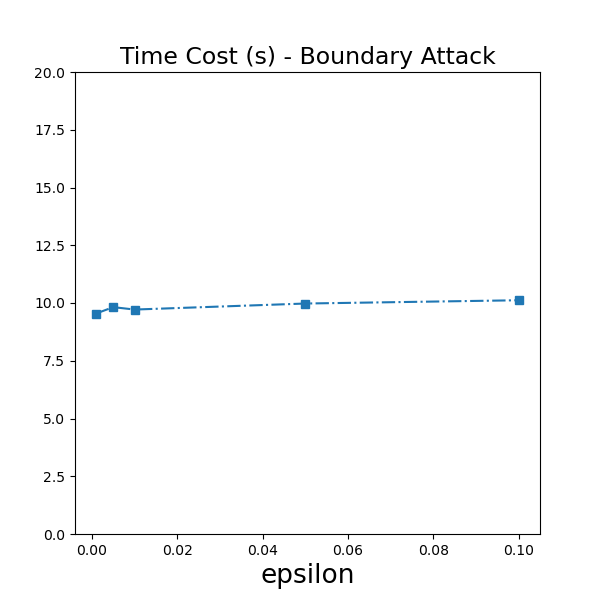}
    \end{subfigure}
    \begin{subfigure}{0.121\linewidth}
    \includegraphics[width=1.05\linewidth]{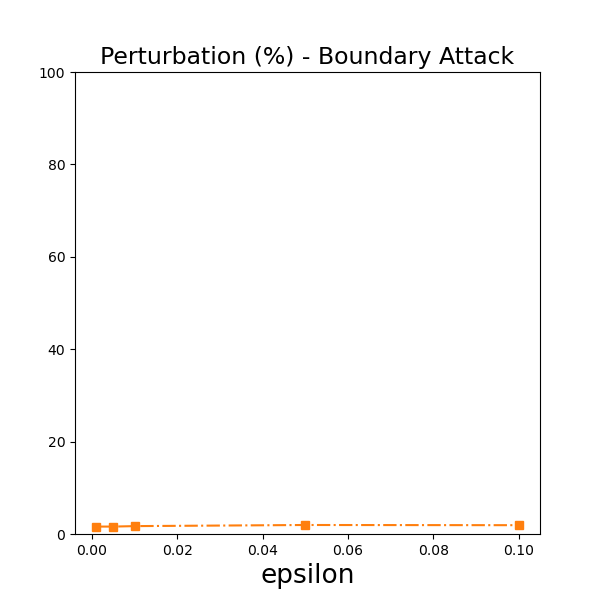}
    \end{subfigure}
    \begin{subfigure}{0.121\linewidth}
    \includegraphics[width=1.05\linewidth]{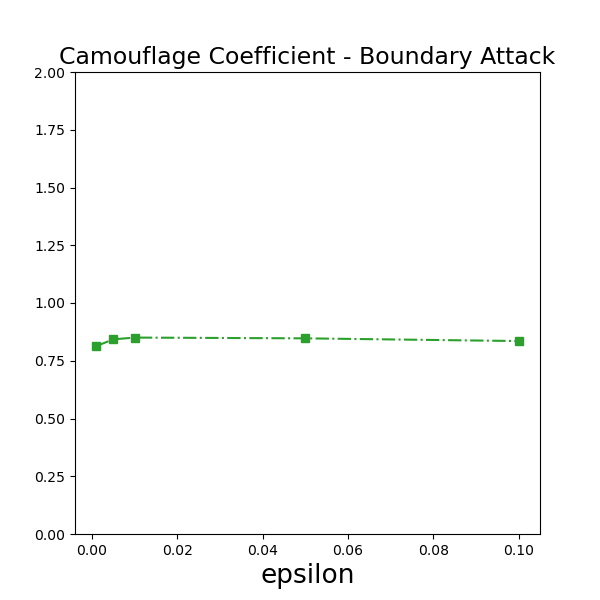}
    \end{subfigure}
    \begin{subfigure}{0.121\linewidth}
    \includegraphics[width=1.05\linewidth]{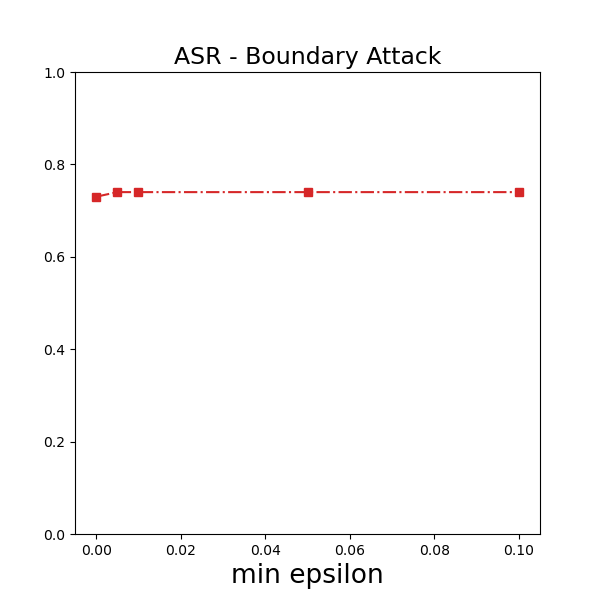}
    \end{subfigure}
    \begin{subfigure}{0.121\linewidth}
    \includegraphics[width=1.05\linewidth]{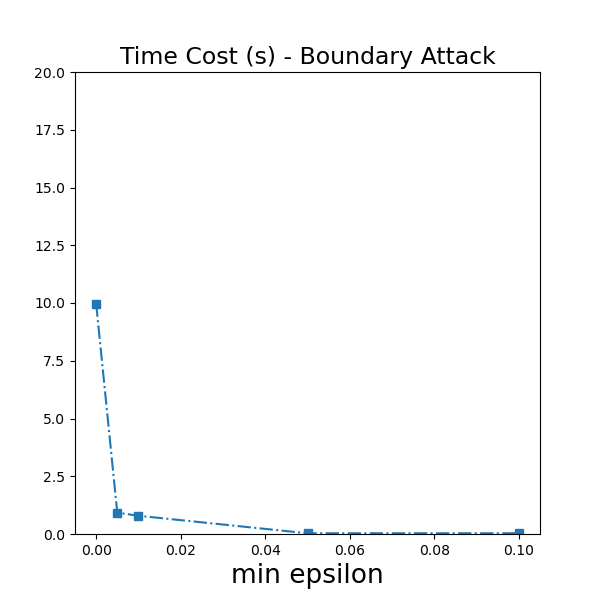}
    \end{subfigure}
    \begin{subfigure}{0.121\linewidth}
    \includegraphics[width=1.05\linewidth]{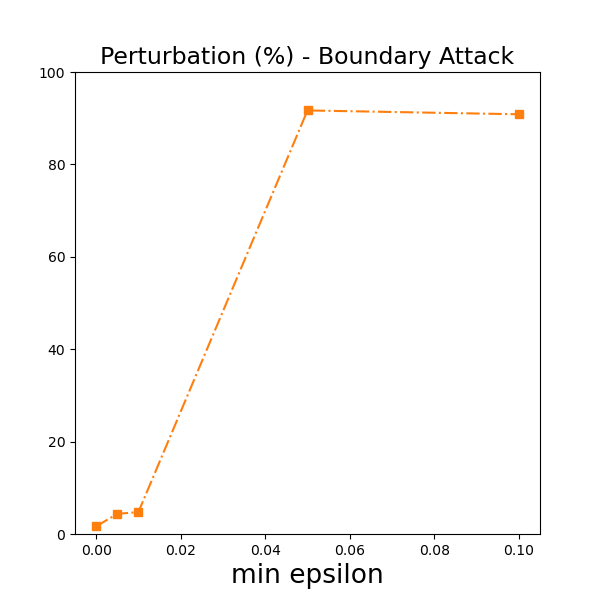}
    \end{subfigure}
    \begin{subfigure}{0.121\linewidth}
    \includegraphics[width=1.05\linewidth]{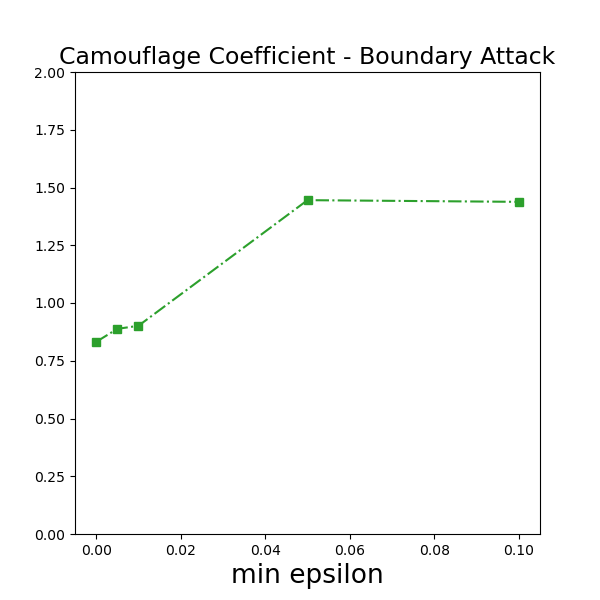}
    \end{subfigure}
    \begin{subfigure}{0.121\linewidth}
    \includegraphics[width=1.05\linewidth]{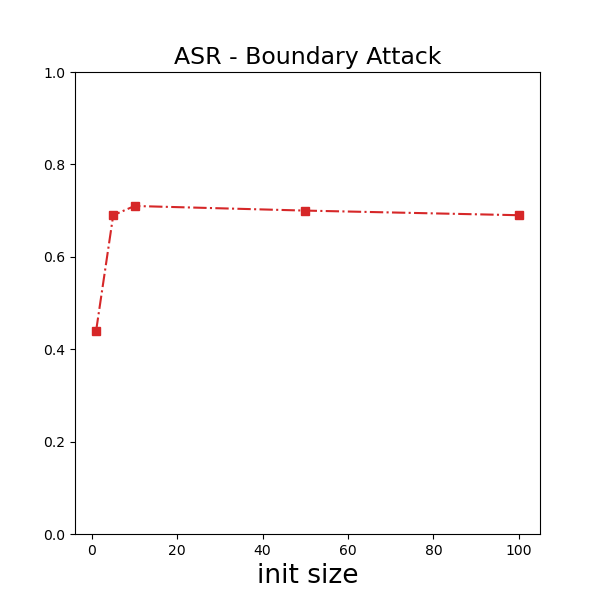}
    \end{subfigure}
    \begin{subfigure}{0.121\linewidth}
    \includegraphics[width=1.05\linewidth]{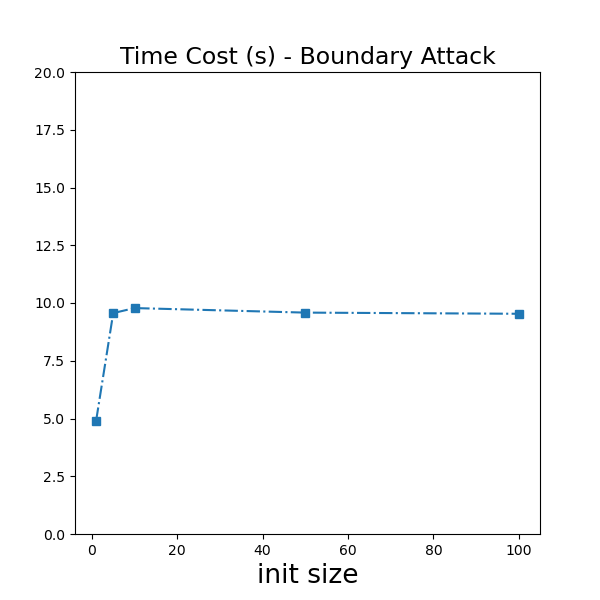}
    \end{subfigure}
    \begin{subfigure}{0.121\linewidth}
    \includegraphics[width=1.05\linewidth]{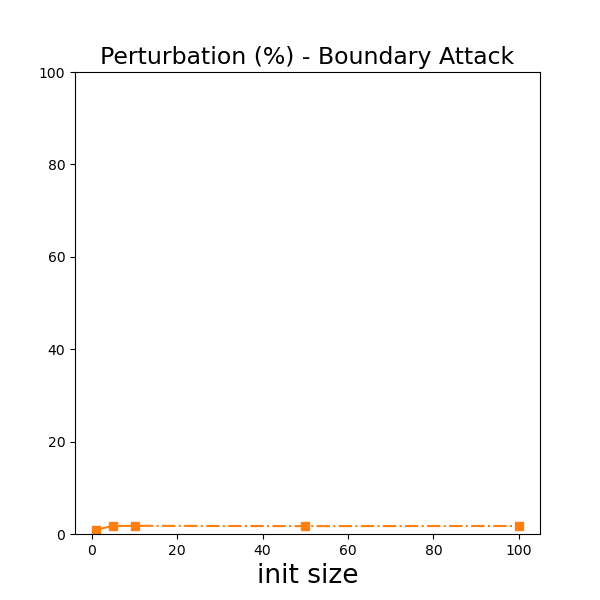}
    \end{subfigure}
    \begin{subfigure}{0.121\linewidth}
    \includegraphics[width=1.05\linewidth]{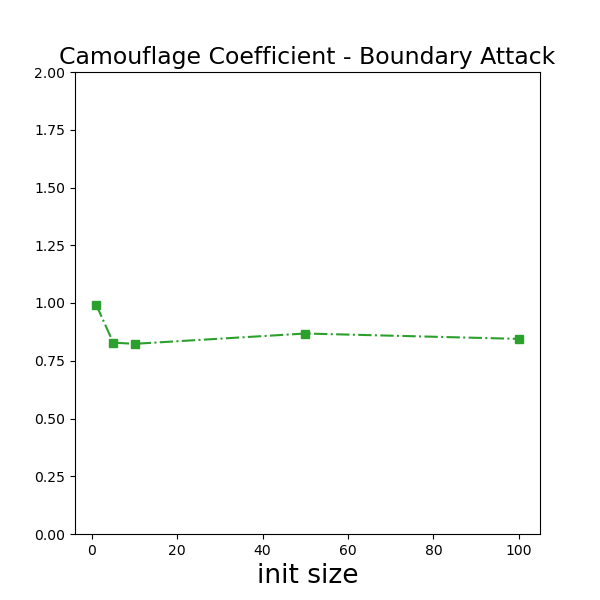}
    \end{subfigure}
    \begin{subfigure}{0.121\linewidth}
    \includegraphics[width=1.05\linewidth]{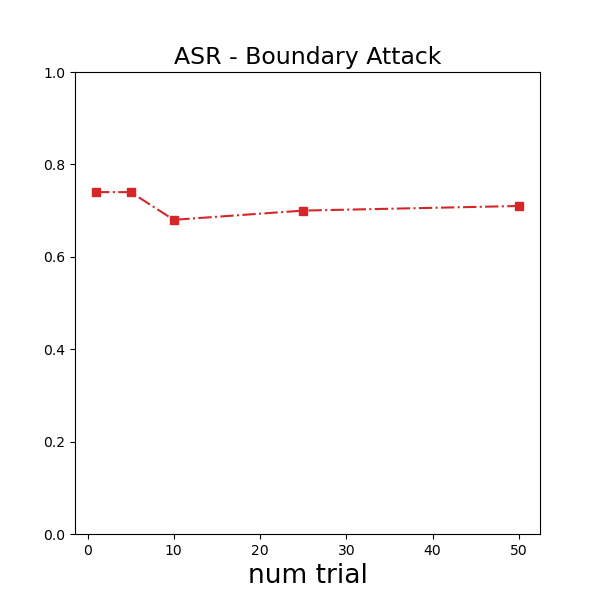}
    \end{subfigure}
    \begin{subfigure}{0.121\linewidth}
    \includegraphics[width=1.05\linewidth]{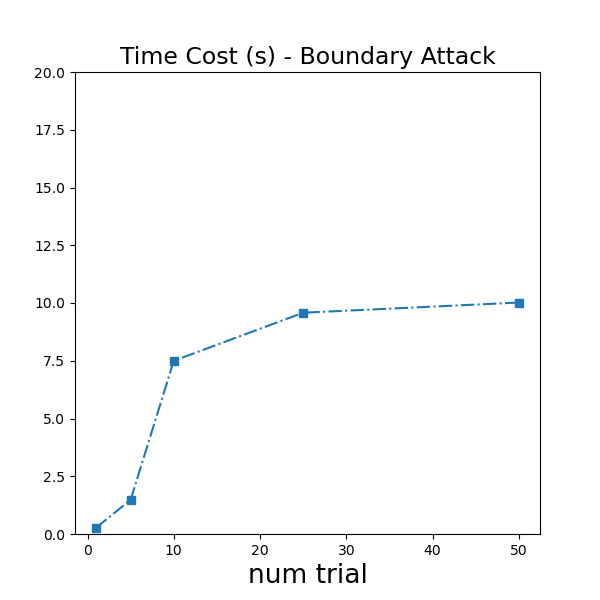}
    \end{subfigure}
    \begin{subfigure}{0.121\linewidth}
    \includegraphics[width=1.05\linewidth]{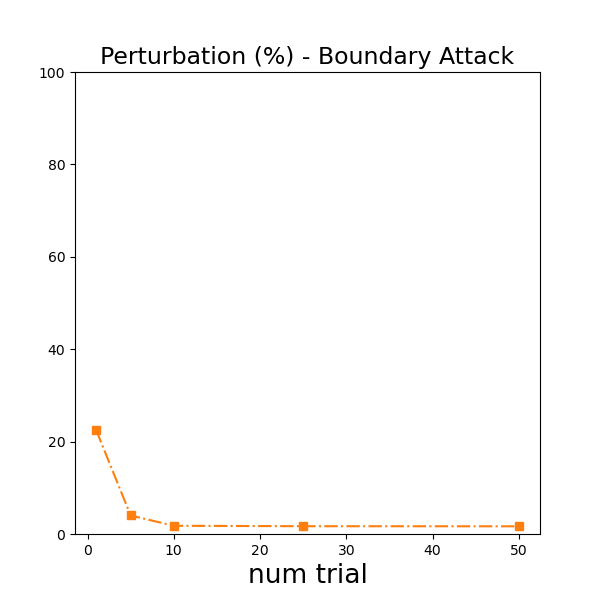}
    \end{subfigure}
    \begin{subfigure}{0.121\linewidth}
    \includegraphics[width=1.05\linewidth]{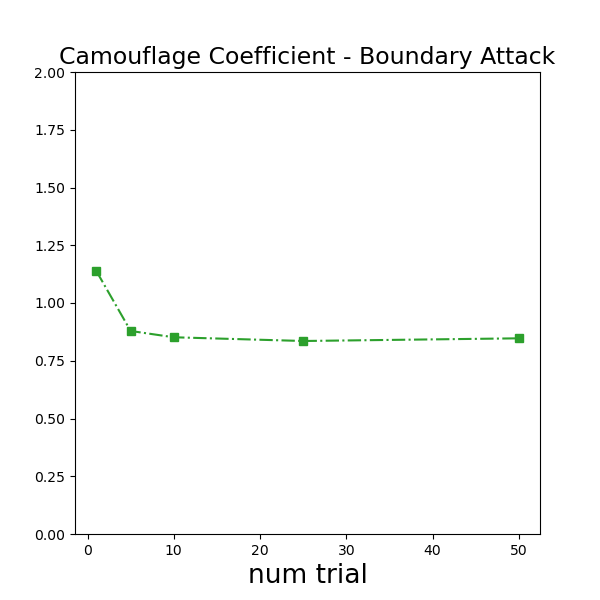}
    \end{subfigure}
    \begin{subfigure}{0.121\linewidth}
    \includegraphics[width=1.05\linewidth]{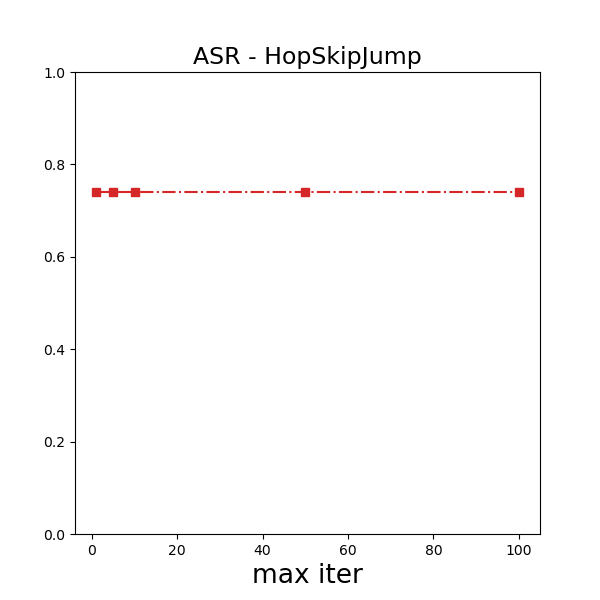}
    \end{subfigure}
    \begin{subfigure}{0.121\linewidth}
    \includegraphics[width=1.05\linewidth]{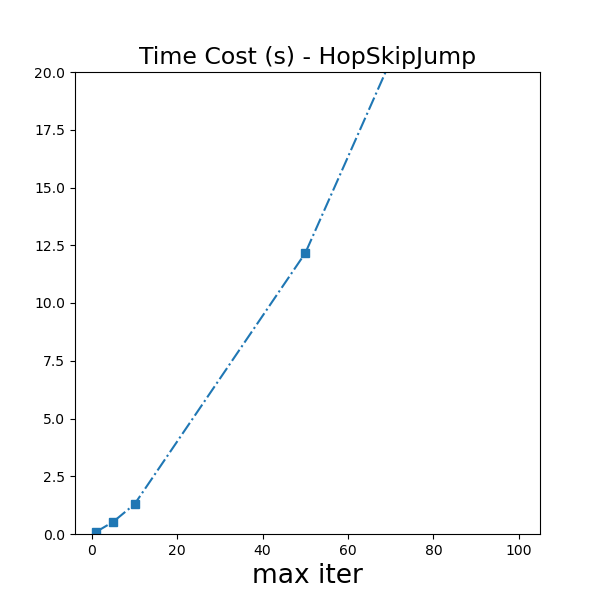}
    \end{subfigure}
    \begin{subfigure}{0.121\linewidth}
    \includegraphics[width=1.05\linewidth]{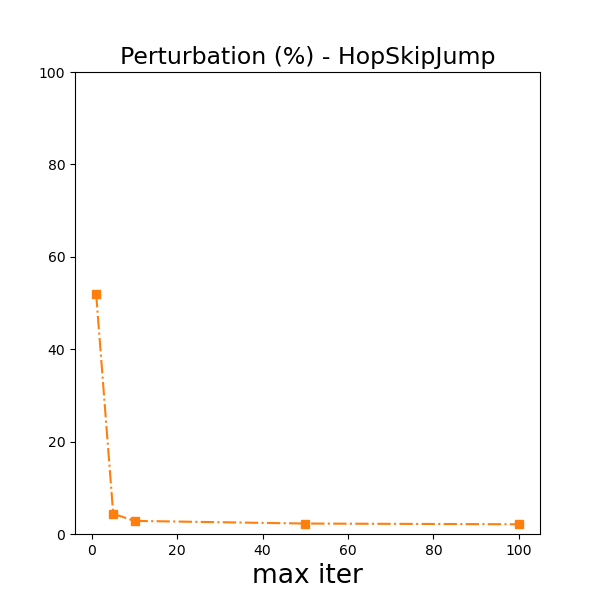}
    \end{subfigure}
    \begin{subfigure}{0.121\linewidth}
    \includegraphics[width=1.05\linewidth]{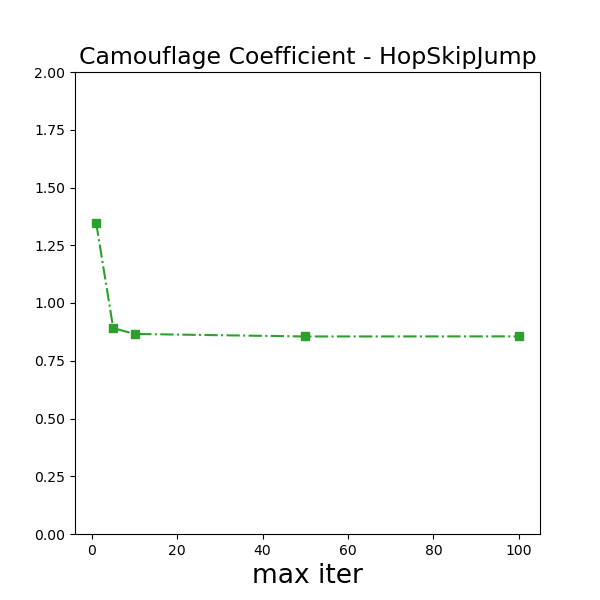}
    \end{subfigure}
    \begin{subfigure}{0.121\linewidth}
    \includegraphics[width=1.05\linewidth]{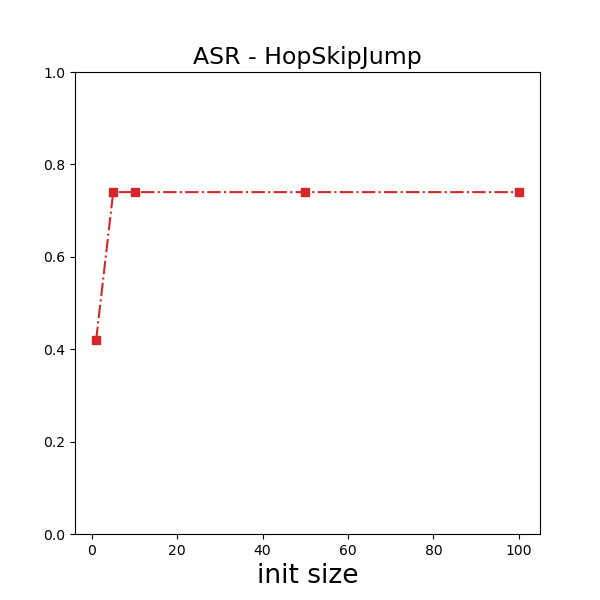}
    \end{subfigure}
    \begin{subfigure}{0.121\linewidth}
    \includegraphics[width=1.05\linewidth]{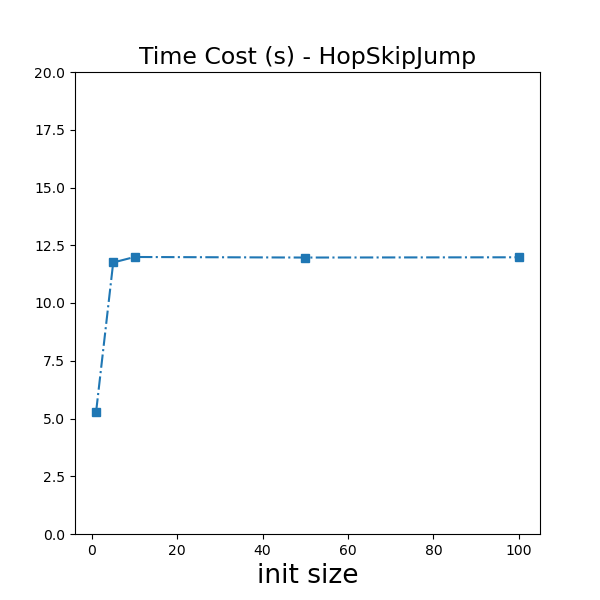}
    \end{subfigure}
    \begin{subfigure}{0.121\linewidth}
    \includegraphics[width=1.05\linewidth]{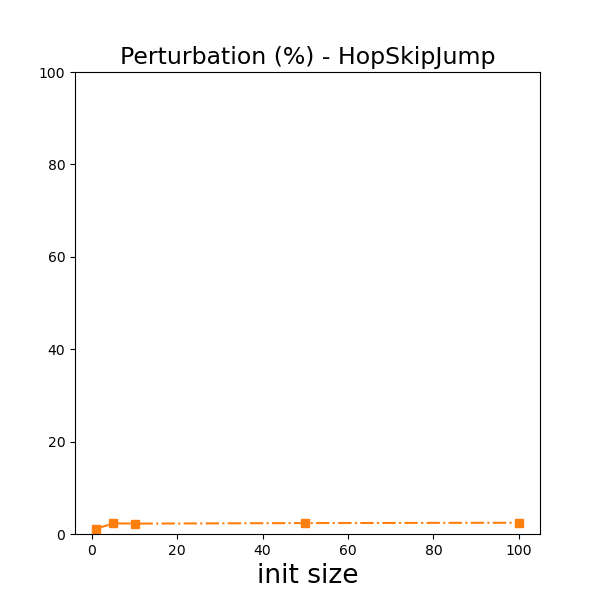}
    \end{subfigure}
    \begin{subfigure}{0.121\linewidth}
    \includegraphics[width=1.05\linewidth]{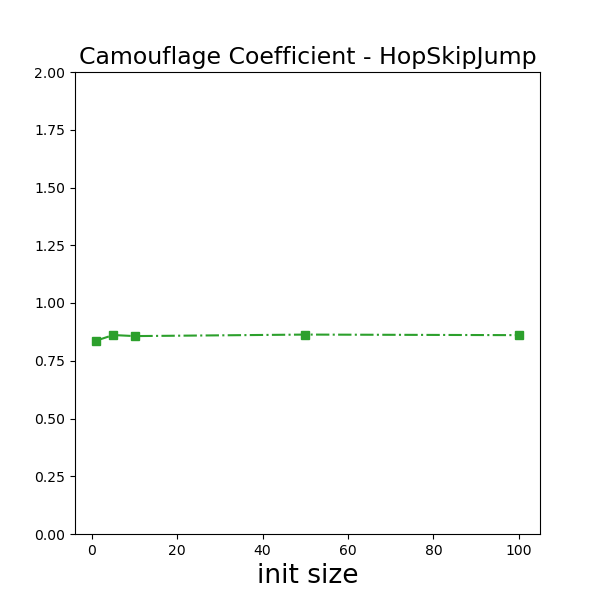}
    \end{subfigure}
\vspace{0.2cm}
\caption{The figures further show the impact of some important hyper-parameters on the performance of the rest of the benchmarks, including the $gamma$ for JSMA, the $learning$ $rate$ for C\&W, the $max$ $iter$ for C\&W and HopSkipJump, the $init$ $size$ for Boundary Attack and HopSkipJump, as well as the $delta$, $epsilon$, $min$ $epsilon$ and $num$ $trial$ for Boundary Attack. The specific values tried for these hyper-parameters are also listed in Table~\ref{tab:hyper_specific}. The reason why we separate these figures from Figure~\ref{fig:addExp2_1} is that the methods here are basically slower by more than one order of magnitude, so to illustrate the results completely, the y-axis of time cost here has to be normalized to a larger range (\ie [0, 20](s) as used in these figures), under which the time cost of all of the faster methods in Figure~\ref{fig:addExp2_1} may look approaching zero with the specific values and trends hard to recognize. \vspace{0.8cm}}
\label{fig:addExp2_2}
\end{figure*}

\subsection{On the Fairness of Results: Hyper-Parameters}\label{subsec:sm_addExp}

In this section, in order to confirm that our experimental results in Section~\ref{subsec:sm_result1} and Section~\ref{subsec:sm_result2} are really fair, representative and of general significance, we provide a series of additional experiments on the UCR-ECG200 dataset for the impact of specific values of some important hyper-parameters to the experimental methods. Specifically, we confirm that neither the benchmark methods merely perform worse under their default settings, which is supported by Table~\ref{tab:hyper_specific} with Figure~\ref{fig:addExp2_1} and Figure~\ref{fig:addExp2_2}, nor do we set the optimal hyper-parameters as the default ones for TSFool to gain any unfair advantages in the comparisons, as shown in Figure~\ref{fig:addExp1}. As aforementioned, Transfer Attack is implemented based on existing adversarial sets, so it can not be directly tested here. But again, the results of PGD (BIM) still provide a solid reference for it.

For TSFool, there are basically three kinds of hyper-parameters: 1) $K$, $T$ and $F$ for the extraction of i-WFA from RNN classifier; 2) $eps$ and $P$ for the control of adversarial perturbation; and 3) $N$, $C$ and $target$ for general task settings like implementing targeted attack. Some more detailed descriptions of them are also provided in our GitHub repository.\footref{fn:1} Here we focus on the $eps$ and $P$, which are directly related to the crafting of perturbation and most significantly impact the performance of TSFool. As shown in Figure~\ref{fig:addExp1}, with the increase of $eps$ and $P$, the perturbation $\rho^*$ and Camouflage Coefficient $\mathcal{C}$ show an upward trend in general. This is because as the definition of TSFool, both of these actions lead to larger distortion. For ASR, recall that different from the benchmark methods, $eps$ in TSFool is not used to directly control perturbation amount, but to fine-tune the amount of the random masking noise under the given constraint, so there is basically an upward trend with certain randomness. In real-world practice, we can adjust it case by case to seek for optimal performance beyond average. On the contrary, the ASR goes up regularly along with the increase of $P$, because the larger its value, the more the crafted adversarial sample approaches the VNS. Finally, smaller $eps$ and $P$ lead to slightly more time costs, because they respectively mean more steps for the interpolation sampling and more random masking noises added.

\subsection{On the Fairness of Results: Compared Data}\label{subsec:sm_fair}

In this section, we further discuss the fairness of the comparison results from the perspective of data. Careful readers may have noticed that, in Table~\ref{tab:average}, Table~\ref{tab:average_full} and Table~\ref{tab:multi}, the sizes of the generated adversarial sets of TSFool shown in ``Generation Number'' are not the same as the benchmarks. This is predictable because the benchmarks always perturb all the test samples without considering their specific distribution, while an important contribution for TSFool is to propose the concept of TPS and VNS, as well as the approach to specifically select them among the test samples. We understand that, compared with the current knowledge, such a novel case may lead to certain concerns. Someone may doubt if our comparison results are really fair as the final compared data are not exactly the same (\ie respectively the TPSs and all test samples for TSFool and benchmarks). Aiming at dealing with this concern, we provide discussions from two aspects as below, with additional experimental results.

\begin{table}[htbp]
\begin{center}
\renewcommand{\arraystretch}{1.15}
\resizebox{8.7cm}{!}{
\begin{tabular}{c|c|c|c|c}
\toprule
\multirow{2}{*}{\textbf{Method}} & \multirow{1.2}{*}{\textbf{Attack}} & \multirow{1.2}{*}{\textbf{Average}} & \multirow{1.2}{*}{\textbf{Perturbation}} & \multirow{1.2}{*}{\textbf{Camouflage}} \\
& \multirow{0.8}{*}{\textbf{Success Rate}} & \multirow{0.8}{*}{\textbf{Time Cost (s)}} & \multirow{0.8}{*}{\textbf{Ratio ($\rho^*$)}} & \multirow{0.8}{*}{\textbf{Coefficient}} \\
\midrule
\midrule
 FGSM & 83.94\% & \cellcolor{Melon}\underline{0.0024} & 77.20\% & 0.8984 \\
 JSMA & 79.43\% & 0.3577 & 8.95\% & 0.8361 \\
 DeepFool & 82.86\% & 0.0249 & 14.90\% & 0.9431 \\
 PGD (BIM) & 81.01\% & 0.1711 & 45.12\% & 0.8685 \\
 C\&W & 81.15\% & 3.1358 & 5.65\% & 0.8974 \\
 Auto-Attack & 84.29\% & 0.1746 & 45.10\% & 0.8667 \\
 \hspace{0.1em} Boundary Attack \hspace{0.1em} & 39.32\% & 5.6949 & 5.22\% & \cellcolor{Melon}\underline{0.6245} \\
 HopSkipJump & 83.22\% & 6.5404 & 5.19\% & 0.8693 \\
 Transfer Attack & 28.40\% & - & 7.07\% & 0.9513 \\
 \textbf{TSFool} & \cellcolor{SpringGreen}\underline{\textbf{87.76\%}} & \textbf{0.0230} & \cellcolor{SpringGreen}\underline{\textbf{4.63\%}} & \textbf{0.8147} \\
\bottomrule
\end{tabular}
}
\end{center}
\caption{Corresponding to Table~\ref{tab:average}, this table also shows the average performance of the experimental methods on the 10 UCR time series datasets, while the only change is that the benchmarks are no longer evaluated on all test samples but on the TPSs selected by TSFool. Even under this unfavorable situation (\ie in fact, the benchmarks should not have known the TPSs), TSFool is still the best in general and our conclusions remain unchanged. This is to deal with a possible concern about the fairness of our comparisons regarding the specific data they use.}
\label{tab:rebut3-2}
\end{table}

Firstly, from \textit{reasonability}, this situation is not a factitious setting, but a natural result due to the difference between TSFool and the benchmarks. Actually, the certain fairness considered here is ``to begin all the algorithms from the same dataset'', which fits the real-world practices. On the contrary, if we just compare the experimental methods in TPSs, then we unreasonably assume that the benchmarks also have the ability to capture them, which is unfair to TSFool. Secondly, to say the least, even if we do make comparisons only on TPSs as shown in Table~\ref{tab:rebut3-2}, the results are very similar and the conclusions in our manuscripts remain unchanged. Specifically, compared with Table~\ref{tab:average} under the same setting, TSFool remains the best in ASR and sub-optimal in efficiency, and even becomes the best in $\rho^*$. This is predictable because, without the guidance provided by VNSs, the benchmarks are not expected to perturb TPSs better than average on all test samples. For the CC, it can be found that the benchmarks do benefit from the TPSs, which also confirms the effectiveness of our idea in selecting them. But still, TSFool only slightly drops from optimal to sub-optimal, behind the Boundary Attack. Considering the unsatisfying performance of Boundary Attack in the other three measures, it can hardly challenge TSFool in general. So we can conclude that, from \textit{necessity}, there is also no need to worry about whether to evaluate the benchmarks on TPSs or all test samples, as this will not impact our conclusions.

\begin{table*}[htbp]
\begin{center}
\renewcommand{\arraystretch}{1.15}
\resizebox{15.5cm}{!}{
\begin{tabular}{c|c|ccccccc|c|c}
\toprule
\multirow{2}{*}{\textbf{Study}} & \multirow{2}{*}{\textbf{Option}} & \multicolumn{7}{c|}{\multirow{0.9}{*}{\textbf{Question}}} & \hspace{0.3em} \multirow{2}{*}{\textbf{Sum}} \hspace{0.3em} & \multirow{2}{*}{\textbf{Count}} \\
\cline{3-9} 
 & & \multirow{1.3}{*}{\textbf{1}} & \multirow{1.3}{*}{\textbf{2}} & \multirow{1.3}{*}{\textbf{3}} & \multirow{1.3}{*}{\textbf{4}} & \multirow{1.3}{*}{\textbf{5}} & \multirow{1.3}{*}{\textbf{6}} & \multirow{1.3}{*}{\textbf{7}} & & \\
\midrule
\midrule
\multirow{3}{*}{Study 1} & The Original Class Cluster & 20 & 13 & \textcolor{Green}{55} & \textcolor{Green}{37} & \textcolor{Green}{47} & \textcolor{Green}{55} & \textcolor{Green}{31} & 258 & \textcolor{Green}{5} \\
 & The Misclassified Class Cluster & \textcolor{Green}{41} & \textcolor{Green}{48} & 4 & 22 & 13 & 8 & 28 & 164 & 2 \\
 & Neutral & 4 & 4 & 6 & 6 & 5 & 2 & 6 & 33 & 0 \\
\midrule
\multirow{3}{*}{Study 2} & The Adversarial Sample from TSFool & \textcolor{Green}{54} & \textcolor{Green}{58} & \textcolor{Green}{51} & \textcolor{Green}{57} & \textcolor{Green}{58} & \textcolor{Green}{60} & \textcolor{Green}{58} & 396 & \textcolor{Green}{7} \\
 & The Adversarial Sample from PGD & 10 & 5 & 12 & 6 & 4 & 3 & 5 & 45 & 0 \\
 & Neutral & 1 & 2 & 2 & 2 & 3 & 2 & 2 & 14 & 0 \\
\bottomrule
\end{tabular}
}
\end{center}
\vspace{0.2cm}
\caption{The table shows the statistical results of the two kinds of human studies, where the ``Sum'' calculates the total number of votes and the ``Count'' reports the final voting results. Notice that the ``Option'' list is just provided here to benefit understanding, while the volunteers were unaware of this information when participating in the human studies. \vspace{0.85cm}}
\label{tab:human_study}
\end{table*}

\subsection{Human Study for Camouflage Coefficient}\label{subsec:sm_human_study}

In this section, through two kinds of subjective human studies with 65 volunteers, we confirm that Camouflage Coefficient is an effective measure in representing the real-world imperceptibility of adversarial samples. Our volunteers consist of 29 men and 36 women aged from 18 to 60. The studies are conducted on the UCR-ECG200, which is a binary classification dataset.

\begin{figure}[htb]
\centerline{\includegraphics[width=8cm]{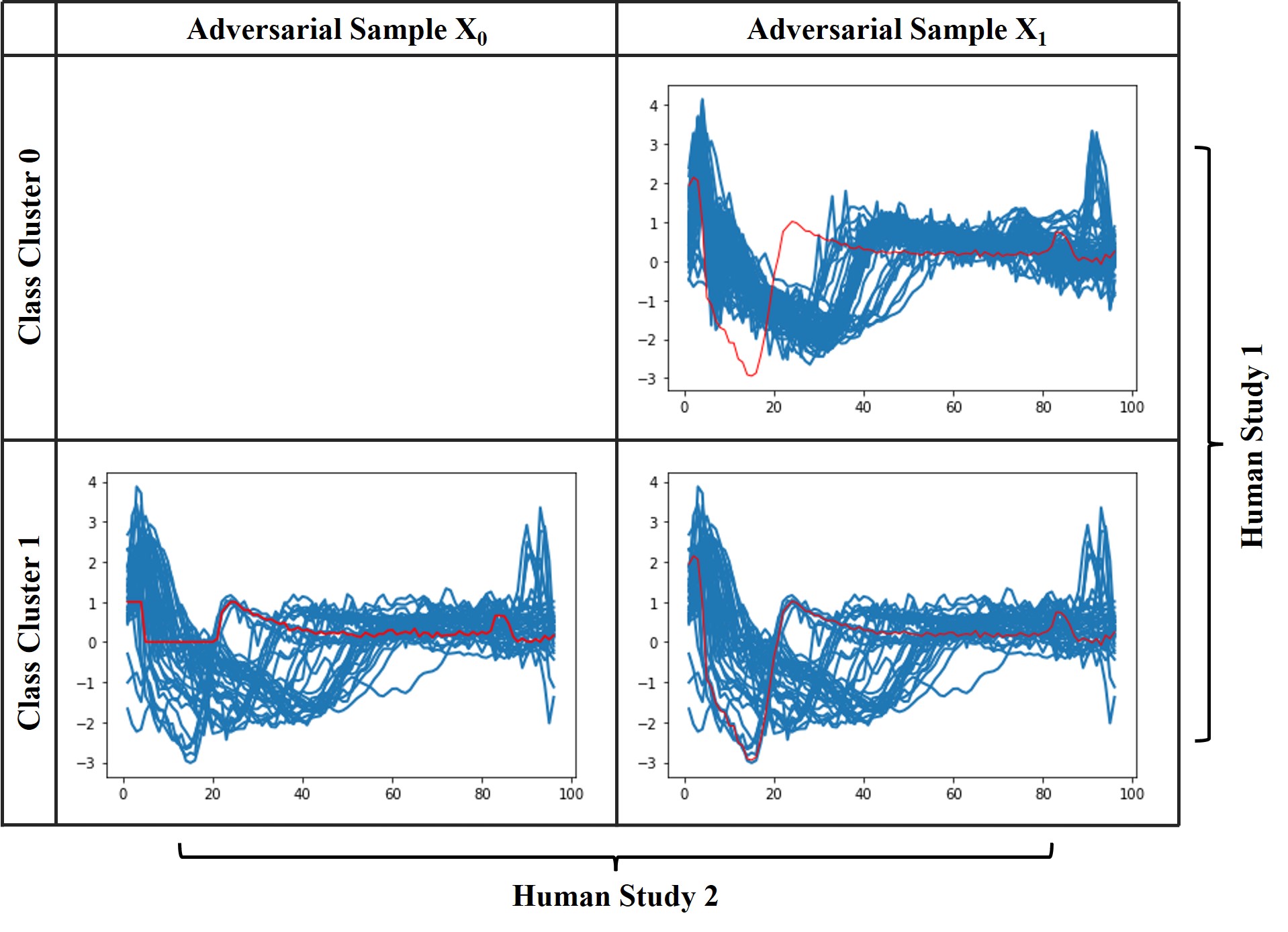}}
\vspace{0.32cm}
\caption{The figure illustrates an instance of our two human studies. The adversarial samples are in \textcolor{Red}{red} and other samples from the class clusters are in \textcolor{cvprblue}{blue}. In brief, the first study is for the imperceptibility of the same adversarial sample with a small CC value in different class clusters, while the second study is for the imperceptibility of two different adversarial samples with different CC values in the same original class cluster. In this case, the adversarial sample $X_0$ is from PGD while $X_1$ is from TSFool, and the original cluster of their target sample is Cluster 1. For Study 1, 84.62\% of volunteers think $X_1$ fits Cluster 1 better than Cluster 0. And for Study 2, 78.46\% of volunteers believe $X_1$ better fits Cluster 1 than $X_0$. Both of these confirm the effectiveness of CC in representing real-world imperceptibility. \vspace{1.3cm}} \label{fig:human_study}
\end{figure}

In the first study, for each of the seven questions, we provide two figures with \textit{\textbf{one}} adversarial sample crafted by TSFool and the \textit{\textbf{two}} class clusters (\ie the original cluster and the misclassified cluster respectively), and ask the volunteers to choose in which figure the adversarial sample fits the class cluster better. This is to show if the specific adversarial samples with small CC values (\ie averagely 0.6291 by TSFool on ECG200) indeed visually fit the original class better than the misclassified class. Then in the second study, also for each of the seven questions, we provide two figures with \textit{\textbf{two}} different adversarial samples generated from the same target sample respectively by TSFool and PGD, and \textit{\textbf{one}} original class cluster of the target sample. The question to be answered is still, in which figure the adversarial sample fits the class cluster better. This is to verify if the adversarial samples with smaller CC values (\ie from TSFool, also 0.6291 on average) can hide in the original class better than the adversarial samples with larger CC values (\ie from PGD, 0.9705 on average). There is an example illustrating both of these two human studies in Figure~\ref{fig:human_study}.

The statistical results of the human studies are illustrated in Table~\ref{tab:human_study}. It can be found in Study 1 that five of seven adversarial samples with small CC values are believed closer to the original class than the misclassified class. And in Study 2, all the seven cases support that adversarial samples with smaller CC values can better fit the original class cluster. As a result, we believe that Camouflage Coefficient can indeed capture the real-world imperceptibility of adversarial samples from the perspective of class distribution. This lays a solid foundation for the imperceptibility of TSFool attack from the perspective of human recognition.

\subsection{Capability in Escaping Anomaly Detection}\label{subsec:sm_anomaly}

In real-world practices, except for avoiding being noticed by humans, another important challenge for adversarial attacks is to escape from automatic defense mechanisms \citep{zhang2021adversarial}. In the field of TSC, anomaly detection is a common choice for the defense \citep{blazquez2021review,schmidl2022anomaly}. So in this section, beyond standard evaluation metrics and human perception, we further demonstrate the imperceptibility of TSFool attack from the perspective of anomaly detection. Specifically, we adopt four popular time series anomaly detection methods, namely One-Class SVM (OCSVM) \citep{scholkopf2001estimating}, Isolation Forest (IF) \citep{liu2012isolation}, Local Outlier Factor (LOF) \citep{breunig2000lof} and LSTM Outlier Detection (LSTMOD) \citep{ergen2019unsupervised}. Our specific implementation of these methods is based on Automated Time-series Outlier Detection System (TODS) \citep{lai2021tods}. As shown in Table~\ref{tab:rebut1-1}, compared with two white-box benchmarks (\ie PGD and C\&W) and one black-box benchmark (\ie HopSkipJump), TSFool is the most hardly detectable attack against anomaly detection in general. This further boosts our confidence in the imperceptibility of TSFool to real-world defense.

\newpage

\begin{table}[htbp]
\begin{center}
\renewcommand{\arraystretch}{1.15}
\resizebox{8.7cm}{!}{
\begin{tabular}{c|c|cccc}
\toprule
\multirow{2}{*}{\textbf{Method}} & \multirow{2}{*}{\textbf{Metric}} & \multicolumn{4}{c}{\multirow{0.7}{*}{\textbf{Time Series Anomaly Detection}}} \\
& & \multirow{1.3}{*}{\textbf{OCSVM}} & \multirow{1.3}{*}{\textbf{IF}} & \multirow{1.3}{*}{\textbf{LOF}} & \multirow{1.3}{*}{\textbf{LSTMOD}} \\
\midrule
\midrule
\multirow{3}{*}{PGD (BIM)} & $Pre$ & \hspace{0.4em} 0.1755 \hspace{0.2em} & \hspace{0.2em} 0.1899 \hspace{0.2em} & \hspace{0.2em} 0.2794 \hspace{0.2em} & \hspace{0.2em} 0.2107 \hspace{0.4em} \\
& $Re$ & 0.4890 & 0.5377 & 0.7018 & 0.7091 \\
& $F1$ & 0.2454 & 0.2692 & 0.3782 & 0.3092 \\
\midrule
\multirow{3}{*}{C\&W} & $Pre$ & 0.0637 & 0.0534 & 0.0463 & 0.0968 \\
& $Re$ & 0.1304 & 0.1201 & \textbf{0.0745} & 0.3377 \\
& $F1$ & 0.0798 & 0.0696 & \textbf{0.0534} & 0.1432 \\
\midrule
\multirow{3}{*}{HopSkipJump} & $Pre$ & 0.0693 & 0.0473 & 0.0801 & 0.1381 \\
& $Re$ & 0.1561 & 0.1115 & 0.2282 & 0.5127 \\
& $F1$ & 0.0897 & 0.0640 & 0.1146 & 0.2129 \\
\midrule
\multirow{3}{*}{TSFool} & $Pre$ & \textbf{0.0505} & \textbf{0.0346} & \textbf{0.0460} & \textbf{0.0741} \\
& $Re$ & \textbf{0.1012} & \textbf{0.0829} & 0.1274 & \textbf{0.3218} \\
& $F1$ & \textbf{0.0622} & \textbf{0.0469} & 0.0639 & \textbf{0.1175} \\
\bottomrule
\end{tabular}
}
\end{center}
\vspace{0.1cm}
\caption{Given adversarial samples respectively generated by TSFool and three common white-box or black-box benchmarks on the 10 UCR time series datasets, this table illustrates the average results of four time series anomaly detection methods on these adversarial samples. Specifically, according to the conventional practice in the field of anomaly detection, we report $Precision$ ($Pre$), $Recall$ ($Re$) and $F_{1} \text{-} \textit{score}$ ($F1$) for every single sub-experiment. It can be found that TSFool performs exactly the best under three of the four detection methods, while the only slight competition is from C\&W under LOF. This further verifies the outstanding imperceptibility of TSFool beyond existing methods in real-world anomaly detection.}
\label{tab:rebut1-1}
\end{table}

\section{Further Discussion}\label{sec:sm_discu}

In this section, we supplement two case studies and an additional experiment to further support our discussion about the potentials of TSFool in Section~\ref{sec:discu}. Specifically, by using DTW as a measure of sample similarity, we first show the smaller distortion between benign and adversarial time series crafted by TSFool compared with PGD in Section~\ref{subsec:sm_dtw}. Secondly, we explore the potential of using TSFool for adversarial training in Section~\ref{subsec:sm_at}, showing that it is competitive with the current standard approach. Finally, in Section~\ref{subsec:sm_extend}, we discuss the weakness of TSFool in not exactly generating one adversarial sample for every single benign one, and further provide an extended version of TSFool to deal with this weakness.

\subsection{Small Distortion under Dynamic Time Warping}\label{subsec:sm_dtw}

Time series poses unique challenges for studying adversarial robustness that are not seen in images \citep{kolter2018tutorial}, and a recent related work proposed that the standard $\ell_p$-norm metric does not capture the true similarity between time series \citep{belkhouja2022dynamic}. It is argued that adversarial time series crafted under $\ell_p$-norm constrain can semantically correspond to a completely different class label, and as a consequence, mislead the improvement of model robustness \citep{belkhouja2022dynamic}. To deal with this problem, \citet{belkhouja2022dynamic} proposed to use DTW as a measure for the realistic distortion from benign time series samples to the corresponding adversarial ones.

The DTW measure between two univariate time series $X_1, X_2 \in \mathbb{R}^{T}$ (\ie $D = 1$) is based on a cost matrix $M \in \mathbb{R}^{T \times T}$, which is built
recursively by the following equation \citep{belkhouja2022dynamic}:
\begin{equation}\label{eq:dtw_m}
\begin{aligned}
    M_{(i, \hspace{0.05em} j)} = \hspace{0.3em} & dist \hspace{0.1em} (X_{1(i)}, X_{2(j)}) \\ & + \text{min} \hspace{0.1em} \{ M_{(i-1, \hspace{0.05em} j)}, M_{(i, \hspace{0.05em} j-1)}, M_{(i-1, \hspace{0.05em} j-1)} \},
\end{aligned}
\end{equation}
where $dist(\cdot, \cdot)$ is the distance under any given metric (\eg $\ell_p$-norm). Then the DTW measure $DTW(X_1, X_2)$ can be calculated as:
\begin{equation}\label{eq:dtw}
\resizebox{7.5cm}{!}{
$\begin{aligned}
    DTW(X_1, X_2) = \sum_{(i, j) \in P}dist \hspace{0.1em} (X_{1(i)}, X_{2(j)}) = M_{(T, \hspace{0.05em} T)},
\end{aligned}$
}
\end{equation}
with the sequence $P = \{ c_{(i, \hspace{0.05em} j)} = (i, j) \}$ contributing to $M_{(T, \hspace{0.05em} T)}$ is the \textit{optimal alignment path} between $X_1$ and $X_2$, where $m_{(\cdot, \cdot)}$ denotes the element in the cost matrix $M$.

\begin{figure}
  \centering
  \begin{subfigure}{0.49\linewidth}
  \includegraphics[width=0.97\linewidth]{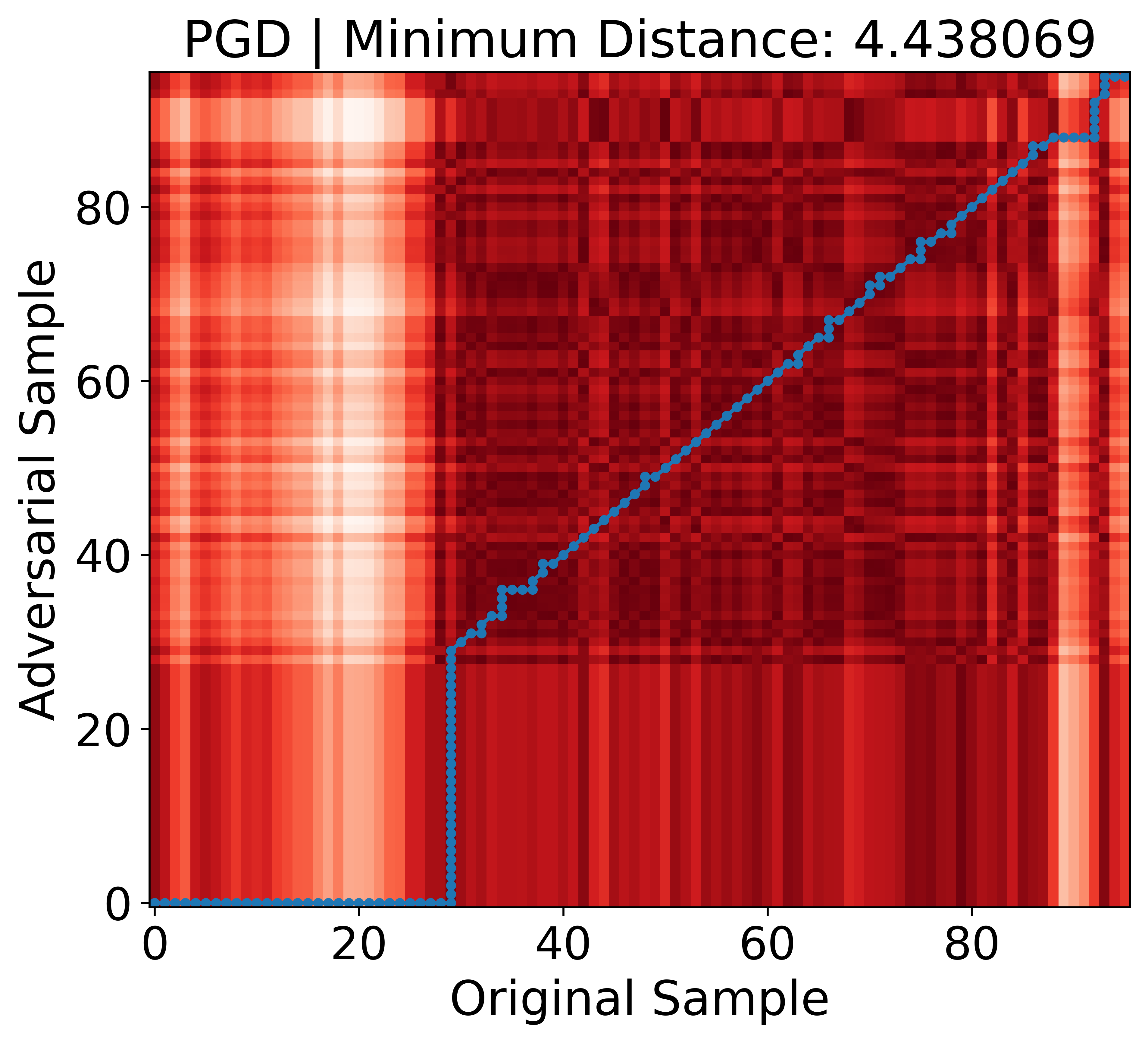}
  \end{subfigure}
  \begin{subfigure}{0.49\linewidth}
  \includegraphics[width=0.97\linewidth]{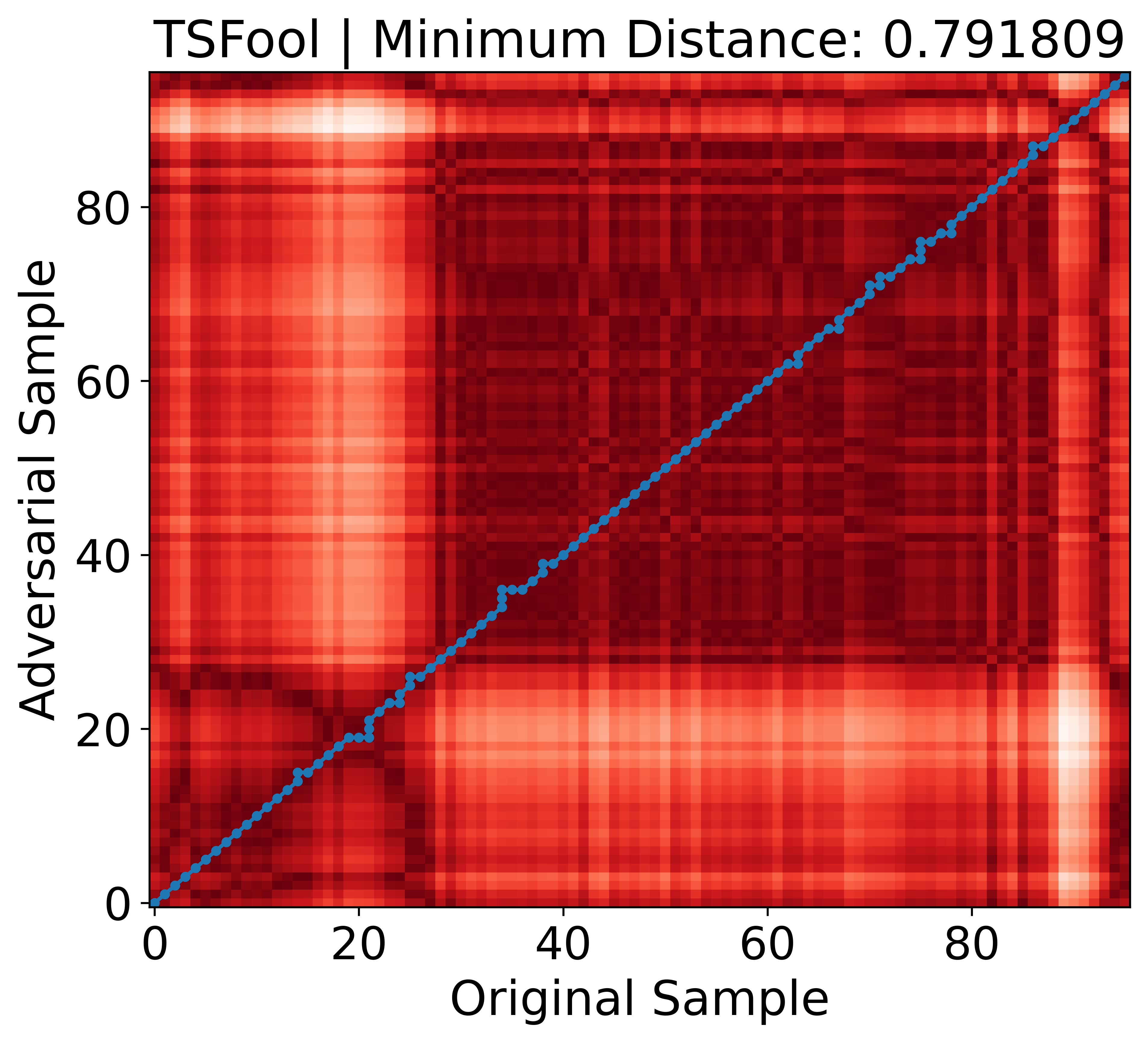}
  \end{subfigure}
  \vspace{0.35cm}
  \caption{An instance of DTW distortion measure. The two figures show the cost matrixes for a specific time series on ECG200 dataset and the corresponding adversarial samples respectively from PGD (left) and TSFool (right), with deeper \textcolor{Red}{red} representing lower cost and the \textit{optimal alignment path} shown in \textcolor{cvprblue}{blue}. Intuitively, with a better diagonal symmetry of matrix pattern and the \textit{optimal alignment path} approaching the diagonal path, the TSFool-crafted adversarial time series is more similar to the original one. The minimum DTW distances calculated (\ie 4.4381 for PGD \textit{vs.} 0.7918 for TSFool) also support this statement. Besides, this even improves our confidence in TSFool to escape from \textit{outlier detection}, as there are basically no outliers observed in the right figure, especially compared with PGD in the left one. \vspace{0.3cm}}
  \label{fig:dtw}
\end{figure}

To facilitate the discussion in Section~\ref{sec:discu}, we implement an additional case study on the UCR-ECG200 dataset, using DTW to measure the similarity between original and adversarial time series respectively crafted by PGD and TSFool. Specifically, a pair of time series $X_{ori}, X_{adv}$ with a lower $DTW(X_{ori}, X_{adv})$ value is closer in minimum DTW distance, which means less distortion under DTW measure, providing another measure of adversarial imperceptibility. Our results show that the average minimum distance between the original and adversarial time series from PGD is 1.1685, and that from TSFool is 0.6397, which means under DTW measure, the adversarial time series crafted by TSFool are still far more similar to the benign ones than by PGD. We also provide an intuitive instance in Figure~\ref{fig:dtw}.

\vspace{0.05cm}

\subsection{Potential of TSFool for Adversarial Training}\label{subsec:sm_at}

With the development of adversarial attack technologies, there are increasing concerns about how to acquire NNs that are robust to adversarial inputs. One of the promising directions is to study adversarial robustness through the lens of robust optimization, which is also known as ``adversarial training'' \citep{goodfellow2015explaining}. Different from the common training based on \textit{empirical risk minimization} (ERM), which is a successful recipe for finding classifiers with small population risk but does not often yield robust models \citep{biggio2013evasion, szegedy2014intriguing}, adversarial training augments the ERM paradigm to:
\begin{equation}\label{eq:at1}
    \mathbb{E}_{\Vec{x}, \Vec{y} \sim D} \left[\max_{\Vec{x}' \in P(\Vec{x})} \mathcal{L}(f, \Vec{x}', \Vec{y})) \right],
\end{equation}
where $D$ is the given data distribution and $P(x)$ is a pre-defined perturbation set, to directly learn the underlying concepts in a robust manner \citep{madry2018towards}. Accordingly, adversarial training directly corresponds to the optimization of the following $\min$-$\max$ problem:
\begin{equation}\label{eq:at2}
    \min_\theta \hspace{0.1em} \sum_{i=1}^n \max_{\Vec{x}' \in P(\Vec{x}_i)} \mathcal{L}(f_\theta, \Vec{x}', \Vec{y}_i)),
\end{equation}
to find model parameters $\theta$ that lead to minimal empirical loss even under the strongest adversarial perturbation. 

To solve this problem, based on \textit{Danskin’s Theorem} \citep{danskin2012theory}, \citet{madry2018towards} propose to find a constrained maximizer of the inner maximization problem, and then use it as the actual data point to compute the gradient for the outer minimization problem. Empirically, adversarial samples crafted by PGD are believed sufficiently close to the optimal solution of the former problem, and common optimizers like \textit{stochastic gradient descent} (SGD) can certainly solve the latter problem well  \citep{madry2018towards}. In practice, the corresponding process is simply generating adversarial samples upon the training set and appending them for model retraining (or directly training from scratch).

\begin{figure}
  \centering
  \begin{subfigure}{0.49\linewidth}
    \includegraphics[width=1\linewidth]{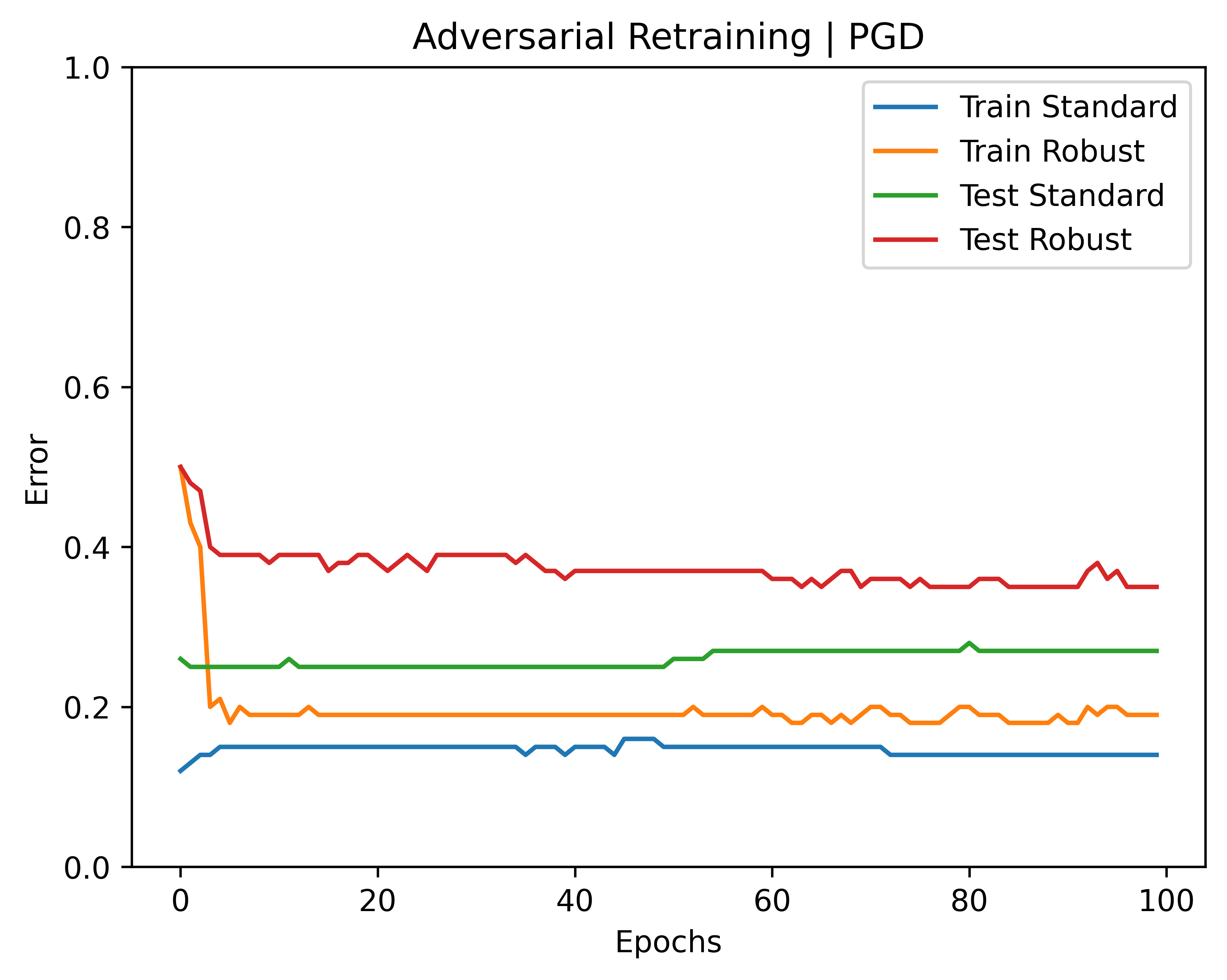}
  \end{subfigure}
  \begin{subfigure}{0.49\linewidth}
    \includegraphics[width=1\linewidth]{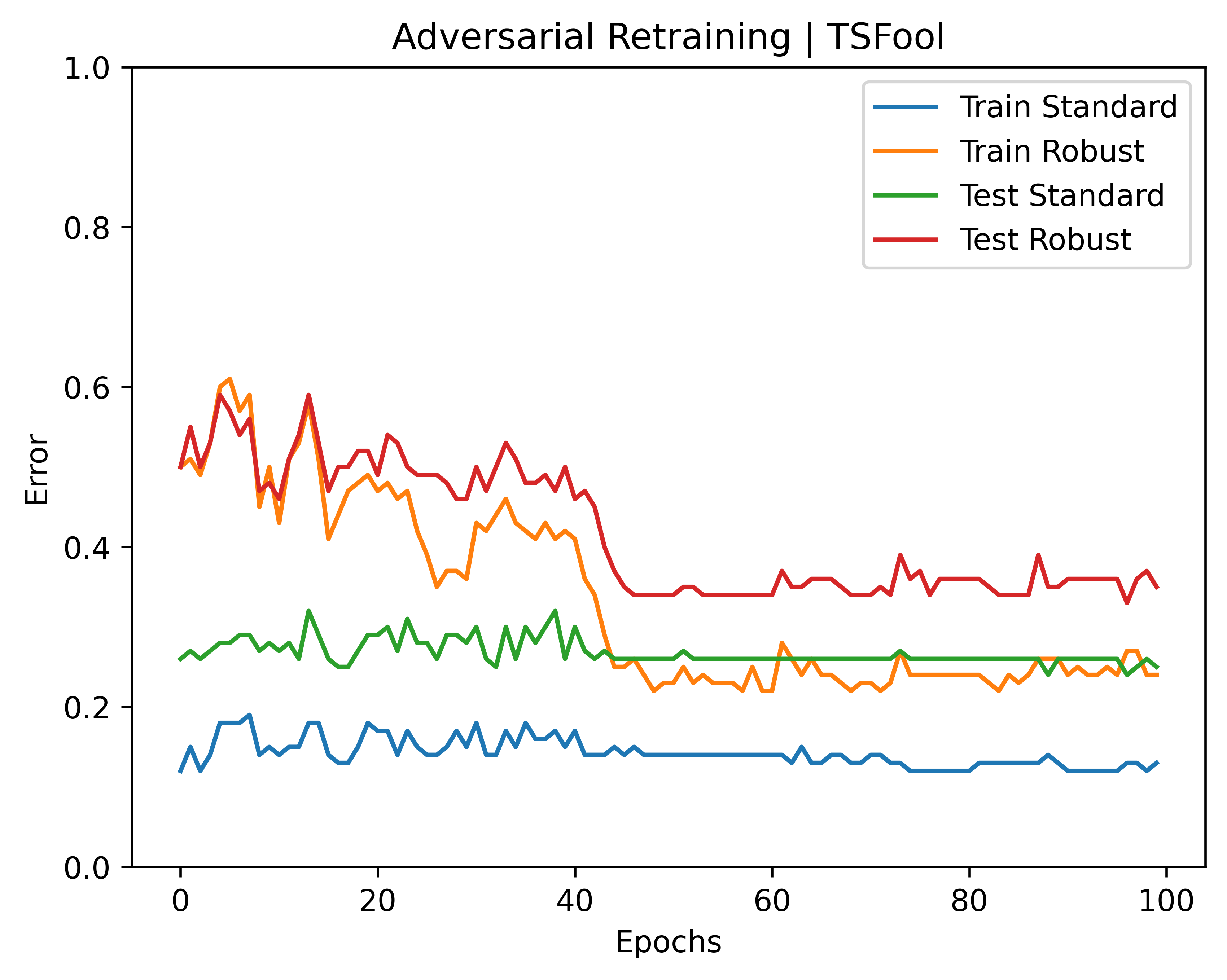}
  \end{subfigure}
  \vspace{0.4cm}
  \caption{The figures respectively show instances for the adversarial retraining process based on PGD and TSFool, with our experimental LSTM classifier on UCR-ECG200 dataset as the target model, and the training epoch set to be 100 as well as the learning rate to be 0.001. The ``Standard'' and ``Robust'' are respectively used to represent the \textit{classification error} under benign and adversarial settings. There are two points to be noticed corresponding to our statements in Section~\ref{subsec:sm_at}. Firstly, the optimal checkpoints of the ``Test Standard'' and the ``Test Robust'' of PGD and TSFool retraining are basically at the same level. Secondly, in general, the ``Train Standard'' and the ``Test Standard'' of PGD increase along with the retraining process, while both of them of TSFool decrease.}
  \label{fig:compare_at}
\end{figure}

In this case study, we use TSFool to replace PGD in the adversarial retraining of our experimental LSTM classifier on UCR-ECG200 dataset, to reveal the potential of TSFool in improving its robustness. Notice that adversarial training is not a direct target of our work, so we do not expect TSFool to beat PGD, which is still a general state-of-the-art method in this field. However, benefits from the comparison between them, we can still give some interesting statements. On the one hand, as shown in Figure~\ref{fig:compare_at}, the ``Test Standard'' and ``Test Robust'' of TSFool are basically at the same level of PGD at the end of the retraining process. According to the definition, in order to train a classifier minimizing Equation~(\ref{eq:at2}), we have to approach the strongest adversary possible that can find gradients of the robust loss sufficiently close to optimal \citep{madry2018towards}. So the comparison results naturally support two statements: firstly, TSFool is a sufficiently strong adversarial attack method; and secondly, the multi-objective optimization adopted by TSFool does not make its attack results seriously violate the results under the conventional objective of adversarial attack.

\vspace{0.2cm}

\newpage

\begin{table}[htbp]
\begin{center}
\resizebox{8.7cm}{!}{
\begin{tabular}{c|c|c|c|c}
\toprule
\hspace{0.3em} \multirow{2}{*}{\textbf{Dataset}} \hspace{0.3em} & \multirow{1.2}{*}{\textbf{Attack}} & \multirow{1.2}{*}{\textbf{Average}} & \multirow{1.2}{*}{\textbf{Perturbation}} & \multirow{1.2}{*}{\textbf{Camouflage}} \\
& \multirow{0.8}{*}{\textbf{Success Rate}} & \multirow{0.8}{*}{\textbf{Time Cost (s)}} & \multirow{0.8}{*}{\textbf{Ratio ($\rho^*$)}} & \multirow{0.8}{*}{\textbf{Coefficient}} \\
\midrule
\midrule
 CBF & 73.78\% & 0.0465 & 7.42\% & 0.7194 \\
 DPOAG & 84.89\% & 0.0326 & 8.53\% & 0.8458 \\
 DPOC & 93.48\% & 0.0472 & 6.79\% & 0.8439 \\
 ECG200 & 92.00\% & 0.0365 & 10.44\% & 0.7698 \\
 GP & 90.67\% & 0.1080 & 16.61\% & 0.8381 \\
 IPD & 87.56\% & 0.0131 & 12.60\% & 0.8257 \\
 MPOAG & 84.42\% & 0.0338 & 12.65\% & 0.7809 \\
 MPOC & 91.07\% & 0.0366 & 17.11\% & 0.8366 \\
 PPOAG & 60.98\% & 0.0838 & 8.54\% & 0.6486 \\
 PPOC & 80.41\% & 0.0566 & 6.36\% & 0.6806 \\
\midrule
 \textbf{Average} & \cellcolor{SpringGreen}\underline{\textbf{83.93\%}} & \textbf{0.0495} & \textbf{10.71\%} & \cellcolor{SpringGreen}\underline{\textbf{0.7789}} \\
\bottomrule
\end{tabular}
}
\end{center}
\vspace{0.1cm}
\caption{Corresponding to Table~\ref{tab:average} and Table~\ref{tab:average_full}, this table 
shows the specific performance of the extended TSFool on the 10 UCR time series datasets, as well as the average of them. Compared with the standard TSFool, these results are acquired directly on all test samples, exactly following the default setting in the evaluation of the benchmarks. It can be found that, just like the standard one, the extended version of TSFool still performs the best, beating all the benchmarks in general. This is to dispel a possible doubt about the applicability of TSFool in real-world practices (\ie if the novel design in TSFool is acceptable for the specific case, the standard version can be used, while if not, the extended version can serve as a complete backup option as it exactly fits the existing knowledge). \vspace{0.5cm}}
\label{tab:rebut3-1}
\end{table}

On the other hand, although it is widely believed that the standard error of NNs will undesirably increase as a result of adversarial training \citep{madry2018towards}, and this is indeed confirmed by the ``Train Standard'' and ``Test Standard'' of PGD retraining in Figure~\ref{fig:compare_at}, we found that these two errors in TSFool retraining show a decreasing trend in the late range of the training process, and the optimal points even better than the points before retraining. In other words, TSFool-based adversarial training is less destructive to standard performance. A possible reason for this observation is that the adversarial samples from TSFool sufficiently fit the benign ones from the perspective of the distribution of class clusters, so that the model can smoothly learn the new knowledge under the adversarial setting without breaking the already-learned perception from the benign samples. This might be instructive for future exploration in the field of adversarial training.

\subsection{Weakness and Extension of TSFool}\label{subsec:sm_extend}

In Section~\ref{subsec:sm_fair}, on the fairness of evaluation for the benchmarks, we have confirmed the consistency of the experimental results on TPSs and all test samples. At the same time, someone may also wonder if there is the same case for TSFool. Specifically, whether TSFool can directly work for every single test sample, instead of beginning from all test samples but finally targeting at the specifically selected TPSs? However, it should be noted that, for the benchmarks, the discussion is just about two different evaluation settings. But for TSFool, that directly corresponds to the specific design of its mechanism. So in this section, given the above question, we discuss a specific weakness in the mechanism of TSFool regarding its applicability, and further provide an extended version of TSFool against the weakness.

As proposed in Section~\ref{subsec:reconsider} and further explained in Section~\ref{subsec:sm_fair}, selecting specific samples (\ie TPSs and VNSs) among the test samples based on the specific distribution can benefit the control of perturbation. On one side, this is an advantage of TSFool as it fills in a gap in the current knowledge and helps to reduce the distortion of adversarial samples. While on the other side, this certainly violates the current default setting in the evaluation of attack methods to exactly generate one adversarial sample for every single benign one. So from the latter view, this also becomes a weakness of TSFool. Fortunately, by simply modifying the strategy in matching the corresponding TPS and VNS, TSFool can be easily extended to finally target at all test samples. Specifically, the only change is, in Section~\ref{subsubsec:perturb1}, from ``for each VNS, finding the closest test sample as TPS'' to ``for each test sample, finding the closest VNS and serving as its TPS'' in the ``Interpolation Sampling'' process. 

As shown in Table~\ref{tab:rebut3-1}, we evaluate this extended version of TSFool on all test samples, just as the default setting for the benchmarks. Compared with the results on Table~\ref{tab:average} and Table~\ref{tab:average_full}, similarly supported by VNS, the extended TSFool still performs the best in ASR and CC. On the other hand, the $\rho^*$ drops by the middle level, which is also predictable as this version no longer benefits from the specific 

\newpage

\noindent selection of TPS. An interesting angle is to view this as an ablation study for the standard TSFool, which also supports our motivation and contribution to introduce the concept of TPS. All in all, even if just evaluated on all test samples without considering TPSs, the extended TSFool is still the most competitive method, outperforming all the benchmarks in general. We will also release the extended TSFool together with the basic one to support different real-world practices. Specifically, if there is a requirement to generate adversarial samples for exactly every appointed data, the extended TSFool would be the certain choice. On the contrary, if the target is to pursue more threatening attacks against specific applications or systems to evaluate their robustness and find their defect, the standard TSFool would be recommended more, so that the target can benefit from the more imperceptible adversarial samples crafted.

\end{document}